\documentclass[a4paper,fleqn]{cas-dc}

\usepackage[numbers]{natbib}
\usepackage{framed,multirow}
\usepackage{lscape}
\usepackage{subcaption}
\usepackage{longtable}
\usepackage{academicons}
\usepackage{threeparttable}
\usepackage{placeins}
\usepackage{float}

\usepackage{url}
\usepackage{xcolor}

\def\tsc#1{\csdef{#1}{\textsc{\lowercase{#1}}\xspace}}
\tsc{WGM}
\tsc{QE}
\tsc{EP}
\tsc{PMS}
\tsc{BEC}
\tsc{DE}

\begin{document}
\let\WriteBookmarks\relax
\def\floatpagepagefraction{1}
\def\textpagefraction{.001}

\shorttitle{Recent Advancements in End-to-End Autonomous Driving using Deep Learning: A Survey}

\shortauthors{ P.S. Chib et~al.}

\title [mode = title]{Recent Advancements in End-to-End Autonomous Driving using Deep Learning: A Survey}

\author[1]{Pranav Singh Chib}[type=editor,
                        auid=000,bioid=1,
                        prefix=,
                        orcid=0000-0003-4930-3937]

\affiliation[1]{organization={Department of Computer Science and Engineering, Indian Institute of Technology Roorkee},
    city={Roorkee},
    state={Uttarakhand},
    country={India}}

\author[1]{Pravendra Singh}[type=editor,
                        auid=000,bioid=1,
                        prefix=,
                        orcid=0000-0003-1001-2219]

\cormark[1]

\ead{pravendra.singh@cs.iitr.ac.in}

\cortext[cor1]{Corresponding author: Pravendra Singh}

\begin{abstract}
End-to-End driving is a promising paradigm as it circumvents the drawbacks associated with modular systems, such as their overwhelming complexity and propensity for error propagation. Autonomous driving transcends conventional traffic patterns by proactively recognizing critical events in advance, ensuring passengers safety and providing them with comfortable transportation, particularly in highly stochastic and variable traffic settings. This paper presents a comprehensive review of the End-to-End autonomous driving stack. It provides a taxonomy of automated driving tasks wherein neural networks have been employed in an End-to-End manner, encompassing the entire driving process from perception to control. Recent developments in End-to-End autonomous driving are analyzed, and research is categorized based on underlying principles, methodologies, and core functionality. These categories encompass sensorial input, main and auxiliary output, learning approaches ranging from imitation to reinforcement learning, and model evaluation techniques. The survey incorporates a detailed discussion of the explainability and safety aspects. Furthermore, it assesses the state-of-the-art, identifies challenges, and explores future possibilities. We maintain the latest advancements and their corresponding open-source implementations at this \href{https://github.com/Pranav-chib/End-to-End-Autonomous-Driving}{link}.

\end{abstract}


\maketitle

\section{Introduction}

Autonomous driving refers to the capability of a vehicle to drive partly or entirely without human intervention. The modular architecture \cite{chen2015deepdrivingc22,maddern20171c26,akai2017autonomousc27,kong2015kinematicc51,zhao2012designc50} is a widely used approach in autonomous driving systems, which divides the driving pipeline into discrete sub-tasks. This architecture relies on individual sensors and algorithms to process data and generate control outputs. It encompasses interconnected modules, including perception, planning, and control. However, the modular architecture has certain drawbacks that impede further advancements in autonomous driving (AD). One significant limitation is its susceptibility to error propagation. For instance, errors in the perception module of a self-driving vehicle, such as misclassification, can propagate to subsequent planning and control modules, potentially leading to unsafe behaviors. Additionally, the complexity of managing interconnected modules and the computational inefficiency of processing data at each stage pose additional challenges associated with the modular approach. To address these shortcomings, an alternative approach called End-to-End driving \cite{hanselmann2022king2orc99,chitta2022transfuserc116,shao2023safetyc117,hu2023planningoriented,chen2022learning5orc98,zhang2022learning} has emerged. This approach aims to overcome the limitations of the modular architecture.

\begin{figure}[t]
\centerline{\includegraphics[scale=0.15]{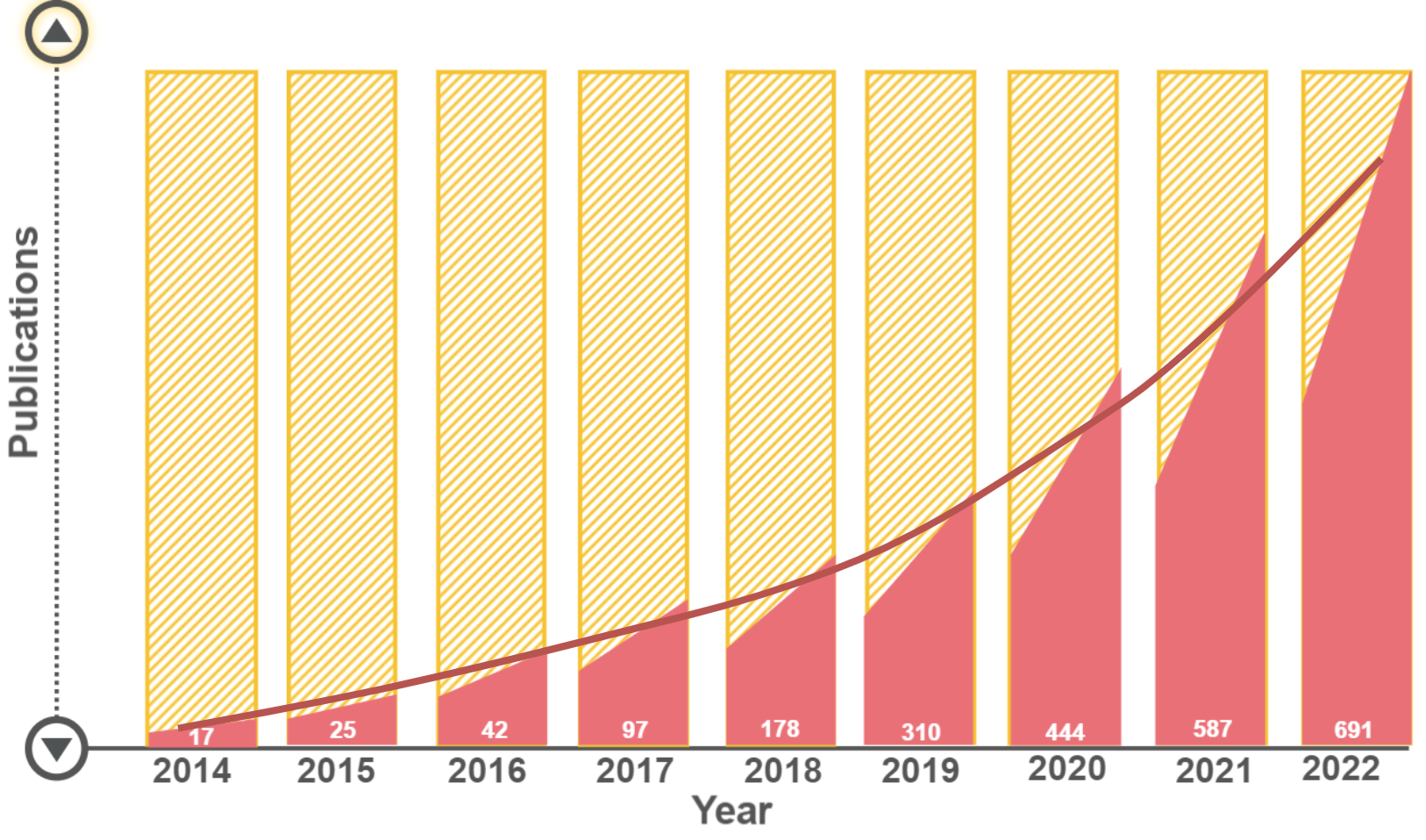}}
\caption{The number of articles in the Web of Science database containing the keywords `End-to-End' and `Autonomous Driving' from 2014 to 2022 illustrates the increasing trend in the research community.}
\label{linegraph}
\end{figure}

The End-to-End approach streamlines the system, improving efficiency and robustness by directly mapping sensory input to control outputs. The benefits of End-to-End autonomous driving have garnered significant attention in the research community as shown in Fig. \ref{linegraph}. Firstly, End-to-End driving addresses the issue of error propagation, as it involves a single learning task pipeline \cite{chitta2021neatc100,wu2022trajectoryc120}  that learns task-specific features, thereby reducing the likelihood of error propagation. Secondly, End-to-End driving offers computational advantages. Modular pipelines often entail redundant computations, as each module is trained for task-specific outputs \cite{kong2015kinematicc51,zhao2012designc50}. This results in unnecessary and prolonged computation. In contrast, End-to-End driving focuses on the specific task of generating the control signal, reducing the need for unnecessary computations and streamlining the overall process. End-to-End models were previously regarded as ``black boxes", lacking transparency. However, recent methodologies have improved interpretability in End-to-End models by generating auxiliary outputs \cite{chitta2022transfuserc116,wu2022trajectoryc120}, attention maps \cite{prakash2021multic121,wu2023policy,shao2023reasonnet,hu2023planningoriented,xiao2023scaling,renz2022plantc121c}, and interpretable maps \cite{renz2022plantc121c,jia2023think,shao2023safetyc117,hu2022st,chitta2021neatc100,li2022human}. This enhanced interpretability provides insights into the root causes of errors and model decision-making. Furthermore, End-to-End driving demonstrates resilience to adversarial attacks. Adversarial attacks \cite{wu2023adversarial} involve manipulating sensor inputs to deceive or confuse autonomous driving systems. In End-to-End models, it is challenging to identify and manipulate the specific driving behavior triggers as it is unknown what causes specific driving patterns. Lastly, End-to-End driving offers ease of training. Modular pipelines require separate training and optimization of each task-driven module, necessitating domain-specific knowledge and expertise. In contrast, End-to-End models can learn relevant features and patterns \cite{codevilla2019exploringc61,hawke2020urban} directly from raw sensor data, reducing the need for extensive engineering and expertise.

\begin{figure*}[H]
\centerline{\includegraphics[scale=0.3]{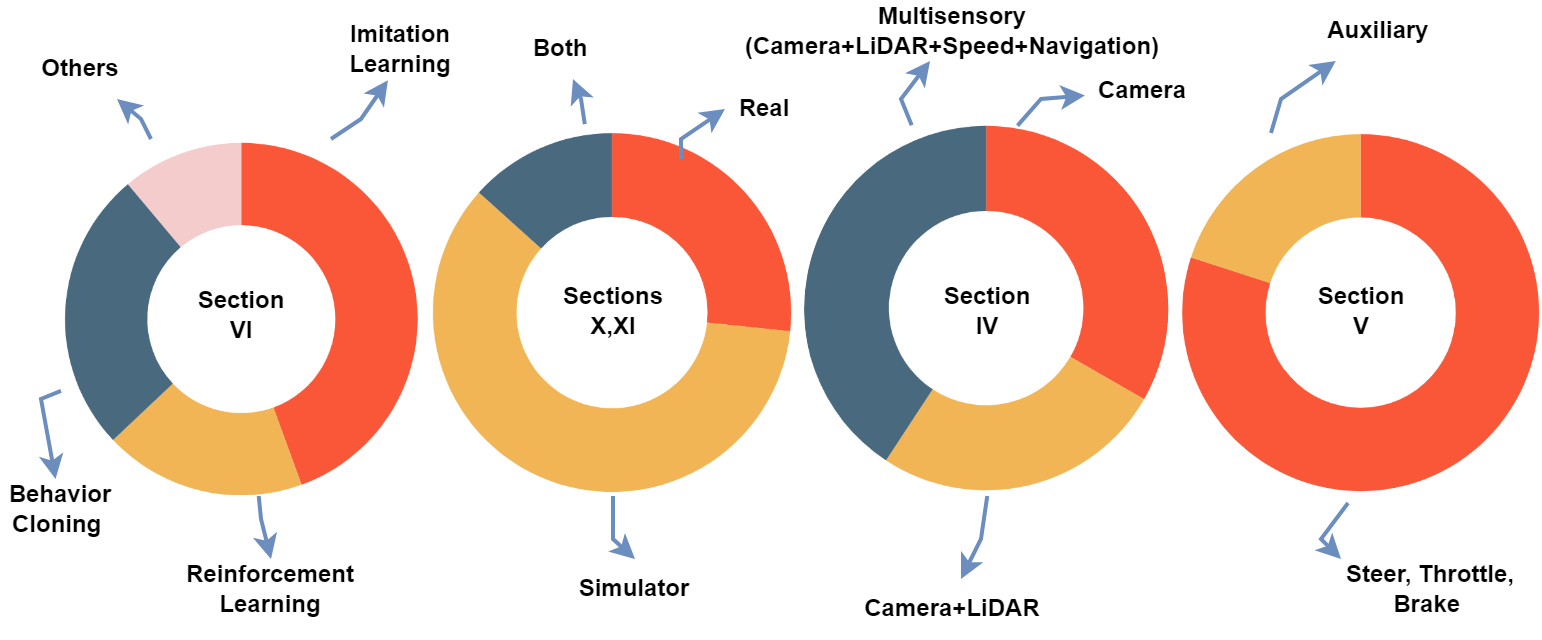}}
\caption{The charts illustrate statistics of the papers included in this survey according to learning approaches (section \ref{learning}), environment being utilized for training (sections \ref{evaluation}, \ref{s_dataset}), input modality (section \ref{input}), and output modality (section \ref{output}).}
\label{piechart}
\end{figure*}

\textbf{Related Surveys}: A number of related surveys are available, though their emphasis differs from ours. The author Yurtsever et al. \cite{yurtsever2020survey} covers the autonomous driving domain with a primary emphasis on the modular methodology. Several past surveys center around specific learning techniques, such as imitation learning \cite{le2022survey} and reinforcement learning \cite{zhu2021survey}. A few end-to-end surveys, including the work by Tampuu et al. \cite{tampuu2020survey}, provide an architectural overview of the complete end-to-end driving pipeline. Recently, Chen et al. \cite{chen2023e2esurvey} discuss the methodology and challenges in end-to-end autonomous driving in their survey. Our focus, however, is on the latest advancements, including modalities, learning principles, safety, explainability, and evaluation (see Table~\ref{literature}).

\textbf{Motivation and Contributions}: The End-to-End architectures have significantly enhanced autonomous driving systems. As elaborated earlier, these architectures have overcome the limitations of modular approaches. Motivated by these developments, we present a survey on recent advancements in End-to-End autonomous driving. The key contributions of this paper are threefold. First, this survey exclusively explores End-to-End autonomous driving using deep learning. We provide a comprehensive analysis of the underlying principles, methodologies, and functionality, delving into the latest state-of-the-art advancements in this domain. Second, we present a detailed investigation in terms of modality, learning, safety, explainability, and results, and provide a quantitative summary in Table~\ref{literature}. Third, we present an evaluation framework based on both open and closed-loop assessments and compile a summarized list of available datasets and simulators.

\textbf{Paper Organization}: The survey is organized as per the underlying principles and methodologies (see Fig. \ref{piechart}). We present the background of modular systems in Section \ref{modularsection}. Section \ref{s_architecture} provides an overview of the End-to-End autonomous driving pipeline architecture. This is followed by sections \ref{input} and \ref{output}, which discuss the input and output modalities of the End-to-End system. Section \ref{learning} comprehensively covers End-to-End learning methods, from imitation learning to reinforcement learning. The domain adaptation is explained in section \ref{domain}. Next, we explore the safety aspect of End-to-End approaches in section \ref{safety}. The importance of explainability and interpretability is discussed in section \ref{explainability}. The evaluation of the End-to-End system consists of open and closed-loop evaluations, which are discussed in section \ref{evaluation}. The relevant datasets and the simulator are presented in section \ref{s_dataset}. Finally, sections \ref{s_future} and \ref{conclusion} provide the future research direction and conclusion, respectively.


\section{Modular system architecture}\label{modularsection}

The modular pipeline \cite{levinson2011towardsc23,urmson2008autonomousc24,wei2013towardsc25,maddern20171c26,akai2017autonomousc27,somerville2018uberc28,ziegler2014makingc29} begins by inputting the raw sensory data into the perception module for obstacle detection and localization via the localization module, followed by planning and prediction for the optimal and safe trajectory of the vehicle. Finally, the motor controller outputs the control signals. The standard modules of the modular driving pipeline are listed below:

\subsection{Components of modular pipeline}

\paragraph{Preception:}

The perception module seeks to achieve a better understanding \cite{wei2013towardsc25} of the scene. It is built on top of algorithms such as object detection and lane detection. The perception module is responsible for sensor fusion, information extraction, and acts as a mediator between the low-level sensor input and the high-level decision module. It fuses heterogeneous sensors to capture and generalize the environment.
The primary tasks of the perception module include:
(i) Object detection
(ii) Object tracking
(iii) Road and lane detection

\begin{figure*}[t]
\centerline{\includegraphics[scale=0.25]{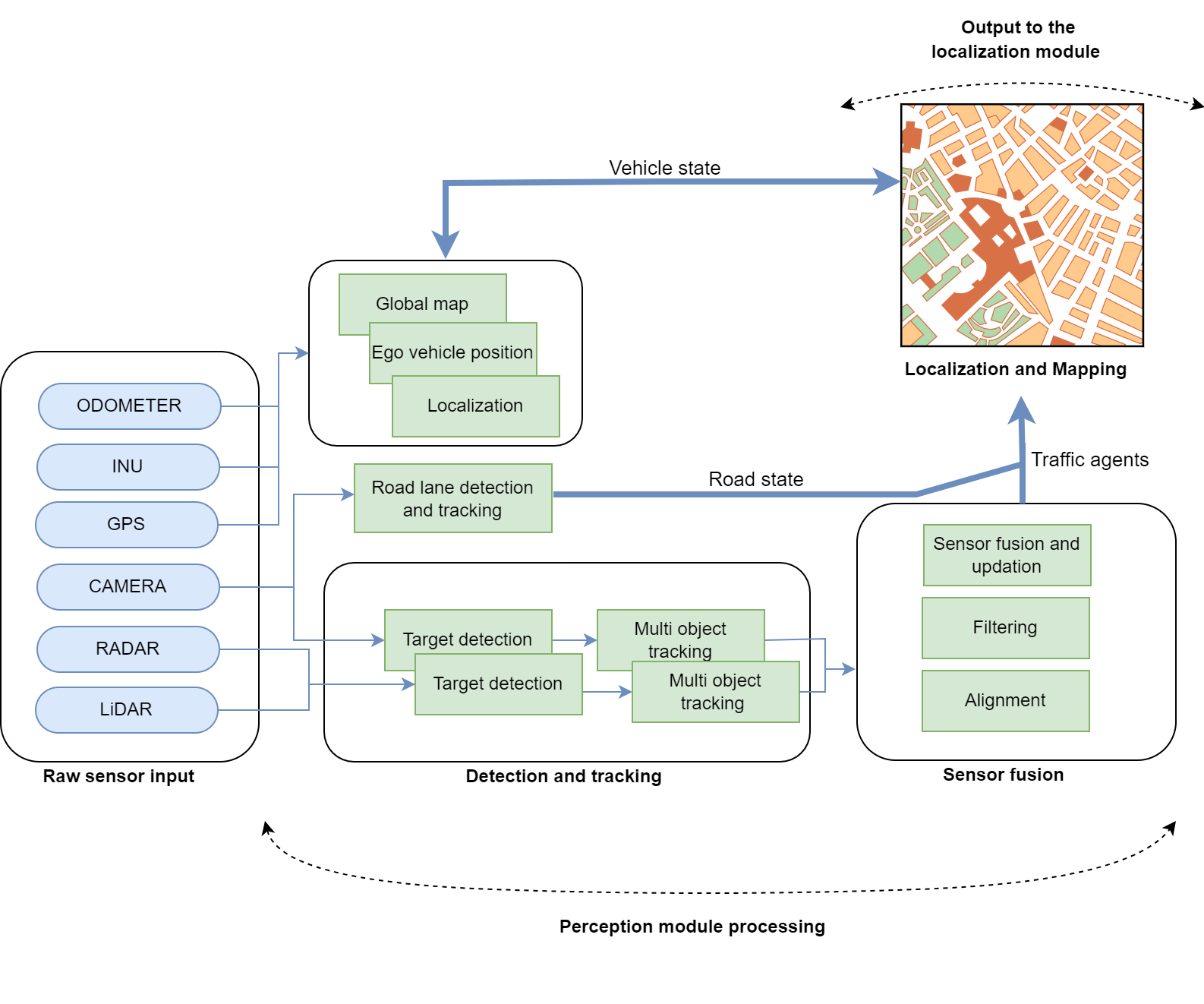}}
\caption{The perception module receives and processes various raw sensor inputs, which are then utilized by the localization and mapping module.}
\label{fig1}
\end{figure*}

\paragraph{Localization and mapping:}

Localization is the next important module, which is responsible for determining the position of the ego vehicle and the corresponding road agents \cite{kuutti2018surveyc30}. It is crucial for accurately positioning the vehicle and enabling safe maneuvers in diverse traffic scenarios. The end product of the localization module is an accurate map. Some of the localization techniques include High Definition map (HD map) and Simultaneous Localization And Mapping (SLAM), which serve as the online map and localize the traffic agents at different time stamps. The localization mapping can further be utilized for driving policy and control commands.

\paragraph{Planning and driving policy:}

The planning and driving policy module \cite{bergamini2021simnet} is responsible for computing a motion-level command that determines the control signal based on the localization map provided by the previous module. It predicts the optimal future trajectory \cite{lin2000vehicle} based on past traffic patterns. The categorization of trajectory prediction techniques (Table \ref{table2}) is as follows:

\begin{itemize}
\item Physics-based methods: These methods are suitable for vehicle motion that can be accurately characterized by kinematics or dynamics models. They can simulate various scenarios quickly with minimal computational cost.
\item Classic machine learning-based methods: Compared to physics-based approaches, this class of methods can consider more variables, provide reasonable accuracy, and have a longer forecasting span, but at a higher computing cost. Most of these techniques use historical motion data to estimate future trajectories.
\item Deep learning-based methods: Deep learning algorithms are capable of reliable prediction over a wider range of prediction horizons. In contrast, standard trajectory prediction methods are only suited for basic scenes and short-term prediction. Deep learning-based systems can make precise predictions across a wider time horizon. As shown in Table \ref{learningapproches}, deep learning utilizes RNN, CNN, GNN, and other networks for feature extraction, calculating interaction strength, and incorporating map information.
\item Reinforcement learning-based methods: These methods aim to mimic how people make decisions and learn the reward function by studying expert demonstrations to produce the best driving strategy. However, most of these techniques are computationally costly.
\end{itemize}

\begin{table*}[t]
\caption{Performance of different driving policy approaches: PB (Physics-Based), CML (Classical Machine Learning), DL (Deep Learning), RL (Reinforcement Learning).}
\label{table2}
\begin{tabular}{cccc}
\hline
Techniques & Accuracy        & Computation cost & Prediction distance \\ \hline
PB         & Medium          & Low              & Short               \\ \hline
CML        & Low             & Medium           & Medium              \\ \hline
DL         & Highly accurate & High             & Wide                \\ \hline
RL         & Highly accurate & High             & Wide                \\ \hline
\end{tabular}
\end{table*}

\begin{table*}[t]
\caption{Summary of deep learning-based approaches for motion prediction utilizing different backbone networks.}
\label{learningapproches}
\begin{tabular}{cccccc}
\hline
Methods                                                             & Detail                                                                                                                                    & Classification                                              & \begin{tabular}[c]{@{}c@{}}Agent and context \\ encoder\end{tabular}  & Context encoder                                                       & Decoder                                                                              \\ \hline
CoverNet {\cite{phan2020covernetc126}}                                                 & \begin{tabular}[c]{@{}c@{}}Formulation of \\ classification problem\\ over the set of diverse trajectories\end{tabular}                      & CNN                                                         & CNN                                                                   & CNN                                                                   & \begin{tabular}[c]{@{}c@{}}Trajectory set \\ generator\end{tabular}                  \\ \hline
HOME {\cite{gilles2021homec127}}                                                     & \begin{tabular}[c]{@{}c@{}}It outputs the \\ 2D top view representation \\ of agent possible future\end{tabular}                             & CNN                                                         & CNN,GRU                                                               & CNN                                                                   & CNN                                                                                  \\ \hline
TPCN {\cite{Ye_2021_CVPRc128}}                                                     & \begin{tabular}[c]{@{}c@{}}Splitting of trajectory prediction\\ into both temporal and \\ spatial dimension\end{tabular}                     & CNN                                                         & PointNET ++                                                           & PointNET++                                                            & \begin{tabular}[c]{@{}c@{}}Displacement \\ prediction\end{tabular}                   \\ \hline
\begin{tabular}[c]{@{}c@{}}MHA-JAM \\ {\cite{ye2021tpcnc129}}\end{tabular}       & \begin{tabular}[c]{@{}c@{}}Attention head is used \\ to generate distinct \\ future trajectories \\ while addressing multimodality\end{tabular} & Attention                                                   & LSTM                                                                  & CNN                                                                   & LSTM                                                                                 \\ \hline
\begin{tabular}[c]{@{}c@{}}MMTransformer\\  {\cite{liu2021multimodalc130}}\end{tabular} & \begin{tabular}[c]{@{}c@{}}Network architecture \\ based on stacked \\ transformer to model the \\ feature multimodality\end{tabular}           & Attention                                                   & Transformer                                                           & VectorNet                                                             & MLP                                                                                  \\ \hline
DenseTNT {\cite{gu2021densetntc131}}                                                 & \begin{tabular}[c]{@{}c@{}}It is a anchor-free model which\\  directly outputs from \\ the dense goal candidates\end{tabular}                & GNN                                                         & VectorNet                                                             & VectorNet                                                             & \begin{tabular}[c]{@{}c@{}}Goal Bases \\ multi-trajectory\\  prediction\end{tabular} \\ \hline
TS-GAN {\cite{song2022learningc132c133}}                                                   & \begin{tabular}[c]{@{}c@{}}Collaborative learning \\ and GAN for modeling\\ motion behavior\end{tabular}                                     & \begin{tabular}[c]{@{}c@{}}Generative \\ model\end{tabular} & LSTM                                                                  & -                                                                     & LSTM                                                                                 \\ \hline
PRIME {\cite{song2022learningc132c133}}                                                    & \begin{tabular}[c]{@{}c@{}} Utilizes the model generator and\\ learning based evaluator\end{tabular}                                 & \begin{tabular}[c]{@{}c@{}}Generative \\ model\end{tabular} & CNN,LSTM                                                              & LSTM                                                                  & \begin{tabular}[c]{@{}c@{}}Model-based \\ generator\end{tabular}                     \\ \hline
\begin{tabular}[c]{@{}c@{}}MotionDiff\\  {\cite{wei2022humanc134}}\end{tabular}    & \begin{tabular}[c]{@{}c@{}}Diffusion probability based \\ kinematics model to diffuse \\ original states to \\ noise distribution\end{tabular}        & CNN                                                         & \begin{tabular}[c]{@{}c@{}}Spatial \\ transformer,\\ GRU\end{tabular} & \begin{tabular}[c]{@{}c@{}}Spatial\\  transformer,\\ GRU\end{tabular} & MLP                                                                                  \\ \hline
ScePT {\cite{chen2022learning5orc98}}                                                    & \begin{tabular}[c]{@{}c@{}}A policy planning \\ based trajectory prediction for \\ accurate motion planning.\end{tabular}                    & GNN                                                         & LSTM                                                                  & CNN                                                                   & GRU, PonitNET                                                                        \\ \hline
\end{tabular}
\end{table*}

\paragraph{Control:}

The motion planner generates the trajectory, which is then updated by the obstacle subsystem and sent to the controller subsystem.
The computed command is sent to the actuators of the driving components, including the throttle, brakes, and steering, to follow the desired trajectory, which is optimized and safer in real-world scenarios. The Proportional Integral Derivative (PID) \cite{zhao2012designc50} and Model Predictive Control (MPC) \cite{kong2015kinematicc51} are some of the controllers used to generate the aforementioned control signals.

\subsection{Input and output modality of modular pipeline}
The output modality of each module is designed to be compatible with the input modality of subsequent modules in the pipeline to ensure that information is correctly propagated through the modular system.

\paragraph{Sensory data:}

At this level (Fig. \ref{fig1}), the raw data from the embedded multi-sensor array is retrieved, filtered, and processed for semantic mapping. LiDAR, RADAR, Camera, GPS, and Odometer are some of the sensor inputs to the perception stack. LiDAR and RADAR are used for depth analysis, while cameras are employed for detection. The INU, GPS, and Odometer sensors capture and map the vehicle's position, state, and the corresponding environment, which can be further utilized by decision-level stages.

\paragraph{Input to the mapping and localization:}

Localization aims to estimate the vehicle's position at each time stamp. Utilizing information from the perception module, the vehicle's position and the environment are mapped based on parameters such as position, orientation, pose, speed, and acceleration. Localization techniques \cite{akai2017autonomousc27} allow for the integration of multiple objects and the identification of their relationships, resulting in a more comprehensive, augmented, and enriched representation.

\begin{figure}[htbp]
\centerline{\includegraphics[scale=0.33]{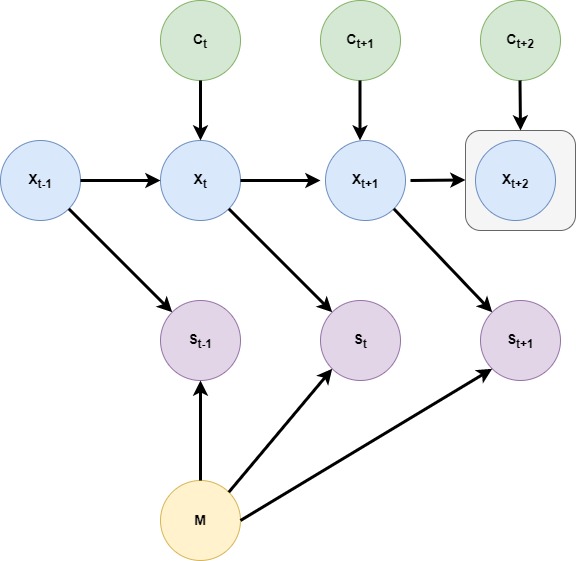}}
\caption{Visualization of temporal correlations from localization that can be used to identify specific behaviors and predict future positions.}
\label{fig}
\end{figure}

As shown in Fig. \ref{fig}, we define $X_t$ as the vehicle's position estimate at time $t$, and $M$ as the environment map. These variables can be estimated using control inputs $C_t$, which are typically derived from wheel encoders or sensors capable of estimating the vehicle's displacement. The measurements derived from sensor readings are denoted by $S_t$ and are used to aid in the estimation of the vehicle's pose.

\paragraph{Input to the path planning and decision module:}

Path planning is broadly categorized into local and global path planners. The purpose of the local planner is to execute the goals set by the global path planner. It is responsible for finding trajectories that avoid obstacles and satisfy optimization requirements within the vehicle's operational space.

The problem of local trajectory prediction can be formulated as estimating the future states ($t_f$) of various traffic actors ($R^t$) in a given scenario based on their current and past states ($t_h$). The state of traffic actors includes vehicles or pedestrians with historical trajectories at different time stamps.

\begin{equation}
    Input = \{ R^1,R^2,R^3,R^3.\;.\;.\;.\;.\;.,R^{t_{h}} \}
\end{equation}

where $R^t$ contains the coordinates of different traffic actors at each time stamp $t$ (up to $h$ past time stamps).
\begin{equation}
    R^t = \{x_0^t,y_0^t,x_1^t,y_1^t \;.\; .\;.\; x_n^t,y_n^t\}
\end{equation}
where $n$ represents all traffic vehicles detected by the ego vehicle; ($x_i^t,y_i^t$) are the coordinates of the vehicle at the $t$ time stamp. $X$ is the input to the path planning module, and the vehicle trajectory $Y$ is predicted from the model at future time stamp $t_f$.
\begin{equation}
    Y = \left \{ R^{t_{h}+1}, R^{t_{h}+2},R^{t_{h}+3}\cdot \cdot \cdot R^{t_{h}+t_{f}}  \right \}
\end{equation}

\paragraph{Input to the control module:}

There are two primary forms of trajectory command that the controller receives: (i) as a series of commands ($T_c$) and (ii) as a series of states ($T_s$).
Controller subsystems that receive a $T_s$ trajectory may be categorized as path tracking techniques, while those that receive a $T_c$ trajectory can be classified as direct hardware actuation control methods.

\begin{itemize}
\item Direct hardware actuation control methods: The Proportional Integral Derivative (PID) \cite{zhao2012designc50} control system is a commonly used hardware actuation technique for self-driving automobiles. It involves determining a desired hardware input and an error measure gauging how much the output deviates from the desired outcome.
\item Path tracking methods: Model Predictive Control (MPC) \cite{kong2015kinematicc51} is a path-tracking approach that involves choosing control command inputs that will result in desirable hardware outputs, then simulating and optimizing those outputs using the motion model of the car over a future prediction horizon.
\end{itemize}

\section{End-to-End system architecture} \label{s_architecture}
 
\begin{figure*}[t]
\centerline{\includegraphics[scale=0.22]{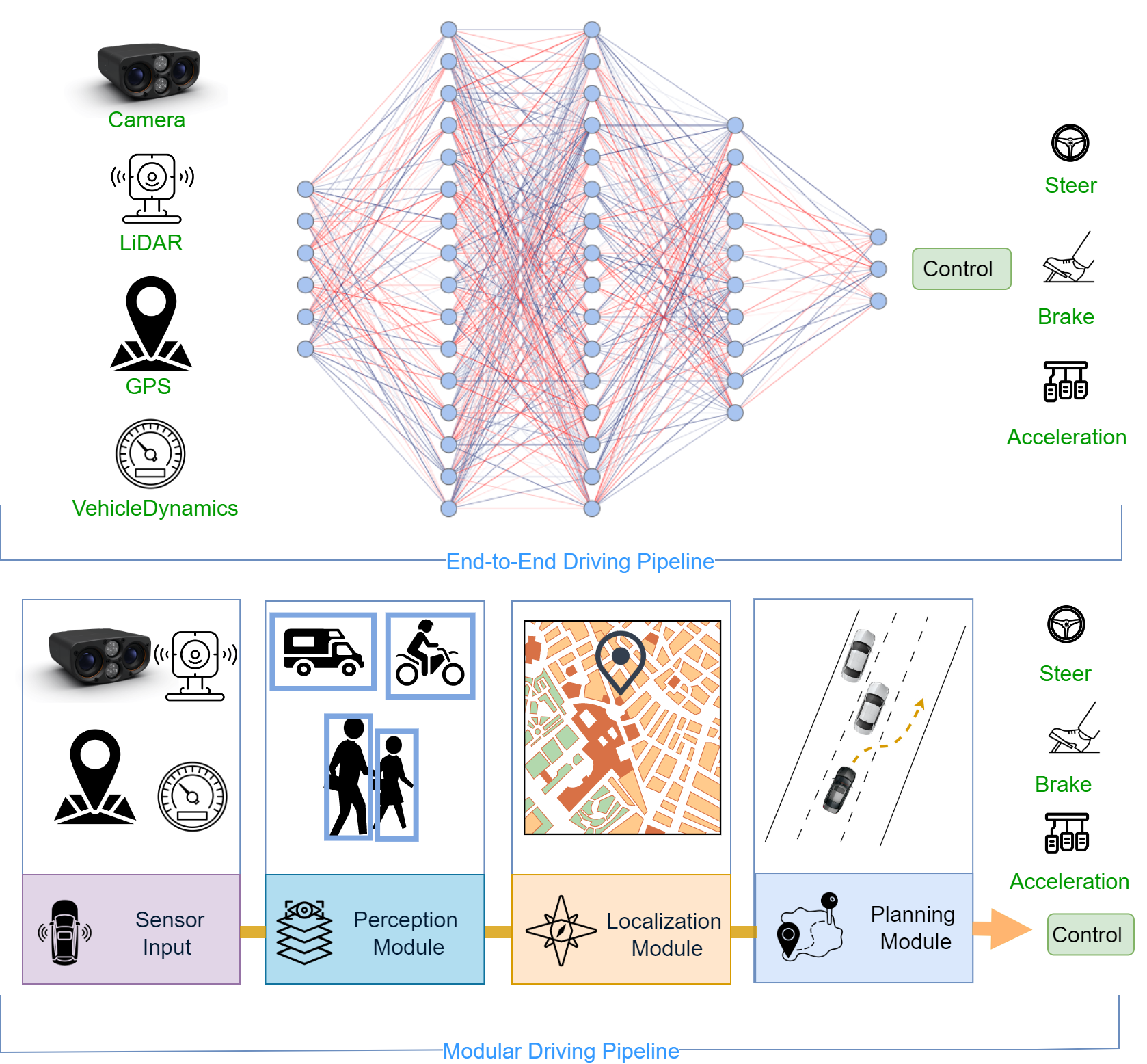}}
\caption{Comparison between End-to-End and modular pipelines. End-to-End is a single pipeline that generates the control signal directly from perception input, whereas a modular pipeline consists of various sub-modules, each with task-specific functionalities.}
\label{architecture}
\end{figure*}

In general, modular systems are referred to as the mediated paradigm and are constructed as a pipeline of discrete components (Fig. \ref{architecture}) that connect sensory inputs and motor outputs. The core processes of a modular system include perception, localization, mapping, planning, and vehicle control \cite{chen2015deepdrivingc22}. The modular pipeline starts by inputting raw sensory data to the perception module for obstacle detection \cite{patil2019h3d} and localization via the localization module \cite{akai2017autonomousc27}. This is followed by planning and prediction \cite{song2022learningc132c133} to determine the optimal and safe trajectory for the vehicle. Finally, the motor controller generates commands for safe maneuvering.

On the other hand, direct perception or End-to-End driving directly generates ego-motion from the sensory input. It optimizes the driving pipeline (Fig. \ref{architecture}) by bypassing the sub-tasks related to perception and planning, allowing for continuous learning to sense and act, similar to humans. The first attempt at End-to-End driving was made by Pomerleau Alvinn \cite{pomerleau1988alvinnc1}, which trained a 3-layer sensorimotor fully connected network to output the car's direction. End-to-End driving generates ego-motion based on sensory input, which can be of various modalities. However, the prominent ones are the camera \cite{chen2020learningc60,ohn2020learningc62,zhao2021samc65}, Light Detection and Ranging (LiDAR) \cite{hanselmann2022king2orc99,chen2022learning5orc98,chitta2022transfuserc116}, navigation commands \cite{zhang2021learning,ohn2020learningc62,codevilla2019exploringc61}, and vehicle dynamics, such as speed \cite{chen2021learningc136,toromanoff2020endc66,zhao2021samc65}. This sensory information is utilized as the input to the backbone model, which is responsible for generating control signals. Ego-motion can involve different types of motions, such as acceleration, turning, steering, and pedaling. Additionally, many models also output additional information, such as a cost map for safe maneuvers, interpretable outputs, and other auxiliary outputs.

There are two main approaches for End-to-End driving: either the driving model is explored and improved via Reinforcement Learning (RL) \cite{zhang2021endc94,toromanoff2020endc66,li2022efficient4,peng2022safe,li2022human,zhao2022cadre}, or it is trained in a supervised manner using Imitation Learning (IL) \cite{renz2022plantc121c,hanselmann2022king2orc99,jia2023think,wu2023policy,xiao2023scaling,zhang2023coaching,chitta2022transfuserc116} to resemble human driving behavior. The supervised learning paradigm aims to learn the driving style from expert demonstrations, which serve as training examples for the model. However, expanding an autonomous driving system based on IL \cite{codevilla2019exploringc61} is challenging since it is impossible to cover every instance during the learning phase. On the other hand, RL works by maximizing cumulative rewards \cite{8793742,li2022efficient4} over time through interaction with the environment, and the network makes driving decisions to obtain rewards or penalties based on its actions. While RL model training occurs online and allows exploration of the environment during training, it is less effective in utilizing data compared to imitation learning. Table \ref{literature} summarizes recent methods in End-to-End driving.

\section{Input modalities in End-to-End system}\label{input}
The following section explores the input modalities essential for end-to-end autonomous driving. These encompass cameras for visual insights, LiDAR for precise 3D point clouds, multi-modal inputs, and navigational inputs. Fig. \ref{inputoutput} illustrates some of the input and output modalities.

\subsection{Camera}
Camera-based methods \cite{prakash2021multic121,hu2023planningoriented,hanselmann2022king2orc99,codevilla2019exploringc61,ohn2020learningc62,prakash2020exploringc64,zhao2021samc65,toromanoff2020endc66,chitta2021neatc100,wu2022trajectoryc120} have shown promising results in End-to-End driving. For instance, Toromanoff et al. \cite{toromanoff2020endc66} demonstrated their capabilities by winning the CARLA 2019 autonomous driving challenge using vision-based approaches in an urban context. The use of monocular \cite{wu2022trajectoryc120,zhang2022learning,ishihara2021multi,toromanoff2020endc66} and stereo vision \cite{zhang2023coaching,xiao2023scaling,wu2023policy} camera views is a natural input modality for image-to-control End-to-End driving. Xiao et al. \cite{xiao2020multimodalc118} employed inputs consisting of a monocular RGB image from a forward-facing camera and the vehicle speed. Chen et al. LAV \cite{chen2022learning5orc98} utilize only the camera image input as shown in Fig. \ref{inputoutput}(d). Wu et al. \cite{wu2023policy}, Xiao et al. \cite{xiao2023scaling}, Zhang et al. \cite{zhang2023coaching} utilize camera-only modality to generate high-level instructions for lane following, turning, stopping and going straight using  imitation learning.

\subsection{LiDAR}

Another significant input source in self-driving is the Light Detection and Ranging (LiDAR) sensor. LiDAR \cite{Hu_2021_CVPR,Cai2021CarlLeadLE,Zeng2020DSDNetDS,hu2022st} is resistant to lighting conditions and offers accurate distance estimates. Compared to other perception sensors, LiDAR data is the richest and provides the most comprehensive spatial information. It utilizes laser light to detect distances and generates PointClouds, which are 3D representations of space where each point includes the $(x, y, z)$ coordinates of the surface that reflected the sensor's laser beam. When localizing a vehicle, generating odometry measurements is critical. Many techniques utilize LiDAR for feature mapping in Birds Eye View (BEV) \cite{hu2023planningoriented,jia2023think,shao2023reasonnet}, High Definition (HD) map \cite{hu2022st,zhang2022mmfn}, and Simultaneous Localization and Mapping (SLAM) \cite{bailey2006simultaneousc37}. Shenoi et al. \cite{shenoi2020jrmotc55} have shown that adding depth and semantics via LiDAR has the potential to enhance driving performance. Liang et al. \cite{liang2019multic76,liang2018deepc77} utilized point flow to learn the driving policy in an End-to-End manner.

\subsection{Multi-modal}

Multimodality \cite{shao2023safetyc117,renz2022plantc121c,chen2022learning5orc98,shao2023reasonnet,Jaeger2023ICCV} outperforms single modality in crucial perception tasks and is particularly well-suited for autonomous driving applications, as it combines multi-sensor data. There are three broad categorizations for utilizing information depending on when to combine multi-sensor information. In early fusion, sensor data is combined before feeding them into the learnable End-to-End system. Chen et al. \cite{chen2022learning5orc98} as shown in Fig. \ref{inputoutput}(d) use a network that accepts (RGB + Depth) channel inputs, Xiao et al. \cite{xiao2020multimodalc118} model also input the same modality. The network modifies just the first convolutional layer to account for the additional input channel, while the remaining network remains unchanged. Renz et al. \cite{renz2022plantc121c} fuse object-level input representation using a transformer encoder.  The author combinedly represents a set of objects as vehicles and segments of the routes.

In mid-fusion, information fusion is done either after some preprocessing stages or after some feature extraction. Zhou et al. \cite{zhou2020endc72} perform information fusion at the mid-level by leveraging the complementary information provided by both the bird's-eye view (BEV) and perspective views of the LiDAR point cloud. Transfuser \cite{chitta2022transfuserc116} as shown in Fig. \ref{inputoutput}(a) addresses the integration of image and LiDAR modalities using self-attention layers. They utilized multiple transformer modules at multiple resolutions to fuse intermediate features. Obtained feature vector forms a concise representation which an MLP then processes before passing to an auto-regressive waypoint prediction network. In late fusion, inputs are processed separately, and their output is fused and further processed by another layer. Some authors \cite{sobh2018endc79,liang2019multic76,liang2018deepc77,xiao2020multimodalc118} use a late fusion architecture for LiDAR and visual modalities, in which each input stream is encoded separately and concatenated.

\subsection{Navigational inputs}

End-to-End navigation input can originate from the route planner \cite{shao2023safetyc117,prakash2021multic121,zhou2019doesc69c98} and navigation commands \cite{chen2020learningc60,hubschneider2017addingc122,codevilla2018endc123c124,liang2018cirlc125}. Routes are defined by a sequence of discrete endpoint locations in Global Positioning System (GPS) coordinates provided by a global planner \cite{prakash2021multic121}. The TCP model \cite{wu2022trajectoryc120} as illustrated in Fig. \ref{inputoutput}(c) is provided with correlated navigation directives like lane keeping, left/right turns, and the destination. This information is used to produce the control actions.
 
Shao et al. \cite{shao2023safetyc117} propose a technique that guides driving using these sparse destination locations instead of explicitly defining discrete navigation directives. PlanT \cite{renz2022plantc121c} utilizes point-to-point navigation based on the input of the goal location. FlowDriveNet \cite{wang2021flowdrivenetc119} considers both the global planner's discrete navigation command and the coordinates of the navigation target. Hubschneider et al. \cite{hubschneider2017addingc122} include a turn indicator command in the driving model, while Codevilla et al. \cite{codevilla2018endc123c124} utilize a CNN block for specific navigation tasks and a second block for subset navigation. In addition to the aforementioned inputs, End-to-End models also incorporate vehicle dynamics, such as ego-vehicle speed \cite{chen2021learningc136,ohn2020learningc62,toromanoff2020endc66,chitta2021neatc100}.

\section{Output modalities in End-to-End system}\label{output}

Usually, an End-to-End autonomous driving system outputs control commands, waypoints, or trajectories. In addition, it may also produce additional representations, such as a cost map and auxiliary outputs.

\subsection{Waypoints}
Predicting future waypoints is a higher-level output modality. Several authors \cite{chen2022learning5orc98,hanselmann2022king2orc99,chitta2022transfuserc116,fong2019understandingc110} use an auto-regressive waypoint network to predict differential waypoints. Trajectories \cite{shao2023safetyc117,zhou2019doesc69c98,cui2021lookout,wu2022trajectoryc120,jia2023think} can also represent sequences of waypoints in the coordinate frame. The network's output waypoints are converted into low-level steering and acceleration using Model Predictive Control (MPC) \cite{kong2015kinematicc51} and Proportional Integral Derivative (PID) \cite{zhao2012designc50}. The longitudinal controller considers the magnitude of a weighted average of vectors between successive time-step waypoints, while the lateral controller considers their direction. The ideal waypoint \cite{toromanoff2020endc66} relies on desired speed, position, and rotation. 

\clearpage
\onecolumn
\begin{figure*}
    \centering
    \begin{tabular}{@{}c@{}}
    \includegraphics[width=.8\linewidth,height=90pt]{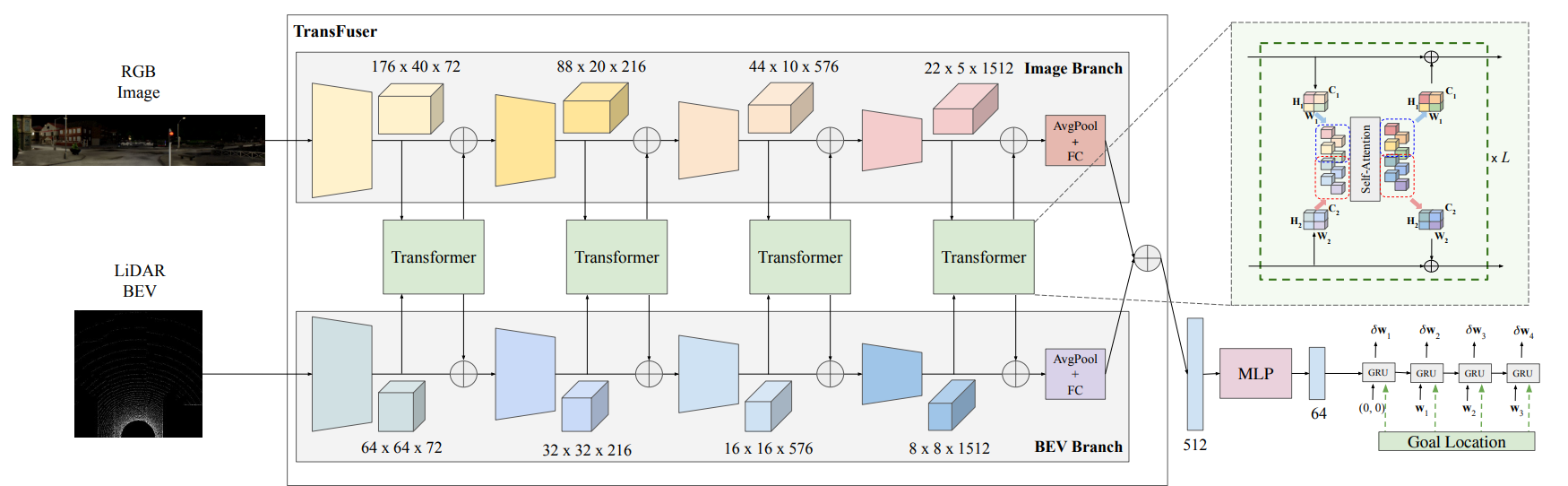} \\[\abovecaptionskip]
    \small (a) Fusion Transformer
    \label{figg1}
    \end{tabular}
     \begin{tabular}{@{}c@{}}
    \includegraphics[width=.8\linewidth,height=100pt]{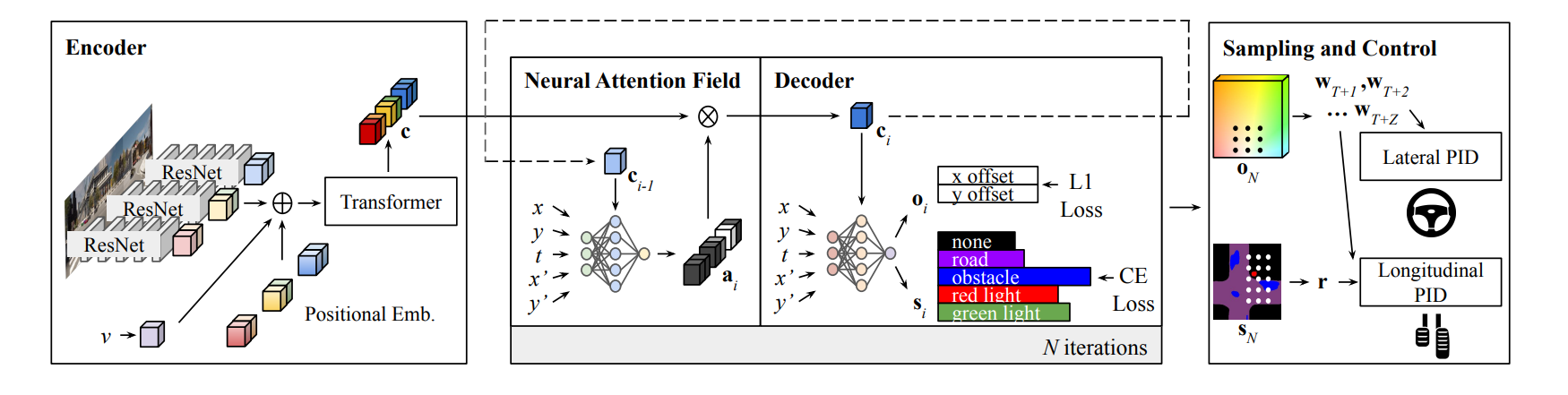} \\[\abovecaptionskip]
    \small b) NEAT
  \end{tabular}
   \begin{tabular}{@{}c@{}}
    \includegraphics[width=.4\linewidth,height=100pt]{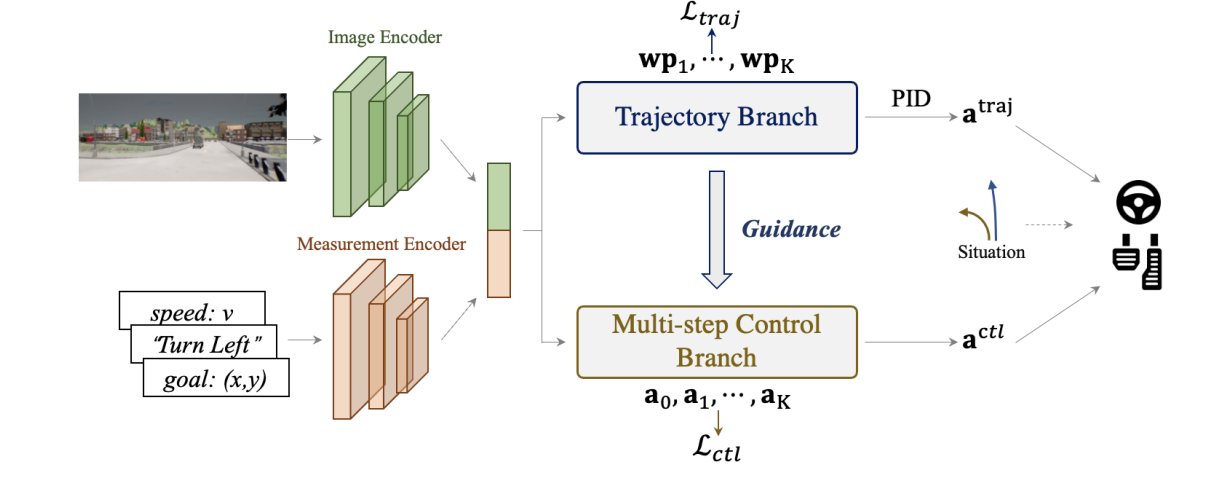} \\[\abovecaptionskip]
    \small (c) TCP
  \end{tabular}
   \begin{tabular}{@{}c@{}}
    \includegraphics[width=.4\linewidth,height=120pt]{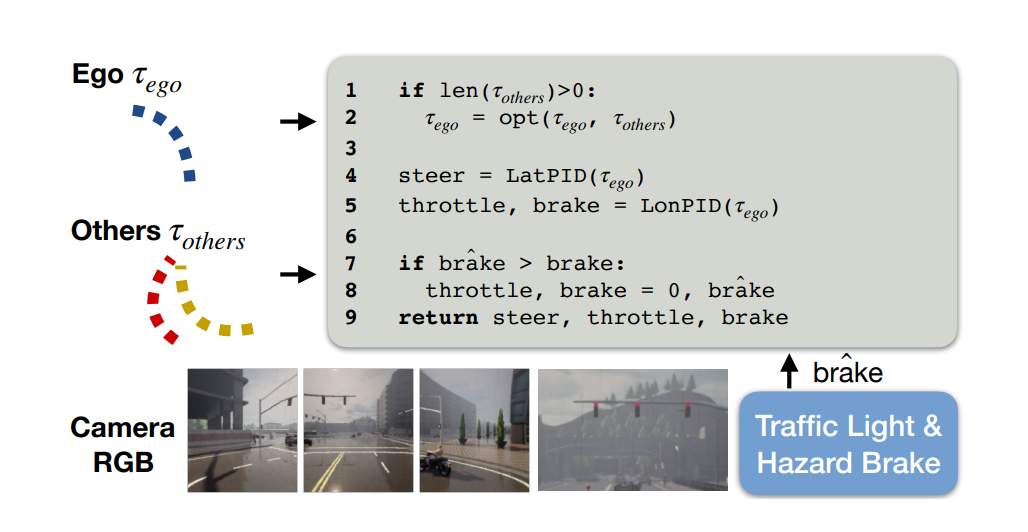} \\[\abovecaptionskip]
    \small (d) LAV
  \end{tabular}
  \begin{tabular}{@{}c@{}}
    \includegraphics[width=.4\linewidth,height=120pt]{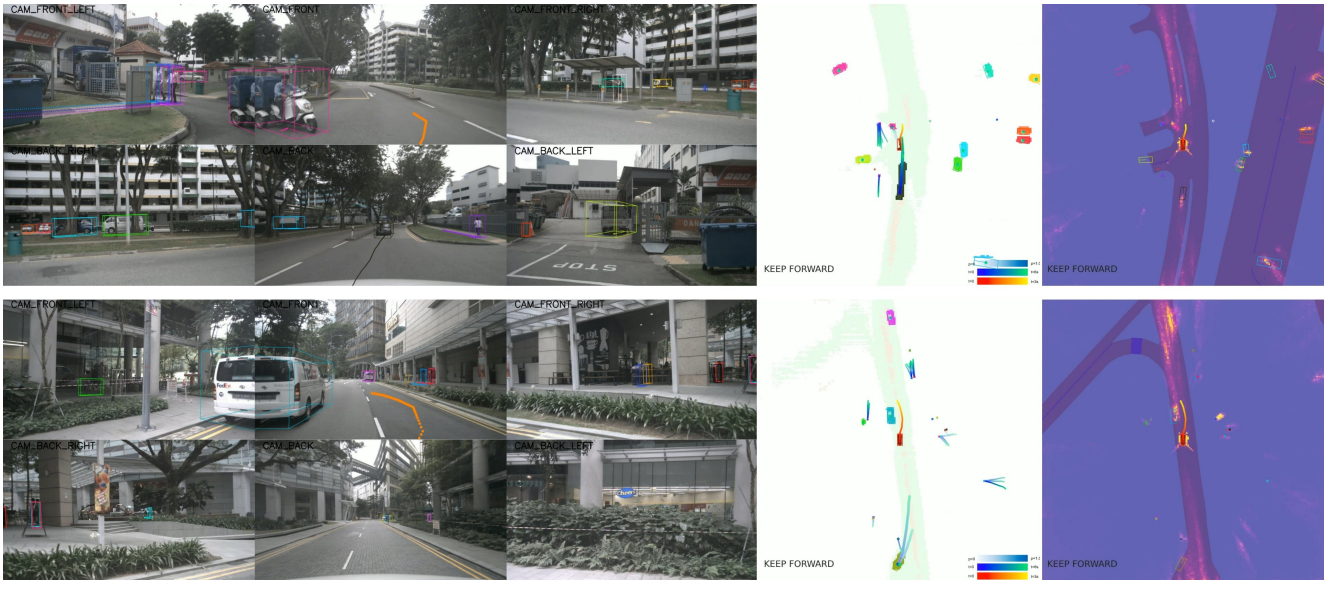} \\[\abovecaptionskip]
    \small (e) UniAD
  \end{tabular} \label{inputoutpute}
  \begin{tabular}{@{}c@{}}
    \includegraphics[width=.4\linewidth,height=120pt]{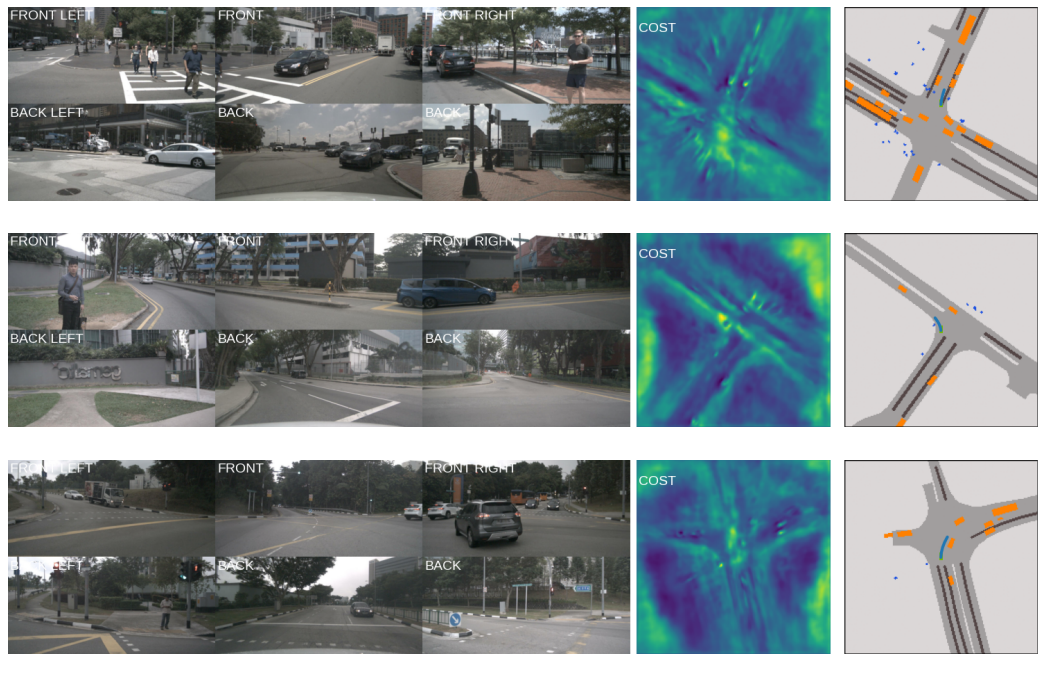} \\[\abovecaptionskip]
    \small (f) ST-P3
  \end{tabular} \label{inputoutputf}
    \caption{The input-output representation of various End-to-End models: (a) Considered RGB image and LiDAR BEV representations as inputs to the multi-modal fusion transformer \cite{chitta2022transfuserc116} and predicts the differential ego-vehicle waypoints. (b) NEAT \cite{chitta2021neatc100} inputs the image patch and velocity features to obtain a waypoint for each time-step used by PID controllers for driving. (c) TCP \cite{wu2022trajectoryc120} takes input image i, navigation information g, current speed v, to generate the control actions guided by the trajectory branch and control branch. (d) LAV \cite{chen2022learning5orc98} uses an image-only input and predicts multi-modal future trajectories used for braking and handling traffic signs and obstacles. (e) UniAD \cite{hu2023planningoriented} generates attention mask visualization which shows how much attention is paid to the goal lane as well as the critical agents that are yielding to the ego-vehicle.(f) ST-P3 \cite{hu2022st} outputs the sub cost map from the prediction module (darker color indicates a smaller cost value). By incorporating the occupancy probability field and leveraging pre-existing knowledge, the cost function effectively balances safety considerations for the final trajectory.}
    \label{inputoutput}
\end{figure*}
\clearpage
\twocolumn

 The lateral distance and angle must be minimized to maximize the reward (or minimize the deviation). The benefit of utilizing waypoints as an output is that they are not affected by vehicle geometry. Additionally, waypoints are easier to analyze by the controller for control commands such as steering. Waypoints in continuous form can be transformed into a specific trajectory. Zhang et al. \cite{zhang2021endc94} and Zhou et al. \cite{zhou2019doesc69c98} utilize a motion planner to generate a series of waypoints that describe the future trajectory. LAV \cite{chen2022learning5orc98}  predicts multi-modal future trajectories (Fig. \ref{inputoutput}(d)) for all detected vehicles, including the ego-vehicle. They use future waypoints to represent the motion plan.

\subsection{Cost function}
Many trajectories and waypoints are possible for the safe maneuvering of the vehicle. The cost \cite{hu2022st,zeng2019end,peng2022safe,li2022human,cui2021lookout,casas2021mp3} is used to select the optimal one among the possibilities. It assigns a weight (positive or negative score) to each trajectory based on parameters defined by the end user, such as safety, distance traveled, comfort, and others. Rhinehart et al. \cite{rhinehart2018deepc67c92} and Chen et al. \cite{chen2022learning5orc98} refine control using the predictive consistency map, which updates knowledge at test time. They also evaluate the trajectory using an ensemble expert likelihood model. Prakash et al. \cite{prakash2021multic121} utilize object-level representations to analyze collision-free routes. Zeng et al. \cite{zeng2021endtoend} employ a neural motion planner that uses a cost volume to predict future trajectories. Hu et al. \cite{hu2022st} employ a cost function illustrated in Fig. \ref{inputoutput}(f) that takes advantage of the learned occupancy probability field, represented by segmentation maps, and prior knowledge such as traffic rules to select the trajectory with the minimum cost. Regarding safety cost functions, Zhao et al. \cite{zhao2021samc65}, Chen et al. \cite{chen2017multic71}, and Shao et al. \cite{shao2023safetyc117} employ safety maps. They analyze actions within the safe set to create causal insights regarding hazardous driving situations.

\begin{figure*}[H]
  \centering
  \begin{tabular}{@{}c@{}}
    \includegraphics[width=.9\linewidth]{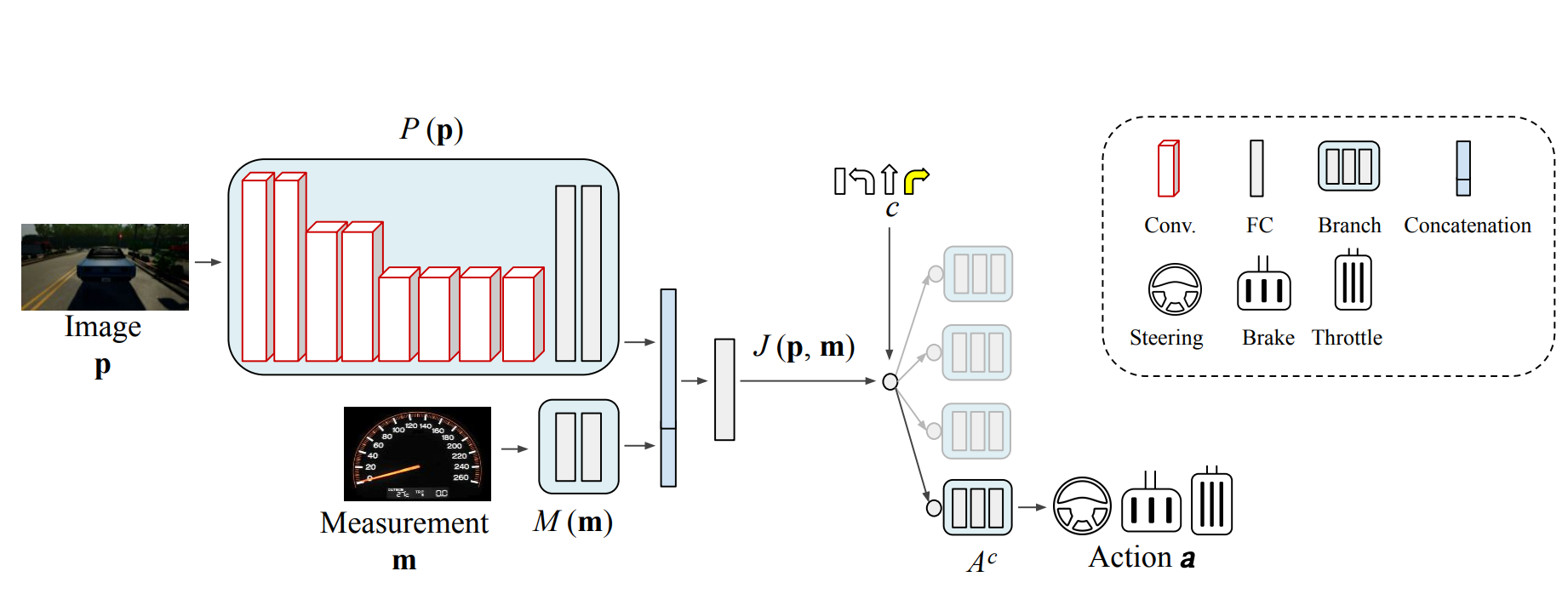} \\[\abovecaptionskip]
    \small
  \end{tabular}

  \vspace{\floatsep}

  \caption{Vehicle maneuvers, represented by a triplet of steering angle, throttle, and brake, depend on a high-level route navigation command (e.g., turn-left, turn-right, go-straight, continue), as well as perception data (e.g., RGB image) and vehicle state measurements (e.g., speed). These inputs guide the specific actions taken by the vehicle, enabling it to navigate the environment effectively through conditional imitation learning \cite{xiao2020multimodalc118}.}\label{f_imitation}
  
\end{figure*}

\subsection{Direct control and acceleration}

Most of the End-to-End models \cite{rhinehart2018deepc67c92,zhou2019doesc69c98,toromanoff2020endc66,chen2020learningc60,codevilla2019exploringc61,hubschneider2017addingc122,codevilla2018endc123c124,liang2018cirlc125,chen2020learningc60} provide the steering angle and speed as outputs at a specific timestamp. The output control needs to be calibrated based on the vehicle's dynamics, determining the appropriate steering angle for turning and the necessary braking for stopping at a measurable distance.

\subsection{Auxiliary output}

The auxiliary output can provide additional information for the model's operation and the determination of driving actions. Several types of auxiliary outputs include the segmentation map \cite{hu2023planningoriented,chitta2022transfuserc116} (Fig. \ref{inputoutput}(e)), BEV map \cite{chitta2021neatc100,hu2023planningoriented,jia2023think,shao2023reasonnet}, future occupancy \cite{renz2022plantc121c,hu2023planningoriented,zeng2021endtoend,casas2021mp3} of the vehicle (Fig. \ref{inputoutput}(e)), and interpretable feature map \cite{renz2022plantc121c,shao2023safetyc117,chitta2022transfuserc116,hu2022st,chitta2021neatc100,zeng2019end} (Fig. \ref{inputoutput}(b)(f)). These outputs provide additional functionality to the End-to-End pipeline and help the model learn better representations. The auxiliary output also facilitates the explanation of the model's behavior \cite{chitta2022transfuserc116,casas2021mp3}, as one can comprehend the information and infer the reasons behind the model's decisions.

\section{Learning approaches for End-to-End system}\label{learning}

The following sections discuss various learning approaches in End-to-End Driving, including imitation learning and reinforcement learning.

\subsection{Imitation learning}\label{AAI}

Imitation learning (IL) \cite{chen2022learning5orc98,renz2022plantc121c,wu2022trajectoryc120,prakash2021multic121,kim2020multi} is based on the principle of learning from expert demonstrations. These demonstrations train the system to mimic the expert's behavior in various driving scenarios. Large-scale expert driving datasets are readily available, which can be leveraged by imitation learning \cite{xiao2020multimodalc118} to train models that perform at human-like standards. 

The main objective is to train a policy \(\pi_{\theta}(s)\) that maps each given state to a corresponding action (Fig. \ref{f_imitation}) as closely as possible to the given expert policy \(\pi^*\), given an expert dataset with state action pair \begin{math}
\left ( s,a \right )
\end{math}:
\begin{equation}
\arg\min_{\theta}E_{s}\sim _{P(s\mid \theta)}L(\pi^*(s),\pi_{\theta}(s))
\end{equation}

where \(P(s\mid\theta)\) represents the state distribution of the trained policy \(\pi_{\theta}\)\label{eq1}.

Behavioural Cloning (BC), Direct Policy Learning (DPL), and Inverse Reinforcement Learning (IRL) are extensions of imitation learning in the domain of autonomous driving.

\subsubsection{Behavioural cloning}

Behavioural cloning \cite{bain1995frameworkc2,chitta2022transfuserc116,wu2022trajectoryc120,chitta2021neatc100,ohn2020learningc62,codevilla2019exploringc61,cui2022coopernaut} is the supervised imitation learning task where the goal is to treat each state-action combination in expert distribution as an  Independent and Identically Distributed (I.I.D) example and minimise imitation loss for the trained policy:
\begin{equation}
 \arg\min_{\theta}E_{(s,a^*)}\sim_{P^*}L(a^*,\pi_{\theta}(s))\label{eq2}   
\end{equation}

where \(p^*(s\mid\pi^*)\) is an expert policy state distribution, and (state s, action $a^*$) is provided by expert policy \(p^*\).

\begin{figure}[!htbp]
\centerline{\includegraphics[scale=0.28]{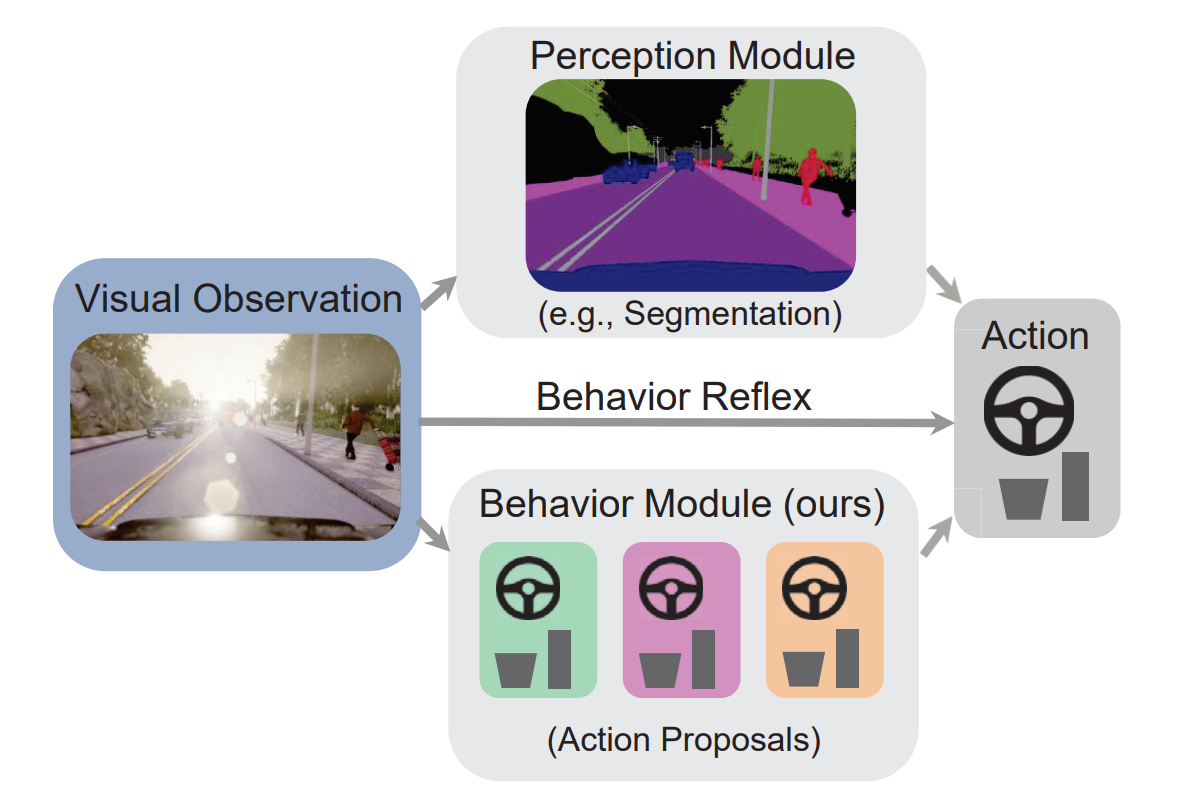}}
\caption{Behavior cloning \cite{ohn2020learningc62} is a perception-to-action driving model that learns behavior reflex for various driving scenarios. The agent acquires the ability to integrate expert policies in a context-dependent and task-optimized manner, allowing it to drive confidently.}
\label{f_behaviour}
\end{figure}

Prakash et al. \cite{prakash2021multic121}, Chitta et al. \cite{chitta2022transfuserc116}, NEAT \cite{chitta2021neatc100}, Ohn et al. \cite{ohn2020learningc62} utilize a policy that maps input frames to low-level control signals in terms of waypoints. These waypoints are then fed into a PID to obtain the steering, throttle, and brake commands based on the predicted waypoints. Behavior cloning \cite{ohn2020learningc62}  assumes that the expert's actions can be fully explained by observation, as it trains a model to
directly map from input to output based on the training dataset (Fig. \ref{f_behaviour}). However, this leads to the distribution shift problem, where the actual observations diverge from the training observations. Many latent variables impact and govern driving agent's actions in real-world scenarios. Therefore, it is essential to learn these variables effectively.

\subsubsection{Direct policy learning}

Within the context of BC, which maps sensor inputs to control commands and is limited by the training dataset, DLP aims to learn an optimal policy \cite{chen2020learningc60} directly that maps inputs to driving actions. The DLP algorithm obtains expert evaluations \cite{peng2022safe} during runtime to gather more training data, particularly for scenarios where the initial policy falls short. It combines an expert dataset with Imitation Learning for initial training and iteratively augments the dataset with additional trajectories collected by the trained policy. The agent can explore its surroundings and discover novel and efficient driving policies.

The online imitation learning algorithm DAGGER \cite{ross2011reductionc5} provides robustness against cascading errors and accumulates additional training instances. Chen et al. \cite{chen2022learning5orc98} introduced automated dagger-like monitoring, where the privileged agent's supervision is collected through online learning and transformed into an agent that provides on-policy supervision. However, the main drawback of direct policy learning is the continuous need for expert access during the training process, which is both costly and inefficient.

\subsubsection{Inverse reinforcement learning}
Inverse Reinforcement Learning (IRL) \cite{sadat2020perceive,chen2020learningc60} aims to deduce the underlying specific behaviours through the reward function. Expert demonstrations \(D = \left \{ \zeta _{1},\zeta _{2},\zeta _{3}, ......,\zeta _{n} \right \}\) are fed into IRL. Each  \(\zeta _{i}=\left \{ (s_{1},a_{2}),(s_{2},a_{2}),.....(s_n,a_n) \right \}\) consists of a state-action pair. The principal goal is to get the underlying reward which can be used to replicate the expert behaviour. Feature-based IRL \cite{sadigh2016planningc10} teaches the different driving styles in the highway scenario. The human-provided examples are used to learn different reward functions and capabilities of interaction with road users. Maximum Entropy (MaxEnt) inverse reinforcement learning \cite{ziebart2008maximumc11} is an extension of the feature-based IRL based on the principle of maximum entropy. This paradigm robustly addresses reward ambiguity and handles sub-optimization. The major drawback is that IRL algorithms are expensive to run. They are also computationally demanding, unstable during training, and may take longer to converge on smaller datasets.

\begin{figure*}[!htbp]
  \centering

  \begin{tabular}{@{}c@{}}
    \includegraphics[width=.3\linewidth,height=100pt]{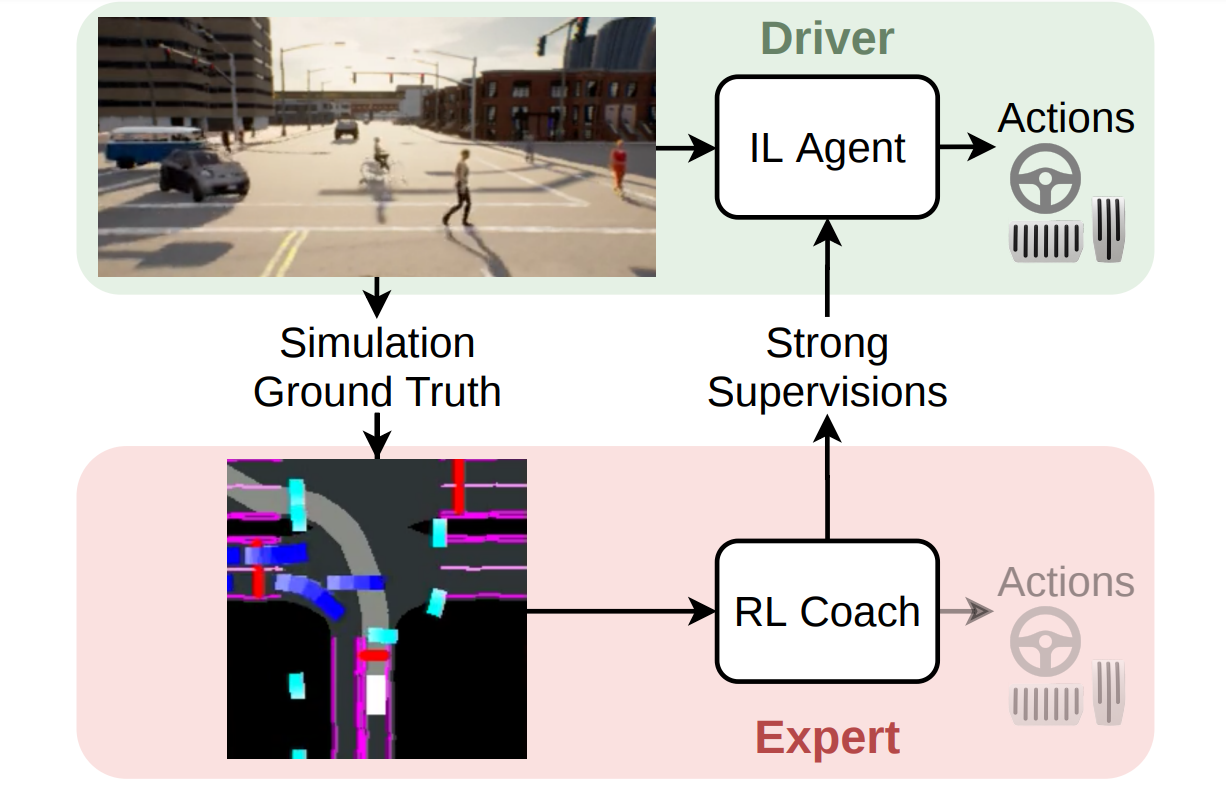} \\[\abovecaptionskip]
    \small (a) Roach Expert Supervision
  \end{tabular}
  \begin{tabular}{@{}c@{}}
    \includegraphics[width=.6\linewidth,height=100pt]{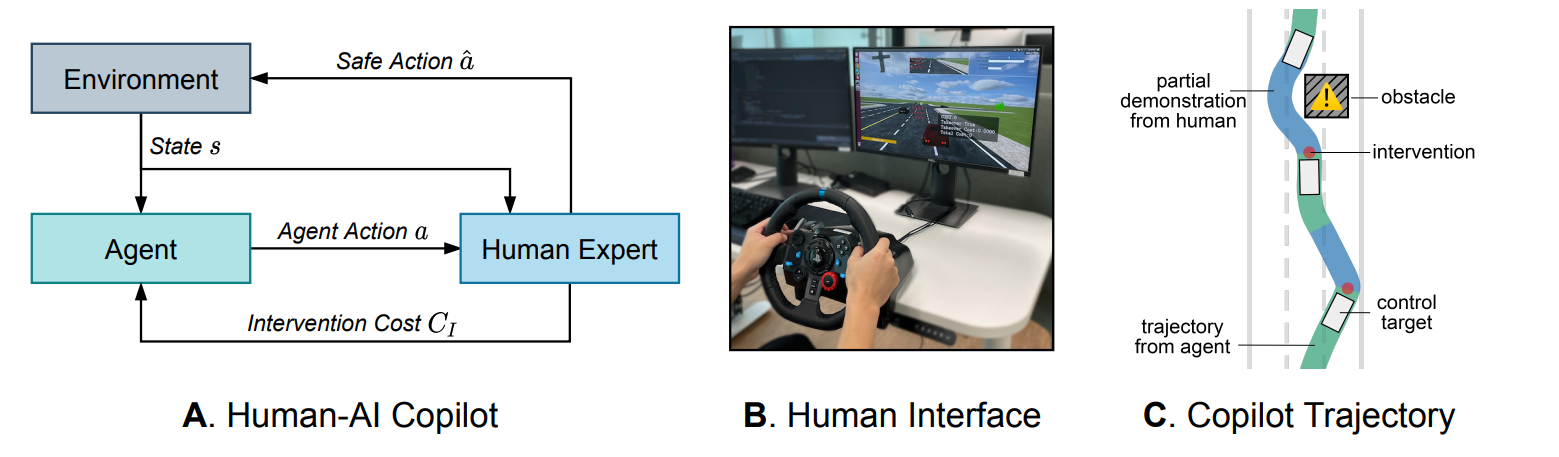} \\[\abovecaptionskip]
    \small (b) Human Copilot Learning
  \end{tabular}

  \caption{RL-based learning method for training the agent to drive optimally: (a) Illustrating the reinforcement learning expert \cite{zhang2021endc94} that maps the BEV to the low-level driving actions; the expert can also provide supervision to the imitation learning agent. (b) Human-in-the-loop learning \cite{li2022efficient4} allows the agent to explore the environment, and in danger scenarios, the human expert takes over the control and provides the safe demonstration.}\label{fig6}
  
\end{figure*}

\subsection{Reinforcement learning}\label{AA}

Reinforcement Learning (RL) \cite{wang2021versatile,zhao2022cadre,peng2022safe,toromanoff2020endc66} is a promising approach to address the distribution shift problem. It aims to maximize cumulative rewards \cite{sutton2018reinforcementc14} over time by interacting with the environment, and the network makes driving decisions to obtain rewards or penalties based on its actions. IL cannot handle novel situations significantly different from the training dataset, RL is robust to this issue as it explores scenario under given environment. Reinforcement learning encompasses various models, including value-based models such as Deep Q-Networks (DQN) \cite{chen2020conditionalc16}, actor-critic based models like Deep Deterministic Policy Gradient (DDPG) and Asynchronous Advantage Actor Critic (A3C) \cite{chen2020conditionalc16}, maximum entropy models \cite{ziebart2008maximumc11} such as Soft Actor Critic (SAC) \cite{chen2021interpretablec18}, and policy-based optimization methods such as Trust Region Policy Optimization (TRPO) and Proximal Policy Optimization (PPO) \cite{zhang2022recedingc19}.

Liang et al. \cite{liang2018cirl10or41orc125} demonstrated the first effective RL approach for vision-based driving pipelines that outperformed the modular pipeline at the time. Their method is based on the Deep Deterministic Policy Gradient (DDPG), an extended version of the actor-critic algorithm. Chen et al. \cite{chen2021learning11orc136} uses tabular-RL to first learn an expert policy and then uses policy distillation to learn a student policy in an imitation learning approach.

Recently, Human-In-The-Loop (HITL) approaches \cite{chekroun2021gric20, li2022efficient4, peng2022safe, zhang2021endc94,prakash2021multic121} have gained attention in the literature. These approaches are based on the premise that expert demonstrations provide valuable guidance for achieving high-reward policies. Several studies have focused on incorporating human expertise into the training process of traditional RL or IL paradigms. One such example is EGPO \cite{peng2022safe}, which aims to develop an expert-guided policy optimization technique where an expert policy supervises the learning agent.

HACO \cite{li2022efficient4} allows the agent to explore hazardous environments while ensuring training safety. In this approach, a human expert can intervene and guide the agent to avoid potentially harmful situations or irrelevant actions (see Fig. \ref{fig6}(b)). Another reinforcement learning expert, Roach \cite{zhang2021endc94}, translates bird's-eye view images into continuous low-level actions (see Fig. \ref{fig6}(a)). Experts can provide high-level supervision for imitation learning or reinforcement learning in general. Policies can be initially taught using imitation learning and then refined using reinforcement learning, which helps reduce the extensive training period required for RL. Jia et al. \cite{jia2023think} utilize features extracted from Roach to learn the ground-truth action/trajectory, providing supervision in their study. Therefore, reinforcement learning provides a solution to address the challenges of imitation learning by enabling agents to actively explore and learn from their environment. There are also associated challenges, such as sample inefficiency, exploration difficulties leading to suboptimal behaviors, and difficulties generalizing learned policies to new scenarios.

\section{Learning domain adaptation from simulator to real} \label{domain}

Large-scale virtual scenarios can be constructed in virtual engines, enabling the collection of a significant quantity of data more readily. However, there still exist significant domain disparities between virtual and real-world data, which pose challenges in creating and implementing virtual datasets. By leveraging the principle of domain adaptation, we can extract critical features directly from the simulator and transfer the knowledge learned from the source domain to the target domain, consisting of accurate real-world data.

The H-Divergence framework \cite{chen2018domain136} resolves the domain gap at both the visual and instance levels by adversarially learning a domain classifier and a detector simultaneously. Zhang et al. \cite{8947932137} propose a simulator-real interaction strategy that leverages disparities between the source domain and the target domain. The authors create two components to align differences at the global and local levels and ensure overall consistency between them. The realistic-looking synthetic images may subsequently be used to train an End-to-End model. A number of techniques rely on an open pipeline that introduces obstacles in the current environment. PlaceNet \cite{ZhangWMWHS20130} places objects into the image space for detection and segmentation operations. GeoSim \cite{chen2021geosim131} is a geometry-aware approach that dynamically inserts objects using LiDAR and HD maps. DummyNet \cite{vobecký2020artificial} is a pedestrian augmentation framework based on a GAN architecture that takes the background image as input and inserts pedestrians with consistent alignment. Some works take advantage of virtual LiDAR data \cite{sallab2019unsupervised141, 8972449142, manivasagam2020lidarsim151}. Sallab et al. \cite{sallab2019unsupervised141} perform learning on virtual LiDAR point clouds from CARLA \cite{dosovitskiy2017carla} and utilize CycleGAN to transfer styles from the virtual domain to the real KITTI \cite{geiger2013vision} dataset. Fang et al. \cite{8972449142} propose a LiDAR simulator that

\clearpage
\onecolumn
\begin{landscape}
\tiny
\begin{longtable}{|cccccccccc|}
\caption{RECENT METHODS IN END-TO-END AUTONOMOUS DRIVING}\\\toprule
\label{literature}\\
\hline
\rowcolor[HTML]{C0C0C0} 
Paper, Year &
  Environment &
  Input modality &
  Output modality &
  Learning &
  Evaluation &
  Explaibility &
  Safety &
  Results &
  Data \\ \hline
\endfirsthead
\caption* {Table \ref{literature} CONTINUED: RECENT METHODS IN END-TO-END AUTONOMOUS DRIVING}\\\toprule
\multicolumn{10}{c}%
{{ }} \\
\hline
\rowcolor[HTML]{C0C0C0} 
Paper, Year &
  Environment &
  Input modality &
  Output modality &
  Learning &
  Evaluation &
  Explaibility &
  Safety &
  Results &
  Data \\ \hline
\endhead
\multicolumn{1}{|c|}{\begin{tabular}[c]{@{}c@{}}PAD \cite{hu2023planningoriented},\\ 2023 \\ \end{tabular}} &
  \multicolumn{1}{c|}{NuScenes} &
  \multicolumn{1}{c|}{\begin{tabular}[c]{@{}c@{}}6 Multi-camera\\  images\end{tabular}} &
  \multicolumn{1}{c|}{\begin{tabular}[c]{@{}c@{}}Segmentation map,\\ agent future trajectories,\\ future \\ occupancy\end{tabular}} &
  \multicolumn{1}{c|}{\begin{tabular}[c]{@{}c@{}}Multi-task \\ learning.\end{tabular}} &
  \multicolumn{1}{c|}{\begin{tabular}[c]{@{}c@{}}min ADE, \\ min FDE,  IoU, L2\\ error\end{tabular}} &
  \multicolumn{1}{c|}{\begin{tabular}[c]{@{}c@{}}Exp. Via attention mask \\ in planner module\end{tabular}} &
  \multicolumn{1}{c|}{\begin{tabular}[c]{@{}c@{}}Lowest collision rate,\\ lowest  L2 error, Safety \\ boost via goal planner.\end{tabular}} &
  \multicolumn{1}{c|}{\begin{tabular}[c]{@{}c@{}}L2: 1.65\\ CR: 0.71\end{tabular}} &
  \begin{tabular}[c]{@{}c@{}}Perception training: 6 epochs \\ Joint training: 20 epochs\end{tabular} \\ \hline
\multicolumn{1}{|c|}{\begin{tabular}[c]{@{}c@{}}ReasonNet \cite{shao2023reasonnet}, \\ 2023\end{tabular}} &
  \multicolumn{1}{c|}{CARLA} &
  \multicolumn{1}{c|}{\begin{tabular}[c]{@{}c@{}}Vehicle measurements \\ navigation, \\ LiDAR, RGB\end{tabular}} &
  \multicolumn{1}{c|}{\begin{tabular}[c]{@{}c@{}}Steering, speed, BEV, \\ brake\end{tabular}} &
  \multicolumn{1}{c|}{Multi-task learning} &
  \multicolumn{1}{c|}{\begin{tabular}[c]{@{}c@{}}DOS and \\ LeaderBoard\end{tabular}} &
  \multicolumn{1}{c|}{Attention map} &
  \multicolumn{1}{c|}{-} &
  \multicolumn{1}{c|}{\begin{tabular}[c]{@{}c@{}}DS: 79.95\\ RC: 89.89\\ IS: 0.89\end{tabular}} &
  \begin{tabular}[c]{@{}c@{}}Train: 2M frames (eight Towns) \\ Test: Town05\end{tabular} \\ \hline
\multicolumn{1}{|c|}{\begin{tabular}[c]{@{}c@{}}Policy Pre-Training \cite{wu2023PPGeo}, \\ 2023\end{tabular}} &
  \multicolumn{1}{c|}{\begin{tabular}[c]{@{}c@{}}NuScenes,\\ CARLA\end{tabular}} &
  \multicolumn{1}{c|}{Camera and intrinsics} &
  \multicolumn{1}{c|}{Steering, throttle} &
  \multicolumn{1}{c|}{Imitation learning} &
  \multicolumn{1}{c|}{\begin{tabular}[c]{@{}c@{}}Longest6, \\ collision rate\end{tabular}} &
  \multicolumn{1}{c|}{\begin{tabular}[c]{@{}c@{}}Attention \\ activation map\end{tabular}} &
  \multicolumn{1}{c|}{\begin{tabular}[c]{@{}c@{}}Handles cases where the \\ agent needs to stop\end{tabular}} &
  \multicolumn{1}{c|}{\begin{tabular}[c]{@{}c@{}}DS: 47.4$\pm$5.6\\ RC: 65.05$\pm$5.1\\ IS: 0.79$\pm$0.08\\ L2: 3.01, \\ CR: 0.94\end{tabular}} &
  \begin{tabular}[c]{@{}c@{}}Train: 40K \\ (Town01, 03, 04, 06) \\ Test: 50 routes\end{tabular} \\ \hline
\multicolumn{1}{|c|}{\begin{tabular}[c]{@{}c@{}}Think Twice \cite{jia2023think}, \\ 2023\end{tabular}} &
  \multicolumn{1}{c|}{CARLA} &
  \multicolumn{1}{c|}{RGB camera, LiDAR} &
  \multicolumn{1}{c|}{\begin{tabular}[c]{@{}c@{}}BEV feature map, \\ agent future trajectories, \\ control actions\end{tabular}} &
  \multicolumn{1}{c|}{\begin{tabular}[c]{@{}c@{}}Open-loop imitation \\ learning\end{tabular}} &
  \multicolumn{1}{c|}{Longest6} &
  \multicolumn{1}{c|}{\begin{tabular}[c]{@{}c@{}}Explainable via \\ expert BEV feature map\end{tabular}} &
  \multicolumn{1}{c|}{\begin{tabular}[c]{@{}c@{}}Safety-critical regions, \\ anticipate \\ future motion\\  to avoid collisions\end{tabular}} &
  \multicolumn{1}{c|}{\begin{tabular}[c]{@{}c@{}}DS: 70.9$\pm$3.4 \\ RC: 95.5$\pm$2.6 \\ IS: 0.75$\pm$0.05\end{tabular}} &
  \begin{tabular}[c]{@{}c@{}}Training: 189K data on four Towns, \\ Roach based teaching\end{tabular} \\ \hline
\multicolumn{1}{|c|}{\begin{tabular}[c]{@{}c@{}}Scaling Vision-based \cite{xiao2023scaling},\\ 2023\end{tabular}} &
  \multicolumn{1}{c|}{CARLA} &
  \multicolumn{1}{c|}{Three cameras (views)} &
  \multicolumn{1}{c|}{\begin{tabular}[c]{@{}c@{}}Ego-vehicle’s \\ steering angle and \\ acceleration\end{tabular}} &
  \multicolumn{1}{c|}{\begin{tabular}[c]{@{}c@{}}Self-supervised \\ imitation learning\end{tabular}} &
  \multicolumn{1}{c|}{\begin{tabular}[c]{@{}c@{}}NoCrash, \\ Leaderboard\end{tabular}} &
  \multicolumn{1}{c|}{Attention map} &
  \multicolumn{1}{c|}{\begin{tabular}[c]{@{}c@{}}Pedestrians caused strong \\ braking (0.794), \\ despite \\ green light\end{tabular}} &
  \multicolumn{1}{c|}{\begin{tabular}[c]{@{}c@{}}DS: 98$\pm$1.7\\ RC: 68$\pm$2.7\end{tabular}} &
  \begin{tabular}[c]{@{}c@{}}Train: 15 hours (540K frames)\\ Test: 25 hours ( 900k frames)\end{tabular} \\ \hline
\multicolumn{1}{|c|}{\begin{tabular}[c]{@{}c@{}}Coaching a Teachable \cite{zhang2023coaching},\\ 2023\end{tabular}} &
  \multicolumn{1}{c|}{CARLA} &
  \multicolumn{1}{c|}{\begin{tabular}[c]{@{}c@{}}Navigation, \\ 3 RGB cameras\end{tabular}} &
  \multicolumn{1}{c|}{\begin{tabular}[c]{@{}c@{}}Waypoints,\\ control commands\end{tabular}} &
  \multicolumn{1}{c|}{\begin{tabular}[c]{@{}c@{}}Knowledge \\ distillation, \\ imitation learning\end{tabular}} &
  \multicolumn{1}{c|}{\begin{tabular}[c]{@{}c@{}}Longest6,  \\ ADE, FDE\end{tabular}} &
  \multicolumn{1}{c|}{-} &
  \multicolumn{1}{c|}{\begin{tabular}[c]{@{}c@{}}Attend to safety-critical \\ entities\end{tabular}} &
  \multicolumn{1}{c|}{\begin{tabular}[c]{@{}c@{}}DS: 73.30$\pm$1.07 \\ RC: 87.44$\pm$0.28\\ ADE: 0.41 \\FDE: 0.36\end{tabular}} &
  Train: Town01 - Town06 \\ \hline
\multicolumn{1}{|c|}{\begin{tabular}[c]{@{}c@{}}Hidden Biases of \cite{Jaeger2023ICCV},\\ 2023\end{tabular}} &
  \multicolumn{1}{c|}{CARLA} &
  \multicolumn{1}{c|}{\begin{tabular}[c]{@{}c@{}}Speed,\\ camera and\\ LiDAR\end{tabular}} &
  \multicolumn{1}{c|}{\begin{tabular}[c]{@{}c@{}}Steer, throttle, and brake, \\ waypoints, path\end{tabular}} &
  \multicolumn{1}{c|}{Imitation learning} &
  \multicolumn{1}{c|}{Longest6} &
  \multicolumn{1}{c|}{-} &
  \multicolumn{1}{c|}{\begin{tabular}[c]{@{}c@{}}Larger safety distance, \\ safety area\end{tabular}} &
  \multicolumn{1}{c|}{\begin{tabular}[c]{@{}c@{}}DS: 72 $\pm$3 \\ RC: 95 $\pm$ 2\end{tabular}} &
  Train: variable size 185k and 555k \\ \hline
\multicolumn{1}{|c|}{\begin{tabular}[c]{@{}c@{}}KING \cite{hanselmann2022king2orc99}, \\ 2022\end{tabular}} &
  \multicolumn{1}{c|}{CARLA} &
  \multicolumn{1}{c|}{\begin{tabular}[c]{@{}c@{}}Front-facing \\ camera and\\ LiDAR\end{tabular}} &
  \multicolumn{1}{c|}{Throttle and steering} &
  \multicolumn{1}{c|}{Behaviour cloning} &
  \multicolumn{1}{c|}{\begin{tabular}[c]{@{}c@{}}Closed-loop (CR),\\ (RC), (IS), (DS)\end{tabular}} &
  \multicolumn{1}{c|}{-} &
  \multicolumn{1}{c|}{\begin{tabular}[c]{@{}c@{}}Safety-critical \\ driving \\ scenarios and\\ stress-testing.\end{tabular}} &
  \multicolumn{1}{c|}{\begin{tabular}[c]{@{}c@{}}DS: 90.20$\pm$0.00\\ RC: 94.42$\pm$0.36\\ IS: 0.96$\pm$0.36\end{tabular}} &
  \begin{tabular}[c]{@{}c@{}}Train: 4 GPU\\ hours 80 routes\\ on Town 3,4,5,6.\end{tabular} \\ \hline
\multicolumn{1}{|c|}{\begin{tabular}[c]{@{}c@{}}LAV \cite{chen2022learning5orc98}, \\ 2022\end{tabular}} &
  \multicolumn{1}{c|}{CARLA} &
  \multicolumn{1}{c|}{\begin{tabular}[c]{@{}c@{}}Front-facing camera, \\ LiDAR\end{tabular}} &
  \multicolumn{1}{c|}{\begin{tabular}[c]{@{}c@{}}Single steering and \\ acceleration command\end{tabular}} &
  \multicolumn{1}{c|}{\begin{tabular}[c]{@{}c@{}}Knowledge \\ distillation, \\ imitation learning\end{tabular}} &
  \multicolumn{1}{c|}{\begin{tabular}[c]{@{}c@{}}Longest6 (DS), \\ (RC),(IS), \\ (CR), (PC), \\ (LC), (RV)\end{tabular}} &
  \multicolumn{1}{c|}{Spatial feature 2D map} &
  \multicolumn{1}{c|}{\begin{tabular}[c]{@{}c@{}}Scenario such \\ as pedestrians crossing, \\ lane change \\ are modeled\end{tabular}} &
  \multicolumn{1}{c|}{\begin{tabular}[c]{@{}c@{}}DS: 61.85 \\ RC: 94.46 \\ IS: 0.64\end{tabular}} &
  \begin{tabular}[c]{@{}c@{}}Train: Four Towns, 186K frames \\ Test: Town02 and Town05\end{tabular} \\ \hline
\multicolumn{1}{|c|}{\begin{tabular}[c]{@{}c@{}}TransFuse \cite{chitta2022transfuserc116}, \\ 2022\end{tabular}} &
  \multicolumn{1}{c|}{CARLA} &
  \multicolumn{1}{c|}{\begin{tabular}[c]{@{}c@{}}RGB, \\ LiDAR\end{tabular}} &
  \multicolumn{1}{c|}{\begin{tabular}[c]{@{}c@{}}Waypoints, steer, throttle, \\ and brake\end{tabular}} &
  \multicolumn{1}{c|}{Behaviour cloning} &
  \multicolumn{1}{c|}{\begin{tabular}[c]{@{}c@{}}Longest 6 (DS), \\ (RC), (IPK), (IS)\end{tabular}} &
  \multicolumn{1}{c|}{\begin{tabular}[c]{@{}c@{}}Explainable via auxiliary \\ output like depth, \\ semantic, HD map, \\ planner A*\end{tabular}} &
  \multicolumn{1}{c|}{Global safety heuristic} &
  \multicolumn{1}{c|}{\begin{tabular}[c]{@{}c@{}}DS: 61.18 \\ RC: 86.69 \\ IS: 0.71\end{tabular}} &
  \begin{tabular}[c]{@{}c@{}}Train: 2500 routes on eight \\ Towns, 228k frames \\ Test: 36 route\end{tabular} \\ \hline
\multicolumn{1}{|c|}{\begin{tabular}[c]{@{}c@{}}Learning to Drive \cite{zhang2022learning}, \\ 2022\end{tabular}} &
  \multicolumn{1}{c|}{\begin{tabular}[c]{@{}c@{}}NuScenes,\\ youtube\end{tabular}} &
  \multicolumn{1}{c|}{Front-facing camera} &
  \multicolumn{1}{c|}{\begin{tabular}[c]{@{}c@{}}Steering, throttle, \\ brake, \\ and velocity\end{tabular}} &
  \multicolumn{1}{c|}{\begin{tabular}[c]{@{}c@{}}Conditional \\ behavior cloning\end{tabular}} &
  \multicolumn{1}{c|}{\begin{tabular}[c]{@{}c@{}}CARLA benchmark, \\ F1 metric\end{tabular}} &
  \multicolumn{1}{c|}{-} &
  \multicolumn{1}{c|}{-} &
  \multicolumn{1}{c|}{\begin{tabular}[c]{@{}c@{}}F1 : 75.0\\ 7 perc. increase in RC\end{tabular}} &
  \begin{tabular}[c]{@{}c@{}}Train: 120 hour YouTube, \\ Test: Town01 and \\ Town02 with \\ 40K transition\end{tabular} \\ \hline
\multicolumn{1}{|c|}{\begin{tabular}[c]{@{}c@{}}HACO \cite{li2022efficient4}, \\ 2022\end{tabular}} &
  \multicolumn{1}{c|}{CARLA} &
  \multicolumn{1}{c|}{\begin{tabular}[c]{@{}c@{}}Current state,\\ navigation \\ information\end{tabular}} &
  \multicolumn{1}{c|}{\begin{tabular}[c]{@{}c@{}}Acceleration, brake \\ and steering\end{tabular}} &
  \multicolumn{1}{c|}{\begin{tabular}[c]{@{}c@{}}Reinforcement \\ learning\end{tabular}} &
  \multicolumn{1}{c|}{\begin{tabular}[c]{@{}c@{}}Safety violation\\ data usage,\\ success rate\end{tabular}} &
  \multicolumn{1}{c|}{-} &
  \multicolumn{1}{c|}{\begin{tabular}[c]{@{}c@{}}Safety violation \\ cost the episode\end{tabular}} &
  \multicolumn{1}{c|}{\begin{tabular}[c]{@{}c@{}}SV: 11.84 \\ SR: 0.35\end{tabular}} &
  \begin{tabular}[c]{@{}c@{}}Train: 50 min \\ HACO training.\\ 35k transition\end{tabular} \\ \hline
\multicolumn{1}{|c|}{\begin{tabular}[c]{@{}c@{}}PlanT \cite{renz2022plantc121c},\\ 2022\end{tabular}} &
  \multicolumn{1}{c|}{CARLA} &
  \multicolumn{1}{c|}{\begin{tabular}[c]{@{}c@{}}Camera ,\\ LiDAR,\\ route, object\end{tabular}} &
  \multicolumn{1}{c|}{\begin{tabular}[c]{@{}c@{}}Waypoints, predicting \\ the future attributes of \\ other vehicles\end{tabular}} &
  \multicolumn{1}{c|}{\begin{tabular}[c]{@{}c@{}}Imitation \\ learning\end{tabular}} &
  \multicolumn{1}{c|}{\begin{tabular}[c]{@{}c@{}}Longest6 CARLA\\ (DS),(RC),\\ (IS), (IPK),\\ (IT)\end{tabular}} &
  \multicolumn{1}{c|}{\begin{tabular}[c]{@{}c@{}}Post hoc explanations,\\ attention weights for \\ identify relevant objects.\end{tabular}} &
  \multicolumn{1}{c|}{\begin{tabular}[c]{@{}c@{}}Identification of collision \\ free routes, \\ RFDS algorithm\\ for collision avoidance\end{tabular}} &
  \multicolumn{1}{c|}{\begin{tabular}[c]{@{}c@{}}DS: 81.36$\pm$6.54 \\ RC: 93.55$\pm$2.62 \\ IS: 0.87$\pm$0.05\end{tabular}} &
  \begin{tabular}[c]{@{}c@{}}Train: 3.2 hours on PlanT,\\  95 hours of \\ driving.\end{tabular} \\ \hline
\multicolumn{1}{|c|}{\begin{tabular}[c]{@{}c@{}}Trajectory-guided \cite{wu2022trajectoryc120}, \\ 2022\end{tabular}} &
  \multicolumn{1}{c|}{CARLA} &
  \multicolumn{1}{c|}{\begin{tabular}[c]{@{}c@{}}Monocular\\ camera\end{tabular}} &
  \multicolumn{1}{c|}{\begin{tabular}[c]{@{}c@{}}Steering, \\ throttle and brake\end{tabular}} &
  \multicolumn{1}{c|}{\begin{tabular}[c]{@{}c@{}}Behaviour \\ cloning\end{tabular}} &
  \multicolumn{1}{c|}{\begin{tabular}[c]{@{}c@{}}CARLA (DS), \\ (RC), (IS)\end{tabular}} &
  \multicolumn{1}{c|}{\begin{tabular}[c]{@{}c@{}}Auxiliary tasks \\ include value \\ and speed head\end{tabular}} &
  \multicolumn{1}{c|}{\begin{tabular}[c]{@{}c@{}}Reduce collision via long \\ prediction, \\ less infraction,\\ control model preform good\end{tabular}} &
  \multicolumn{1}{c|}{\begin{tabular}[c]{@{}c@{}}DS: 75.14 \\ RC: 85.63 \\ IS: 0.87\end{tabular}} &
  \begin{tabular}[c]{@{}c@{}}Train: Town01, 03, 04, 06 \\ Test: Town02, 05\end{tabular} \\ \hline
\multicolumn{1}{|c|}{\begin{tabular}[c]{@{}c@{}}Safety- Enhanced \\ Autonomous\\ Driving \cite{shao2023safetyc117},\\ 2022\end{tabular}} &
  \multicolumn{1}{c|}{CARLA} &
  \multicolumn{1}{c|}{\begin{tabular}[c]{@{}c@{}}3 RGB,\\ 1 LiDAR\end{tabular}} &
  \multicolumn{1}{c|}{\begin{tabular}[c]{@{}c@{}}10 Waypoints,\\ steering,\\ acceleration\end{tabular}} &
  \multicolumn{1}{c|}{InterFuser} &
  \multicolumn{1}{c|}{\begin{tabular}[c]{@{}c@{}}CARLA  leaderboard,\\ CARLA 42 Routes \\ benchmark,\\ (RC),(IS),(DS)\end{tabular}} &
  \multicolumn{1}{c|}{\begin{tabular}[c]{@{}c@{}}Intermediate interpretable \\ features output from \\ transformer decoder\\ (safety map, object density)\end{tabular}} &
  \multicolumn{1}{c|}{\begin{tabular}[c]{@{}c@{}}Safe set contain only \\ safe actions, safety sensitive \\ output via InterFuser\end{tabular}} &
  \multicolumn{1}{c|}{\begin{tabular}[c]{@{}c@{}}DS: 76.18 \\ RC: 88.23 \\ IS: 0.84\end{tabular}} &
  \begin{tabular}[c]{@{}c@{}}Train: 3M frames or \\ 410 hours (eight Towns)\end{tabular} \\ \hline
\multicolumn{1}{|c|}{\begin{tabular}[c]{@{}c@{}}ST-P3 \cite{hu2022st},\\ 2022\end{tabular}} &
  \multicolumn{1}{c|}{\begin{tabular}[c]{@{}c@{}}NuScenes,\\ CARLA\end{tabular}} &
  \multicolumn{1}{c|}{\begin{tabular}[c]{@{}c@{}}6 Camera(NuScenes),\\ 4 cameras (CARLA),\\ navigation command\end{tabular}} &
  \multicolumn{1}{c|}{\begin{tabular}[c]{@{}c@{}}Steering, \\ throttle and brake\end{tabular}} &
  \multicolumn{1}{c|}{ST-P3} &
  \multicolumn{1}{c|}{\begin{tabular}[c]{@{}c@{}}Open loop: IOU, PQ, \\ RQ, SQ, \\ L2 error \\ Closed: (DS), (RC)\end{tabular}} &
  \multicolumn{1}{c|}{\begin{tabular}[c]{@{}c@{}}Interpretable map\\ lanes and \\ area which is drivable\end{tabular}} &
  \multicolumn{1}{c|}{\begin{tabular}[c]{@{}c@{}}Safety Cost function\\ for the jerk action\\ penalize\end{tabular}} &
  \multicolumn{1}{c|}{\begin{tabular}[c]{@{}c@{}}DS: 55.14 \\ RC: 86.74\\ L2: 2.90, CR: 1.27\end{tabular}} &
  \begin{tabular}[c]{@{}c@{}}Train: 26124 samples\\ Validation: 5719 samples\end{tabular} \\ \hline
\multicolumn{1}{|c|}{\begin{tabular}[c]{@{}c@{}}Safe Driving via \\ Expert Guided \\ Policy
optimization \cite{peng2022safe},\\ 2022\end{tabular}} &
  \multicolumn{1}{c|}{MetaDrive} &
  \multicolumn{1}{c|}{Camera} &
  \multicolumn{1}{c|}{Control signal} &
  \multicolumn{1}{c|}{\begin{tabular}[c]{@{}c@{}}Expert-in-the-loop \\ reinforcement \\ learning\end{tabular}} &
  \multicolumn{1}{c|}{MetaDriv benchmark} &
  \multicolumn{1}{c|}{-} &
  \multicolumn{1}{c|}{\begin{tabular}[c]{@{}c@{}}Guardian to ensure \\ training safety\end{tabular}} &
  \multicolumn{1}{c|}{\begin{tabular}[c]{@{}c@{}}ER: 388.37 $\pm$10.01 \\ EC: 0.56 $\pm$0.35 \\ SR: 0.85 $\pm$0.05\end{tabular}} &
  \begin{tabular}[c]{@{}c@{}}Train: 100 scenes\\ Test: 50 scenes\end{tabular} \\ \hline
\multicolumn{1}{|c|}{\begin{tabular}[c]{@{}c@{}}COOPERNAUT \cite{cui2022coopernaut} ,\\ 2022\end{tabular}} &
  \multicolumn{1}{c|}{CARLA} &
  \multicolumn{1}{c|}{\begin{tabular}[c]{@{}c@{}}Front-facing camera, \\ LiDAR\end{tabular}} &
  \multicolumn{1}{c|}{Control signal} &
  \multicolumn{1}{c|}{\begin{tabular}[c]{@{}c@{}}Behaviour \\ cloning\end{tabular}} &
  \multicolumn{1}{c|}{AUTOCASTSIM} &
  \multicolumn{1}{c|}{-} &
  \multicolumn{1}{c|}{\begin{tabular}[c]{@{}c@{}}Reduces safety \\ hazards \\ for line-of \\ sight sensing.\end{tabular}} &
  \multicolumn{1}{c|}{\begin{tabular}[c]{@{}c@{}}SR: 90.5$\pm$1.2 \\ CR: 4.5$\pm$3.1\end{tabular}} &
  \begin{tabular}[c]{@{}c@{}}Train: 12 traces + 84 traces\\ Test : 27 accident \\ prone traces\end{tabular} \\ \hline
\multicolumn{1}{|c|}{\begin{tabular}[c]{@{}c@{}}Human-AI Shared \\ Control via Policy \cite{li2022human},\\ 2022\end{tabular}} &
  \multicolumn{1}{c|}{MetaDrive} &
  \multicolumn{1}{c|}{\begin{tabular}[c]{@{}c@{}}Current state, \\ goal state\end{tabular}} &
  \multicolumn{1}{c|}{Control signal} &
  \multicolumn{1}{c|}{\begin{tabular}[c]{@{}c@{}}Reinforcement\\ learning\end{tabular}} &
  \multicolumn{1}{c|}{\begin{tabular}[c]{@{}c@{}}MetaDrive and \\ Pybullet-A1\end{tabular}} &
  \multicolumn{1}{c|}{\begin{tabular}[c]{@{}c@{}}Interpretable control \\ interface\end{tabular}} &
  \multicolumn{1}{c|}{\begin{tabular}[c]{@{}c@{}}Human-AI, \\ safety guarantee \\ 95 percentage \\ success rate\end{tabular}} &
  \multicolumn{1}{c|}{\begin{tabular}[c]{@{}c@{}}EC: 0.05 $\pm$0.08 \\ SR: 0.95 $\pm$ 0.02\end{tabular}} &
  \begin{tabular}[c]{@{}c@{}}Train: 50 training scenario \\ Test: 20 test scene\end{tabular} \\ \hline
\multicolumn{1}{|c|}{\begin{tabular}[c]{@{}c@{}}MMFN: Multi-Modal\\ Fusion-Net for \cite{zhang2022mmfn},\\ 2022\end{tabular}} &
  \multicolumn{1}{c|}{CARLA} &
  \multicolumn{1}{c|}{\begin{tabular}[c]{@{}c@{}}HD map and \\ radar on top of the\\ LiDAR and \\ camera\end{tabular}} &
  \multicolumn{1}{c|}{\begin{tabular}[c]{@{}c@{}}Steering, \\ throttle and brake\end{tabular}} &
  \multicolumn{1}{c|}{\begin{tabular}[c]{@{}c@{}}Imitation \\ learning\end{tabular}} &
  \multicolumn{1}{c|}{\begin{tabular}[c]{@{}c@{}}CARLA \\ LeaderBoard\end{tabular}} &
  \multicolumn{1}{c|}{-} &
  \multicolumn{1}{c|}{\begin{tabular}[c]{@{}c@{}}Expert has more \\ awareness of safe \\ driving\end{tabular}} &
  \multicolumn{1}{c|}{\begin{tabular}[c]{@{}c@{}}DS: 22.8 \\ RC: 47.22\end{tabular}} &
  \begin{tabular}[c]{@{}c@{}}Train: 207K frames\\ Test: 20 routes\end{tabular} \\ \hline
\multicolumn{1}{|c|}{\begin{tabular}[c]{@{}c@{}}CADRE: A Cascade \\ Deep Reinforcement \cite{zhao2022cadre},\\ 2022\end{tabular}} &
  \multicolumn{1}{c|}{CARLA} &
  \multicolumn{1}{c|}{\begin{tabular}[c]{@{}c@{}}Front-view camera, \\ position, \\ orientation \\ and speed\end{tabular}} &
  \multicolumn{1}{c|}{Control signal} &
  \multicolumn{1}{c|}{\begin{tabular}[c]{@{}c@{}}Imitation \\ learning\end{tabular}} &
  \multicolumn{1}{c|}{\begin{tabular}[c]{@{}c@{}}CARLA \\ NoCrash\end{tabular}} &
  \multicolumn{1}{c|}{-} &
  \multicolumn{1}{c|}{-} &
  \multicolumn{1}{c|}{\begin{tabular}[c]{@{}c@{}}VA: 81/81\\ PA: 76/78\end{tabular}} &
  \begin{tabular}[c]{@{}c@{}}Train:25 training routes\\ Test: Town02\end{tabular} \\ \hline
\multicolumn{1}{|c|}{{\begin{tabular}[c]{@{}c@{}}Model-Based Imitation \\ Learning for \cite{hu2022model}, \\ 2022\end{tabular}}} &
  \multicolumn{1}{c|}{CARLA} &
  \multicolumn{1}{c|}{RGB image, route} &
  \multicolumn{1}{c|}{\begin{tabular}[c]{@{}c@{}}Vehicle control,\\ BEV Segmentation\end{tabular}} &
  \multicolumn{1}{c|}{\begin{tabular}[c]{@{}c@{}}Model-based \\ imitation learning\end{tabular}} &
  \multicolumn{1}{c|}{\begin{tabular}[c]{@{}c@{}}CARLA \\ LeaderBoard\end{tabular}} &
  \multicolumn{1}{c|}{\begin{tabular}[c]{@{}c@{}}BEV semantic \\ segmentation for \\ interpretability\end{tabular}} &
  \multicolumn{1}{c|}{-} &
  \multicolumn{1}{c|}{\begin{tabular}[c]{@{}c@{}}DS: 61.1 $\pm$ 3.2 \\ RC: 97.4$\pm$ 0.8 \\ IS: 63.0 $\pm$ 3.0\end{tabular}} &
  \begin{tabular}[c]{@{}c@{}}Test: Town05, routes\\ Train: Four different training \\ towns total \\ of 2.9M frames.\end{tabular} \\ \hline
\multicolumn{1}{|c|}{\begin{tabular}[c]{@{}c@{}}LookOut: Diverse \\ Multi-Future \\ Prediction and Planning \cite{cui2021lookout},\\ 2021\end{tabular}} &
  \multicolumn{1}{c|}{\begin{tabular}[c]{@{}c@{}}ATG4D,\\ Lidarsim\end{tabular}} &
  \multicolumn{1}{c|}{\begin{tabular}[c]{@{}c@{}}LiDAR, \\ navigation.\end{tabular}} &
  \multicolumn{1}{c|}{\begin{tabular}[c]{@{}c@{}}Trajectory, \\ vehicle control\end{tabular}} &
  \multicolumn{1}{c|}{Cost learning} &
  \multicolumn{1}{c|}{\begin{tabular}[c]{@{}c@{}}Open loop: mAP, \\ mSADE, \\ mSADE, PlanASD\\ Cloose loop: Lidarsim\end{tabular}} &
  \multicolumn{1}{c|}{-} &
  \multicolumn{1}{c|}{\begin{tabular}[c]{@{}c@{}}Cost function include \\ driving including \\ safety, comfort, \\ traffic-rules.\end{tabular}} &
  \multicolumn{1}{c|}{\begin{tabular}[c]{@{}c@{}}CR: 7.93\\ Progress:62.65\\ Jerk: 4.69\end{tabular}} &
  \begin{tabular}[c]{@{}c@{}}Train: one million frames\\ Test: Lidarsim\end{tabular} \\ \hline
\multicolumn{1}{|c|}{\begin{tabular}[c]{@{}c@{}}MP3: A Unified Model \\ to Map, Perceive, \\ Predict and Plan \cite{casas2021mp3}, \\ 2021\end{tabular}} &
  \multicolumn{1}{c|}{URBANEXPERT} &
  \multicolumn{1}{c|}{\begin{tabular}[c]{@{}c@{}}Raw sensor data \\ and a high-level\\ command\end{tabular}} &
  \multicolumn{1}{c|}{\begin{tabular}[c]{@{}c@{}}Control command,\\ dynamic occupancy \\ field\end{tabular}} &
  \multicolumn{1}{c|}{MP3} &
  \multicolumn{1}{c|}{\begin{tabular}[c]{@{}c@{}}Closed-loop: Lidarsim\\ Open loop: L2\end{tabular}} &
  \multicolumn{1}{c|}{\begin{tabular}[c]{@{}c@{}}Interpretable cost \\ functions, dynamic \\ occupancy field, \\ Interpretable Scene repr.\end{tabular}} &
  \multicolumn{1}{c|}{\begin{tabular}[c]{@{}c@{}}Penalize trajectories \\ where the SDV overlaps \\ occupied regions, penalize\\ jerk, lateral acceleration\end{tabular}} &
  \multicolumn{1}{c|}{\begin{tabular}[c]{@{}c@{}}L2: 12.95\\ Collision: 1037.08\\ Jerk: 1.64 \\ Success Pre: 74.39\end{tabular}} &
  \begin{tabular}[c]{@{}c@{}}Train: 5000 scenarios \\ Test: 1000\end{tabular} \\ \hline
\multicolumn{1}{|c|}{\begin{tabular}[c]{@{}c@{}}Object-Aware \\ Regularization \\ for Addressing Causal \cite{park2021object},\\ 2021\end{tabular}} &
  \multicolumn{1}{c|}{CARLA} &
  \multicolumn{1}{c|}{RGB image} &
  \multicolumn{1}{c|}{Control signal} &
  \multicolumn{1}{c|}{\begin{tabular}[c]{@{}c@{}}Behaviour \\ cloning\end{tabular}} &
  \multicolumn{1}{c|}{\begin{tabular}[c]{@{}c@{}}Atari environment,\\ CARLA .\end{tabular}} &
  \multicolumn{1}{c|}{-} &
  \multicolumn{1}{c|}{Policy safe adaptation.} &
  \multicolumn{1}{c|}{\begin{tabular}[c]{@{}c@{}}Straight: 87$\pm$4.4 \\ turn: 70.0$\pm$ 7.2 \\ Nav: 35.7$\pm$10.2\end{tabular}} &
  \begin{tabular}[c]{@{}c@{}}Train: 150 demonstrations\\ Test : 25 routes\end{tabular} \\ \hline
\multicolumn{1}{|c|}{\begin{tabular}[c]{@{}c@{}}GRI: General Reinforced \\ Imitation and its\\application to vision-based \cite{chekroun2021gric20},\\ 2021\end{tabular}} &
  \multicolumn{1}{c|}{CARLA} &
  \multicolumn{1}{c|}{3 RGB camera view} &
  \multicolumn{1}{c|}{Control signal} &
  \multicolumn{1}{c|}{\begin{tabular}[c]{@{}c@{}}Off-policy \\ reinforcement\\ learning\end{tabular}} &
  \multicolumn{1}{c|}{\begin{tabular}[c]{@{}c@{}}CARLA Leaderboard,\\ NoCrash, \\ Mujoco benchmark\end{tabular}} &
  \multicolumn{1}{c|}{-} &
  \multicolumn{1}{c|}{-} &
  \multicolumn{1}{c|}{\begin{tabular}[c]{@{}c@{}}DS: 36.79 \\ RC: 61.85 \\ IS: 0.60\end{tabular}} &
  \begin{tabular}[c]{@{}c@{}}Train: 60M steps \\ Test: with 12M and 16M steps\end{tabular} \\ \hline
\multicolumn{1}{|c|}{\begin{tabular}[c]{@{}c@{}}Multi-Modal\\ Fusion \cite{prakash2021multic121},\\ 2021\end{tabular}} &
  \multicolumn{1}{c|}{CARLA} &
  \multicolumn{1}{c|}{\begin{tabular}[c]{@{}c@{}}Front-facing \\ camera and\\ LiDAR\end{tabular}} &
  \multicolumn{1}{c|}{\begin{tabular}[c]{@{}c@{}}4 Waypoints, \\ steer, throttle, \\ and brake\end{tabular}} &
  \multicolumn{1}{c|}{\begin{tabular}[c]{@{}c@{}}Imitation \\ learning\end{tabular}} &
  \multicolumn{1}{c|}{\begin{tabular}[c]{@{}c@{}}CARLA \\ LeaderBoard\end{tabular}} &
  \multicolumn{1}{c|}{\begin{tabular}[c]{@{}c@{}}Attention map \\ visualizations\end{tabular}} &
  \multicolumn{1}{c|}{\begin{tabular}[c]{@{}c@{}}Handle adversarial \\ scenarios in urban \\ driving, \\ e.g., hard turnings.\end{tabular}} &
  \multicolumn{1}{c|}{\begin{tabular}[c]{@{}c@{}}DS: 33.15 $\pm$ 4.04 \\ RC: 56.36 $\pm$ 7.14\end{tabular}} &
  \begin{tabular}[c]{@{}c@{}}Train: 7 towns \\ Test: Twon05\end{tabular} \\ \hline
\multicolumn{1}{|c|}{\begin{tabular}[c]{@{}c@{}}Learning by \\ Watching \cite{zhang2021learning},\\ 2021\end{tabular}} &
  \multicolumn{1}{c|}{CARLA} &
  \multicolumn{1}{c|}{\begin{tabular}[c]{@{}c@{}}Speed,\\ high-level\\ navigation\end{tabular}} &
  \multicolumn{1}{c|}{\begin{tabular}[c]{@{}c@{}}Waypoints, \\ steer, throttle, \\ and brake\end{tabular}} &
  \multicolumn{1}{c|}{\begin{tabular}[c]{@{}c@{}}Imitation \\ learning\end{tabular}} &
  \multicolumn{1}{c|}{\begin{tabular}[c]{@{}c@{}}CARLA \\ NoCrash \\ benchmarks\end{tabular}} &
  \multicolumn{1}{c|}{BEV visibility map} &
  \multicolumn{1}{c|}{\begin{tabular}[c]{@{}c@{}}Avoid an unsafe\\ maneuver.\end{tabular}} &
  \multicolumn{1}{c|}{\begin{tabular}[c]{@{}c@{}}NC-R: 92\\ NC-D: 24\\ OB: 92\end{tabular}} &
  \begin{tabular}[c]{@{}c@{}}Train: Town 1\\ Test: Town 2\end{tabular} \\ \hline
\multicolumn{1}{|c|}{\begin{tabular}[c]{@{}c@{}}NEAT \cite{chitta2021neatc100},\\ 2021\end{tabular}} &
  \multicolumn{1}{c|}{CARLA} &
  \multicolumn{1}{c|}{\begin{tabular}[c]{@{}c@{}}RGB cameras,\\ and intrinsics,\\ locations,speed\end{tabular}} &
  \multicolumn{1}{c|}{\begin{tabular}[c]{@{}c@{}}Waypoints,\\ BEV as a \\ auxiliary output\end{tabular}} &
  \multicolumn{1}{c|}{\begin{tabular}[c]{@{}c@{}}Behaviour \\ cloning\end{tabular}} &
  \multicolumn{1}{c|}{\begin{tabular}[c]{@{}c@{}}CARLA \\ Leaderboard.\\ (RC), (IS), (DS)\end{tabular}} &
  \multicolumn{1}{c|}{\begin{tabular}[c]{@{}c@{}}NEAT intermediate \\ representations provides\\ interpretable attention map\end{tabular}} &
  \multicolumn{1}{c|}{\begin{tabular}[c]{@{}c@{}}At that time highest safety \\ among other \\ methods on the CARLA.\end{tabular}} &
  \multicolumn{1}{c|}{\begin{tabular}[c]{@{}c@{}}DS: 24.08$\pm$3.30 \\ RC: 59.94$\pm$0.50 \\ IS: 0.49$\pm$0.02\end{tabular}} &
  \begin{tabular}[c]{@{}c@{}}Train: 8 towns\\ Test: (Town01- \\ Town 06) \\ 100 secret routes\end{tabular} \\ \hline
\multicolumn{1}{|c|}{\begin{tabular}[c]{@{}c@{}}End-to-End Urban\\ Driving \cite{zhang2021endc94},\\  2021\end{tabular}} &
  \multicolumn{1}{c|}{CARLA} &
  \multicolumn{1}{c|}{\begin{tabular}[c]{@{}c@{}}Wide-angle camera \\ image with a \\ 100 degree \\ horizontal FOV\end{tabular}} &
  \multicolumn{1}{c|}{\begin{tabular}[c]{@{}c@{}}Steering, \\ throttle and brake\end{tabular}} &
  \multicolumn{1}{c|}{\begin{tabular}[c]{@{}c@{}}Reinforcement \\ learning expert,\\ imitation learning\end{tabular}} &
  \multicolumn{1}{c|}{\begin{tabular}[c]{@{}c@{}}CARLA \\ NoCrash and\\ LeaderBoard\end{tabular}} &
  \multicolumn{1}{c|}{-} &
  \multicolumn{1}{c|}{-} &
  \multicolumn{1}{c|}{\begin{tabular}[c]{@{}c@{}}DS: 55.27$\pm$1.43 \\ RC: 88.16$\pm$1.52 \\ IS: 0.62$\pm$0.02\end{tabular}} &
  \begin{tabular}[c]{@{}c@{}}Off-policy \\ dataset 80 episodes, \\ Train: 50 routes\\ Test: 26 routes\end{tabular} \\ \hline
\multicolumn{1}{|c|}{\begin{tabular}[c]{@{}c@{}}Learning to drive \\ from \cite{chen2021learningc136},\\ 2021\end{tabular}} &
  \multicolumn{1}{c|}{CARLA} &
  \multicolumn{1}{c|}{\begin{tabular}[c]{@{}c@{}}RGB images\\ and \\ speed readings\end{tabular}} &
  \multicolumn{1}{c|}{\begin{tabular}[c]{@{}c@{}}Steering, \\ throttle and \\ brake\end{tabular}} &
  \multicolumn{1}{c|}{\begin{tabular}[c]{@{}c@{}}Reinforcement \\ learning, policy \\ distillation\end{tabular}} &
  \multicolumn{1}{c|}{\begin{tabular}[c]{@{}c@{}}CARLA \\ LeaderBoard\end{tabular}} &
  \multicolumn{1}{c|}{-} &
  \multicolumn{1}{c|}{\begin{tabular}[c]{@{}c@{}}Action-values \\ based on the \\ current ego-vehicle \\ state\end{tabular}} &
  \multicolumn{1}{c|}{\begin{tabular}[c]{@{}c@{}}DS: 17.36$\pm$2.95 \\ RC: 43.46$\pm$2.99 \\ IS: 0.54$\pm$0.06\end{tabular}} &
  \begin{tabular}[c]{@{}c@{}}Train: 69 hours \\ about 1m frames\\ Test: 270K frames\end{tabular} \\ \hline
\multicolumn{1}{|c|}{\begin{tabular}[c]{@{}c@{}}Safe Local Motion \\ Planning with \\ Self-Supervised \cite{Hu_2021_CVPR},\\ 2021\end{tabular}} &
  \multicolumn{1}{c|}{\begin{tabular}[c]{@{}c@{}}NuScenes,\\ CARLA\end{tabular}} &
  \multicolumn{1}{c|}{LiDAR} &
  \multicolumn{1}{c|}{\begin{tabular}[c]{@{}c@{}}Control signal,\\ BEV\end{tabular}} &
  \multicolumn{1}{c|}{\begin{tabular}[c]{@{}c@{}}Behaviour \\ cloning\end{tabular}} &
  \multicolumn{1}{c|}{CARLA NoCrash} &
  \multicolumn{1}{c|}{\begin{tabular}[c]{@{}c@{}}Object-centric\\ representation\end{tabular}} &
  \multicolumn{1}{c|}{\begin{tabular}[c]{@{}c@{}}Safe planner, maintaining \\ a wide safety margin, avoid\\ safety critical situations\end{tabular}} &
  \multicolumn{1}{c|}{\begin{tabular}[c]{@{}c@{}}(Success rate)\\ NC-E:66 $\pm$ 3\\ NC-R: 73$\pm$ 1\\ NC-D: 44 $\pm$ 5\end{tabular}} &
  \begin{tabular}[c]{@{}c@{}}Train:  Town 1 \\ Test:  Town 2 \\ Train: 850 scenes(NuScenes)\\ Test: 150 scenes\end{tabular} \\ \hline
\multicolumn{1}{|c|}{\begin{tabular}[c]{@{}c@{}}Carl-Lead: Lidar-based \\ End-to-End \\ Autonomous \cite{Cai2021CarlLeadLE},\\ 2021\end{tabular}} &
  \multicolumn{1}{c|}{CARLA} &
  \multicolumn{1}{c|}{LiDAR} &
  \multicolumn{1}{c|}{\begin{tabular}[c]{@{}c@{}}Steering, \\ throttle, brake, \\ HD map\end{tabular}} &
  \multicolumn{1}{c|}{\begin{tabular}[c]{@{}c@{}}Reinforcement\\ learning\end{tabular}} &
  \multicolumn{1}{c|}{CARLA NoCrash} &
  \multicolumn{1}{c|}{\begin{tabular}[c]{@{}c@{}}Saliency maps for\\ visualizing \\ model predictions\end{tabular}} &
  \multicolumn{1}{c|}{\begin{tabular}[c]{@{}c@{}}Collision reward \\ punishment for unsafe \\ driving\end{tabular}} &
  \multicolumn{1}{c|}{\begin{tabular}[c]{@{}c@{}}(Success rate)\\ NC-R: 93.50\\ \\ NC-D: 93.00\end{tabular}} &
  \begin{tabular}[c]{@{}c@{}}Train: 677.7K interaction steps\\ Test: 14,400 episodes\end{tabular} \\ \hline
\multicolumn{1}{|c|}{\begin{tabular}[c]{@{}c@{}}A Versatile and \\ Efficient \\ Reinforcement Learning \cite{wang2021versatile},\\ 2021\end{tabular}} &
  \multicolumn{1}{c|}{\begin{tabular}[c]{@{}c@{}}CARLA,\\ BDD100k\end{tabular}} &
  \multicolumn{1}{c|}{RGB image} &
  \multicolumn{1}{c|}{Control signal} &
  \multicolumn{1}{c|}{\begin{tabular}[c]{@{}c@{}}Reinforcement \\ learning,\\ imitation learning\end{tabular}} &
  \multicolumn{1}{c|}{NoGap on CARLA} &
  \multicolumn{1}{c|}{\begin{tabular}[c]{@{}c@{}}Interpretable \\ segmentation\\ map\end{tabular}} &
  \multicolumn{1}{c|}{\begin{tabular}[c]{@{}c@{}}Move the vehicle \\ to safe state\end{tabular}} &
  \multicolumn{1}{c|}{\begin{tabular}[c]{@{}c@{}}MPI (m): Turn:7.2,\\ Straight : 4.1 \\ SR: Turn: 53.2 \\ Straight: 16.7\\ Close loop : MPI:\\ RL: 332.6, IL:180.9\end{tabular}} &
  \begin{tabular}[c]{@{}c@{}}Train: 28h of driving \\ and 2.5M RL steps\end{tabular} \\ \hline
\multicolumn{1}{|c|}{\begin{tabular}[c]{@{}c@{}}Multi-task Learning \\ with  Attention \\ for End-to-end \cite{ishihara2021multi},\\ 2021\end{tabular}} &
  \multicolumn{1}{c|}{CARLA} &
  \multicolumn{1}{c|}{\begin{tabular}[c]{@{}c@{}}Monocular RGB,\\ velocity\end{tabular}} &
  \multicolumn{1}{c|}{\begin{tabular}[c]{@{}c@{}}Steering, \\ throttle and \\ brake\end{tabular}} &
  \multicolumn{1}{c|}{\begin{tabular}[c]{@{}c@{}}Conditional \\ imitation learning,\\ multitask learning\end{tabular}} &
  \multicolumn{1}{c|}{\begin{tabular}[c]{@{}c@{}}CARLA NoCrash ,\\ CoRL2017 \\ benchmark\end{tabular}} &
  \multicolumn{1}{c|}{-} &
  \multicolumn{1}{c|}{-} &
  \multicolumn{1}{c|}{\begin{tabular}[c]{@{}c@{}}Straight :99 $\pm$ 1\\ One Turn: 99 $\pm$ 1\\ NC-E:  81$\pm$ 11\\ NC-R :67 $\pm$ 9\\ NC-D:23 $\pm$ 5\end{tabular}} &
  \begin{tabular}[c]{@{}c@{}}Train: Town01 \\ (466,000 frames)\\ Test: Town02\end{tabular} \\ \hline
\multicolumn{1}{|c|}{\begin{tabular}[c]{@{}c@{}}End-to-End Model-\\ Free Reinforcement \cite{toromanoff2020endc66},\\ 2020\end{tabular}} &
  \multicolumn{1}{c|}{CARLA} &
  \multicolumn{1}{c|}{\begin{tabular}[c]{@{}c@{}}Front-facing \\ camera , \\ speed\end{tabular}} &
  \multicolumn{1}{c|}{\begin{tabular}[c]{@{}c@{}}5 steer actions, \\ 3 values for  \\ throttle,\\ one for brake\end{tabular}} &
  \multicolumn{1}{c|}{\begin{tabular}[c]{@{}c@{}}Reinforcement \\ learning\end{tabular}} &
  \multicolumn{1}{c|}{\begin{tabular}[c]{@{}c@{}}CARLA \\ NoCrash, \\ LeaderBoard\end{tabular}} &
  \multicolumn{1}{c|}{\begin{tabular}[c]{@{}c@{}}Coin implicit \\ affordances\end{tabular}} &
  \multicolumn{1}{c|}{-} &
  \multicolumn{1}{c|}{\begin{tabular}[c]{@{}c@{}}NC-R: 96\\ NC-D:70\\ NC-E: 100\end{tabular}} &
  \begin{tabular}[c]{@{}c@{}}Train: 20M \\ iterations Town05\\ Test: Town02\end{tabular} \\ \hline
\multicolumn{1}{|c|}{\begin{tabular}[c]{@{}c@{}}Learning by \\ cheating \cite{chen2020learningc60},\\ 2020\end{tabular}} &
  \multicolumn{1}{c|}{CARLA} &
  \multicolumn{1}{c|}{\begin{tabular}[c]{@{}c@{}}Front-facing \\ camera , speed,\\ navigation \\ command\end{tabular}} &
  \multicolumn{1}{c|}{\begin{tabular}[c]{@{}c@{}}Steering, \\ throttle and \\ brake\end{tabular}} &
  \multicolumn{1}{c|}{\begin{tabular}[c]{@{}c@{}}On-policy\\ imitation\\ learning\end{tabular}} &
  \multicolumn{1}{c|}{\begin{tabular}[c]{@{}c@{}}CARLA \\ NoCrash \\ and\\ LeaderBoard\end{tabular}} &
  \multicolumn{1}{c|}{Map  representation} &
  \multicolumn{1}{c|}{\begin{tabular}[c]{@{}c@{}}Conduct a \\ separate \\ infraction analysis\end{tabular}} &
  \multicolumn{1}{c|}{\begin{tabular}[c]{@{}c@{}}NC-E: 100\\ NC-R: 94$\pm$4\\ NC-D: 85$\pm$1\end{tabular}} &
  \begin{tabular}[c]{@{}c@{}}DAgger training.\\ Train: 157K frames, \\ 4 hours driving Town1, \\ Test: Town2\end{tabular} \\ \hline
  \multicolumn{1}{|c|}{\begin{tabular}[c]{@{}c@{}}Learning Situational \\ Driving \cite{ohn2020learningc62},\\ 2020\end{tabular}} &
  \multicolumn{1}{c|}{CARLA} &
  \multicolumn{1}{c|}{\begin{tabular}[c]{@{}c@{}}Front-facing camera , \\ speed, navigation \\ command\end{tabular}} &
  \multicolumn{1}{c|}{\begin{tabular}[c]{@{}c@{}}Longitudinal \\ and lateral \\ control values\end{tabular}} &
  \multicolumn{1}{c|}{\begin{tabular}[c]{@{}c@{}}Behaviour \\ cloning\end{tabular}} &
  \multicolumn{1}{c|}{\begin{tabular}[c]{@{}c@{}}CARLA \\ NoCrash,\\ LeaderBoard\end{tabular}} &
  \multicolumn{1}{c|}{\begin{tabular}[c]{@{}c@{}}Situation-specific \\ predictions can be \\ inspected \\ at test time\end{tabular}} &
  \multicolumn{1}{c|}{\begin{tabular}[c]{@{}c@{}}Learned agent to adhere \\ to traffic rules and \\ safety\end{tabular}} &
  \multicolumn{1}{c|}{\begin{tabular}[c]{@{}c@{}}NC-R: 64\\ NC-D: 32\end{tabular}} &
  \begin{tabular}[c]{@{}c@{}}Train: Town1\\ Test: Town2\end{tabular} \\ \hline
  
\multicolumn{1}{|c|}{\begin{tabular}[c]{@{}c@{}}SAM \cite{zhao2021samc65},\\ 2020\end{tabular}} &
  \multicolumn{1}{c|}{CARLA} &
  \multicolumn{1}{c|}{\begin{tabular}[c]{@{}c@{}}Image,\\ self-speed,\\ turning command\end{tabular}} &
  \multicolumn{1}{c|}{\begin{tabular}[c]{@{}c@{}}Brake, \\ gas, and \\ steering angle\end{tabular}} &
  \multicolumn{1}{c|}{\begin{tabular}[c]{@{}c@{}}Conditional\\ imitation\\  learning\end{tabular}} &
  \multicolumn{1}{c|}{\begin{tabular}[c]{@{}c@{}}Traffic-school \\ benchmark,\\ CARLA NoCrash\end{tabular}} &
  \multicolumn{1}{c|}{-} &
  \multicolumn{1}{c|}{\begin{tabular}[c]{@{}c@{}}Stop intentions  help \\ avoid \\ hazardous traffic\\ situations\end{tabular}} &
  \multicolumn{1}{c|}{\begin{tabular}[c]{@{}c@{}}NC-E: 83$\pm$1\\ NC-R: 68$\pm$7\\ NC-D: 29$\pm$2\end{tabular}} &
  \begin{tabular}[c]{@{}c@{}}Train : 10 hours \\ (360K frames) Town01.\end{tabular} \\ \hline
\multicolumn{1}{|c|}{\begin{tabular}[c]{@{}c@{}}Multi-task Learning with \\ Future States \\ for Vision \cite{kim2020multi}, \\ 2020\end{tabular}} &
  \multicolumn{1}{c|}{CARLA} &
  \multicolumn{1}{c|}{\begin{tabular}[c]{@{}c@{}}Three RGB \\ cameras\end{tabular}} &
  \multicolumn{1}{c|}{Control signal} &
  \multicolumn{1}{c|}{\begin{tabular}[c]{@{}c@{}}Multi-task learning,\\ conditional \\ imitation learning\end{tabular}} &
  \multicolumn{1}{c|}{\begin{tabular}[c]{@{}c@{}}NoCrash,\\ AnyWeather\end{tabular}} &
  \multicolumn{1}{c|}{-} &
  \multicolumn{1}{c|}{\begin{tabular}[c]{@{}c@{}}Localization tasks is \\ useful for safe \\ driving.\end{tabular}} &
  \multicolumn{1}{c|}{\begin{tabular}[c]{@{}c@{}}NC-E: 92,\\ NC-R 66,\\ NC-D:  32, \\ SR: 93.2\end{tabular}} &
  Train: 100 hours \\ \hline

  \multicolumn{1}{|c|}{\begin{tabular}[c]{@{}c@{}}DSDNet \cite{Zeng2020DSDNetDS},\\ 2020\end{tabular}} &
  \multicolumn{1}{|c|}{\begin{tabular}[c]{@{}c@{}}NuScenes,\\ ATG4D,\\ CARLA\end{tabular}} &
  \multicolumn{1}{|c|}{\begin{tabular}[c]{@{}c@{}}LiDAR,\\ HD map\end{tabular}} &
  Steering, speed &
  \multicolumn{1}{|c|}{\begin{tabular}[c]{@{}c@{}}Imitation \\ learning\end{tabular}} &
  \multicolumn{1}{|c|}{\begin{tabular}[c]{@{}c@{}}CR, L2,\\ CARLA\end{tabular}} &
  \multicolumn{1}{|c|}{\begin{tabular}[c]{@{}c@{}}Learn interpretable \\ intermediate \\ results\end{tabular}} &
  \multicolumn{1}{|c|}{\begin{tabular}[c]{@{}c@{}}Safer planning by \\ cost function\end{tabular}} &
  \multicolumn{1}{|c|}{\begin{tabular}[c]{@{}c@{}}L2: 1.22, \\ Lane vio: 1.55\\ IOU: 55.4\\ Min MSD: 0.213\end{tabular}} &
  \multicolumn{1}{|c|}{\begin{tabular}[c]{@{}c@{}}Train: 60K, 1000, 5000 samples \\ Test: 17K, 500 samples\end{tabular}} \\ \hline
\multicolumn{1}{|c|}{\begin{tabular}[c]{@{}c@{}}Perceive, Predict, and\\  Plan: Safe Motion \cite{sadat2020perceive},\\ 2020\end{tabular}} &
  Real scenario &
 \multicolumn{1}{|c|}{ \begin{tabular}[c]{@{}c@{}}Raw sensor data,\\ HD map,\\ high level route\end{tabular}}&
  Control action &
  \multicolumn{1}{|c|}{\begin{tabular}[c]{@{}c@{}}Imitation learning,\\ inverse RL\end{tabular}} &
  \multicolumn{1}{|c|}{\begin{tabular}[c]{@{}c@{}}L2, CR,\\ jerk and \\ lateral acceleration\end{tabular}} &
  \multicolumn{1}{|c|}{\begin{tabular}[c]{@{}c@{}}Interpretable \\ Semantic \\ Representations\\ (occupancy map)\end{tabular}} &
  \multicolumn{1}{|c|}{\begin{tabular}[c]{@{}c@{}}Planner learn \\ safety cost,\\ safety buffer\end{tabular}} &
  \multicolumn{1}{|c|}{\begin{tabular}[c]{@{}c@{}}CR: 1.78 \\ L2:3.34\\ Jerk: 1.27 \\ Lat. acc: 2.89\end{tabular}} &
  \multicolumn{1}{|c|}{\begin{tabular}[c]{@{}c@{}}Train: 6100 scenarios \\ Test: 1500 scenarios.\end{tabular}} \\ \hline

\multicolumn{1}{|c|}{\begin{tabular}[c]{@{}c@{}}Urban driving \\ with condition \\ imitation learning \cite{hawke2020urban},\\ 2020\end{tabular}} &
  \multicolumn{1}{c|}{Real scenario} &
  \multicolumn{1}{c|}{\begin{tabular}[c]{@{}c@{}}Camera , \\ navigation \\ command\end{tabular}} &
  \multicolumn{1}{c|}{Steering, speed} &
  \multicolumn{1}{c|}{\begin{tabular}[c]{@{}c@{}}Imitation \\ learning\end{tabular}} &
  \multicolumn{1}{c|}{\begin{tabular}[c]{@{}c@{}}Successful  turns, \\ CR,TV\end{tabular}} &
  \multicolumn{1}{c|}{\begin{tabular}[c]{@{}c@{}}Using pre trained \\ perception\end{tabular}} &
  \multicolumn{1}{c|}{Safety-driver in loop} &
  \multicolumn{1}{c|}{MAE: 0.0715} &
  \begin{tabular}[c]{@{}c@{}}Train: 30h\\ Test: 26 routes\end{tabular} \\ \hline  
\multicolumn{1}{|c|}{\begin{tabular}[c]{@{}c@{}}Exploring the \\ Limitations of Behavior \cite{codevilla2019exploringc61},\\ 2019\end{tabular}} &
  \multicolumn{1}{c|}{CARLA} &
  \multicolumn{1}{c|}{\begin{tabular}[c]{@{}c@{}}Image,\\ self-speed,\\ turning command\end{tabular}} &
  \multicolumn{1}{c|}{\begin{tabular}[c]{@{}c@{}}Waypoints, \\ steer, throttle, \\ and brake\end{tabular}} &
  \multicolumn{1}{c|}{\begin{tabular}[c]{@{}c@{}}Behaviour \\ cloning\end{tabular}} &
  \multicolumn{1}{c|}{\begin{tabular}[c]{@{}c@{}}CARLA \\ NoCrash,\\ LeaderBoard\end{tabular}} &
  \multicolumn{1}{c|}{-} &
  \multicolumn{1}{c|}{\begin{tabular}[c]{@{}c@{}}Potentially inconsistent \\ put the \\ vehicle back \\ to a safe state\end{tabular}} &
  \multicolumn{1}{c|}{\begin{tabular}[c]{@{}c@{}}NC-E: 90$\pm$2\\ NC-R: 56$\pm$2\\ NC-D: 24$\pm$8\end{tabular}} &
  \begin{tabular}[c]{@{}c@{}}Train: 100 hours dataset\\ Test: 80 hours dataset\end{tabular} \\ \hline
\multicolumn{1}{|c|}{\begin{tabular}[c]{@{}c@{}}Learning \\ to drive from \\ from
simulation without \cite{bewley2019learning},\\ 2019\end{tabular}} &
  \multicolumn{1}{c|}{\begin{tabular}[c]{@{}c@{}}Simulation\\ + Real \\ scenario\end{tabular}} &
  \multicolumn{1}{c|}{\begin{tabular}[c]{@{}c@{}}Single frontal\\ camera\end{tabular}} &
  \multicolumn{1}{c|}{Steering} &
  \multicolumn{1}{c|}{\begin{tabular}[c]{@{}c@{}}Imitation \\ learning\end{tabular}} &
  \multicolumn{1}{c|}{\begin{tabular}[c]{@{}c@{}}Average distance\\ per intervention\end{tabular}} &
  \multicolumn{1}{c|}{\begin{tabular}[c]{@{}c@{}}Bi-directional \\ image translator\end{tabular}} &
  \multicolumn{1}{c|}{-} &
  \multicolumn{1}{c|}{\begin{tabular}[c]{@{}c@{}}Sim MAE: 0.017\\ Real MAE: 0.081\end{tabular}} &
  \begin{tabular}[c]{@{}c@{}}Train and Test: \\ 60K frames\end{tabular} \\ \hline

\multicolumn{1}{|c|}{\begin{tabular}[c]{@{}c@{}}Learning accurate,\\ comfortable \\and
human-like driving\cite{hecker2019learning},\\ 2019\end{tabular}} &
  \multicolumn{1}{c|}{Real scenario} &
  \multicolumn{1}{c|}{\begin{tabular}[c]{@{}c@{}}Single frontal\\ camera\end{tabular}} &
  \multicolumn{1}{c|}{Steering, speed} &
  \multicolumn{1}{c|}{\begin{tabular}[c]{@{}c@{}}Imitation \\ learning\end{tabular}} &
  \multicolumn{1}{c|}{\begin{tabular}[c]{@{}c@{}}L1, comfort measure, \\ driving accuracy, \\ human-likeness \\ score\end{tabular}} &
  \multicolumn{1}{c|}{HERE map} &
  \multicolumn{1}{c|}{-} &
  \multicolumn{1}{c|}{\begin{tabular}[c]{@{}c@{}}AS: 7.96\\ HL: 29.3\end{tabular}} &
  \begin{tabular}[c]{@{}c@{}}Train: 60h,\\ Test: 10h\end{tabular} \\ \hline
\multicolumn{1}{|c|}{\begin{tabular}[c]{@{}c@{}}Learning to \\ drive in a day \cite{8793742},\\ 2019\end{tabular}} &
  \multicolumn{1}{c|}{\begin{tabular}[c]{@{}c@{}}Real +\\ Unreal Engine 4\end{tabular}} &
  \multicolumn{1}{c|}{\begin{tabular}[c]{@{}c@{}}Single frontal\\ camera\end{tabular}} &
  \multicolumn{1}{c|}{\begin{tabular}[c]{@{}c@{}}Steering, \\ throttle and \\ brake\end{tabular}} &
  \multicolumn{1}{c|}{\begin{tabular}[c]{@{}c@{}}Reinforcement \\ learning\end{tabular}} &
  \multicolumn{1}{c|}{\begin{tabular}[c]{@{}c@{}}Distance travel\\ reward\end{tabular}} &
  \multicolumn{1}{c|}{-} &
  \multicolumn{1}{c|}{Safer reward function} &
  \multicolumn{1}{c|}{\begin{tabular}[c]{@{}c@{}}Meter/Disengagement\\ (250m): 0\end{tabular}} &
  \begin{tabular}[c]{@{}c@{}}Train: 10 episode\\ Test : 250 meters\end{tabular} \\ \hline
\multicolumn{1}{|c|}{\begin{tabular}[c]{@{}c@{}}Multimodal \\ end-to-end \\autonomous driving \cite{xiao2020multimodalc118},\\ 2019\end{tabular}} &
  \multicolumn{1}{c|}{CARLA} &
  \multicolumn{1}{c|}{\begin{tabular}[c]{@{}c@{}}Front-facing \\ camera and\\ LiDAR\end{tabular}} &
  \multicolumn{1}{c|}{\begin{tabular}[c]{@{}c@{}}Steering, \\ throttle and \\ brake\end{tabular}} &
  \multicolumn{1}{c|}{\begin{tabular}[c]{@{}c@{}}Conditional\\ imitation\\  learning\end{tabular}} &
  \multicolumn{1}{c|}{CARLA} &
  \multicolumn{1}{c|}{-} &
  \multicolumn{1}{c|}{-} &
  \multicolumn{1}{c|}{SR: 94} &
  \begin{tabular}[c]{@{}c@{}}Train: 25h\\ Test : Town 1 and 2\end{tabular} \\ \hline
  \multicolumn{1}{|c|}{\begin{tabular}[c]{@{}c@{}}End-to-end \\ interpretable neural \\ motion planner\cite{zeng2019end},\\ 2019\end{tabular}} &
  \multicolumn{1}{c|}{Real scenario} &
  \multicolumn{1}{c|}{\begin{tabular}[c]{@{}c@{}}Front-facing \\ camera , \\ speed.\end{tabular}} &
  \multicolumn{1}{c|}{\begin{tabular}[c]{@{}c@{}}Cost volume,\\ future trajectories,\\ Object location\end{tabular}} &
  \multicolumn{1}{c|}{\begin{tabular}[c]{@{}c@{}}Imitation \\ learning\end{tabular}} &
  \multicolumn{1}{c|}{\begin{tabular}[c]{@{}c@{}}Collision rate,\\ traffic violation rate\end{tabular}} &
  \multicolumn{1}{c|}{\begin{tabular}[c]{@{}c@{}}Interpretable \\ intermediate representations\end{tabular}} &
  \multicolumn{1}{c|}{\begin{tabular}[c]{@{}c@{}}Cost volume can \\ generate safer planning\end{tabular}} &
  \multicolumn{1}{c|}{\begin{tabular}[c]{@{}c@{}}L2(3sec): 2.353\\ Colli rate: 0.78\end{tabular}} &
  \begin{tabular}[c]{@{}c@{}}Train : 50000 scenarios\\ Val: 500 scenarios\\ Test: 1000 scenarios\end{tabular} \\ \hline
\end{longtable}

\begin{tablenotes}
\item Route Completion (RC), Infraction Score/penalty (IS), Driving score (DS), Collisions pedestrians (CP)/(PC), Collisions vehicles (CV), Collisions layout (CL)/(LC), Red light infractions (RLI), Red light violation (RV), Stop sign infractions (SSI), Off-road infractions (OI), Route deviations (RD), Agent blocked (AB).
\item Average Displacement Error (ADE), Final Displacement Error (FDE), Intersection over Union (IOU), Panoptic Quality (PQ), Recognition Quality (RQ), Segmentation Quality (SQ), Instance Contrastive Pair (ICP), Action Contrastive Pair (ACP), Driving in Occlusion Simulation (DOS), Episodic Cost (EC), Success Rate (SR).
\item Collision Rate (CR), Safety Violation (SV), Episodic Return (ER), Episodic Cost (EC), Vehicle Avoidance (VA), Pedestrian Avoidance (PA), Meters Per Intervention(MPI), Reinforcement Learning (RL), Imitation Learning (IL), NoCrash Regular Traffic (NC-R), NoCrash Dense Traffic (NC-D), NoCrash Empty (NC-E) .

\end{tablenotes}
\end{landscape}
\clearpage
\twocolumn

augments real point clouds with artificial obstacles by blending them appropriately into the surroundings. Regarding planning and decision disparity, Pan et al. \cite{pan2017virtual152} propose learning driving policies in a simulated setting with realistic frames before applying them in the real world. Osinski et al. \cite{osiński2020simulationbased160} propose a driving policy using a simulator, where a segmentation network is developed using annotated real-world data, while the driving controller is learned using synthetic images and their semantics. Mitchell et al. \cite{mitchell2020multivehicle165} and Stocco et al. \cite{Stocco_2023166} enable robust online policy learning adaptation through a mixed-reality arrangement, which includes an actual vehicle and other virtual cars and obstacles, allowing the real car to learn from simulated collisions and test scenarios.

\begin{figure*}
  \centering

  \begin{tabular}{@{}c@{}}
    \includegraphics[width=.8\linewidth,height=130pt]{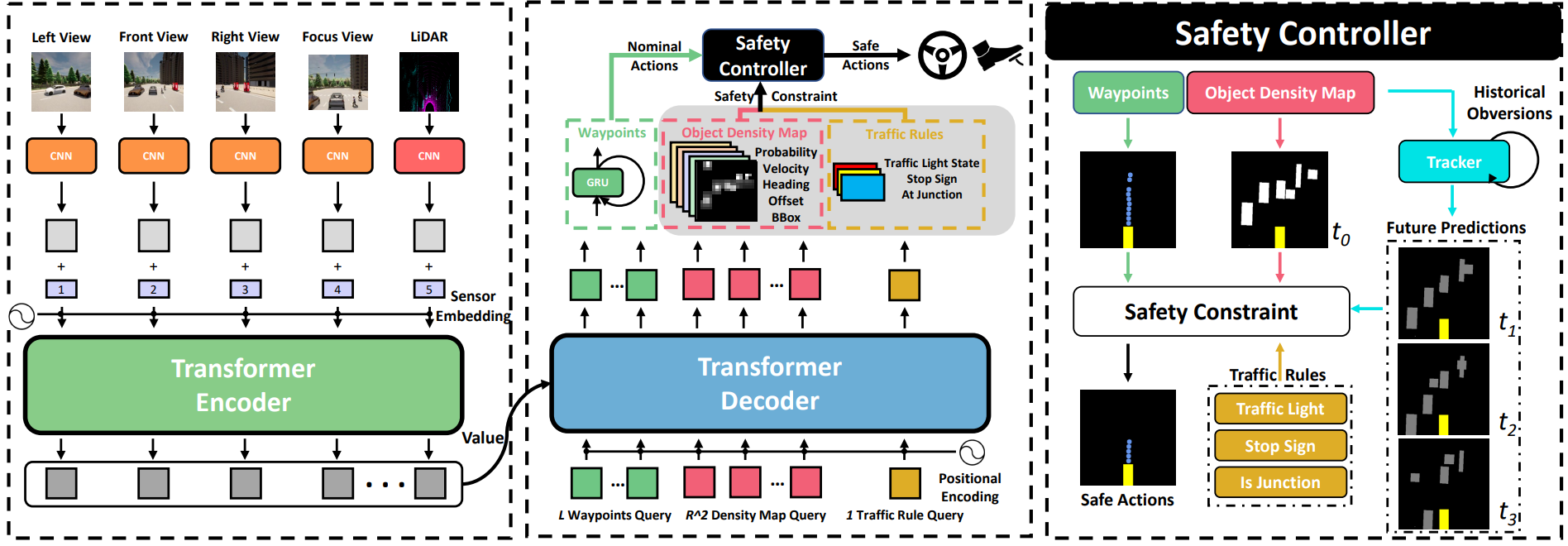} \\[\abovecaptionskip]
    \small (a) InterFuser 
  \end{tabular}
  \begin{tabular}{@{}c@{}}
    \includegraphics[width=.45\linewidth,height=100pt]{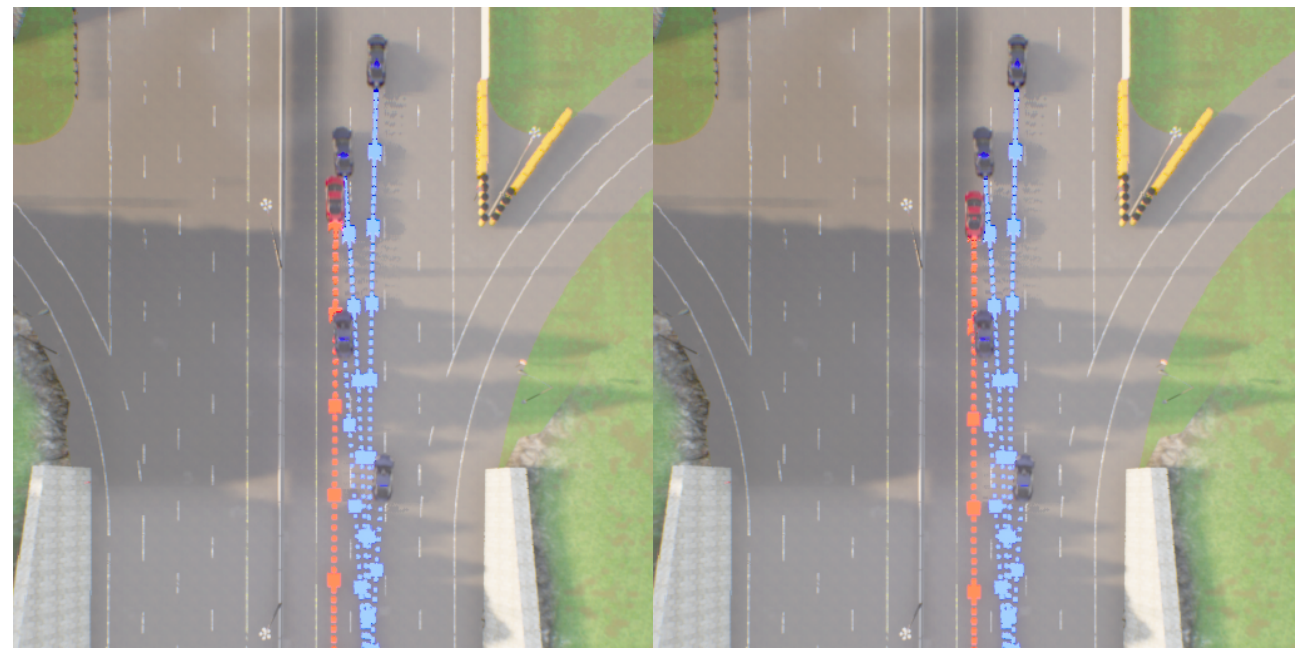} \\[\abovecaptionskip]
    \small (b) KING Lane Merger Scenario 
  \end{tabular}
  \begin{tabular}{@{}c@{}}
    \includegraphics[width=.45\linewidth,height=100pt]{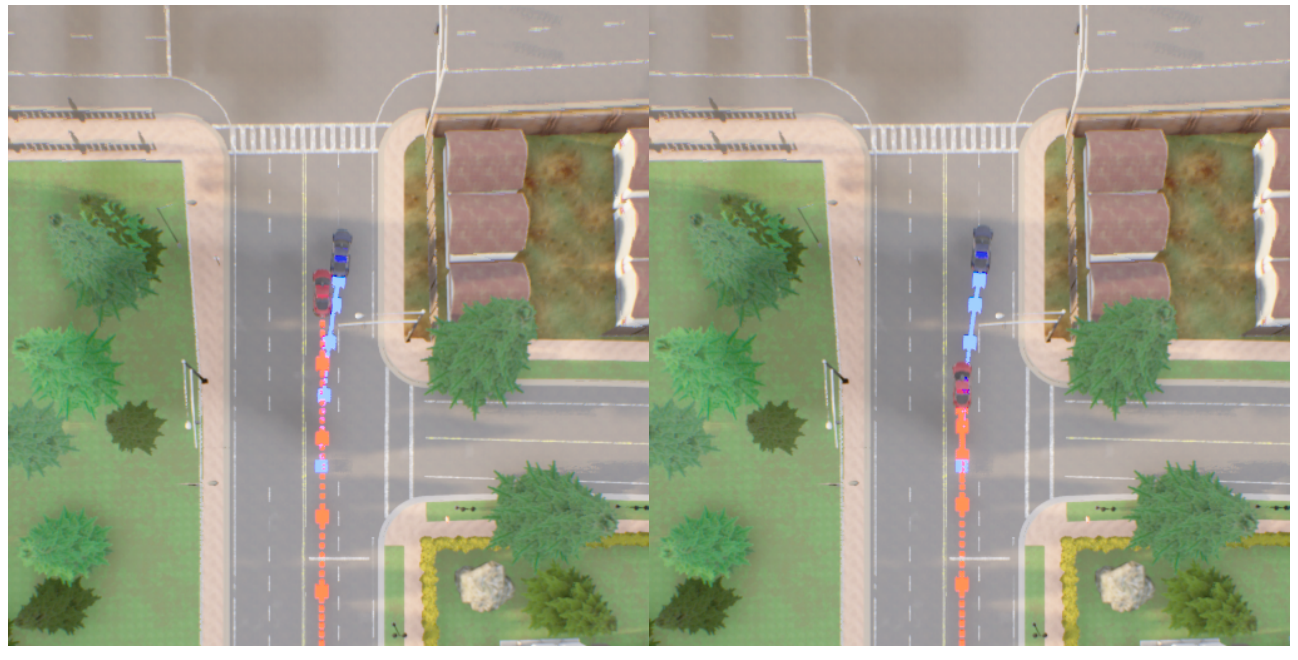} \\[\abovecaptionskip]
    \small (c) KING Collision Avoidance 
  \end{tabular}

  \caption{Demonstration of safe driving methods: (a) InterFuser \cite{shao2023safetyc117} processes multisensorial information to detect adversarial events, which are then used by the controller to constrain driving actions within safe sets. (b) KING \cite{hanselmann2022king2orc99} improves collision avoidance using scenario generation. The image shows the ego vehicle (shown in red) maintaining a safe distance during a lane merge in the presence of an adversarial agent (shown in blue). (c) In the same context, the image illustrates the vehicle slowing down to avoid collision.}\label{safe}
  
\end{figure*}

\section{Safety} \label{safety}

\begin{table*}[]

\caption{END-TO-END DRIVING TESTING TO ENSURE SAFETY}
\label{testing0}
\begin{tabular}{|ccc|}
\hline
\rowcolor[HTML]{C0C0C0} 
Methods &
  Summary &
  Literature \\ \hline
\multicolumn{1}{|c|}{} &
  \multicolumn{1}{c|}{\begin{tabular}[c]{@{}c@{}}Generating neuron coverage to \\ identify false actions\end{tabular}} &
  \cite{tian2018deeptest} \\ \cline{2-3} 
\multicolumn{1}{|c|}{} &
  \multicolumn{1}{c|}{\begin{tabular}[c]{@{}c@{}}Designing an diverse and critical \\ unsafe test cases\end{tabular}} &
  \cite{hanselmann2022king2orc99} \\ \cline{2-3} 
\multicolumn{1}{|c|}{\multirow{-3}{*}{\begin{tabular}[c]{@{}c@{}}Search-based \\ testing \\ \end{tabular}}} &
  \multicolumn{1}{c|}{\begin{tabular}[c]{@{}c@{}}Objective function to search safety \\ sensitive output\end{tabular}} &
  \cite{zeng2021endtoend} \\ \hline
\multicolumn{1}{|c|}{} &
  \multicolumn{1}{c|}{\begin{tabular}[c]{@{}c@{}}Place the original object with an \\ adversarial one\end{tabular}} &
  \cite{ZhangWMWHS20130} \\ \cline{2-3} 
\multicolumn{1}{|c|}{\multirow{-2}{*}{\begin{tabular}[c]{@{}c@{}}Optimization-\\ based attack\end{tabular}}} &
  \multicolumn{1}{c|}{\begin{tabular}[c]{@{}c@{}}Virtual obstacles to generate \\ adversarial attack in natural environment\end{tabular}} &
  \cite{wu2023adversarial} \\ \hline
\multicolumn{1}{|c|}{} &
  \multicolumn{1}{c|}{\begin{tabular}[c]{@{}c@{}}Generate the adversarial realistic-looking \\ representations based on images\end{tabular}} &
  \cite{goodfellow2020generative} \\ \cline{2-3} 
\multicolumn{1}{|c|}{} &
  \multicolumn{1}{c|}{\begin{tabular}[c]{@{}c@{}}Generate pedestrian augmentation from \\ inserting pedestrians in image\end{tabular}} &
  \cite{ouyang2018pedestriansynthesisgan138} \\ \cline{2-3} 
\multicolumn{1}{|c|}{\multirow{-3}{*}{\begin{tabular}[c]{@{}c@{}}GAN-based \\ attack\end{tabular}}} &
  \multicolumn{1}{c|}{\begin{tabular}[c]{@{}c@{}}Designing an objective function to search \\ for the diverse unsafe test cases\end{tabular}} &
  \cite{shao2023safetyc117} \\ \hline
\end{tabular}
\end{table*}

\begin{table*}[]
\caption{END-TO-END TESTING ORACLE MEASURES CORRECT CONTROL DECISION AT DIFFERENT SCENARIOS}

\label{testing1}
\begin{tabular}{|ccc|}
\hline
\rowcolor[HTML]{C0C0C0} 
Test Oracle &
  Detail &
  Literature \\ \hline
\multicolumn{1}{|c|}{\begin{tabular}[c]{@{}c@{}}Metamorphic\\ testing\end{tabular}} &
  \multicolumn{1}{c|}{\begin{tabular}[c]{@{}c@{}}The control signal should not get alter\\ in different condition\end{tabular}} &
  \cite{hanselmann2022king2orc99} \\ \hline
\multicolumn{1}{|c|}{\begin{tabular}[c]{@{}c@{}}Differential\\ testing\end{tabular}} &
  \multicolumn{1}{c|}{\begin{tabular}[c]{@{}c@{}}The End-to-End system must give the same\\safe control for same scenario\end{tabular}} &
  \cite{zhao2021samc65} \\ \hline
\multicolumn{1}{|c|}{\begin{tabular}[c]{@{}c@{}}Model-based\\ oracle\end{tabular}} &
  \multicolumn{1}{c|}{\begin{tabular}[c]{@{}c@{}}Predicting the critical scenario that\\ cause system failure\end{tabular}} &
  \cite{codevilla2019exploringc61} \\ \hline
\end{tabular}
\end{table*}

\begin{table*}[]
\centering

\caption{POPULAR SAFETY METRICS USED FOR SAFETY EVALUATION OF DRIVING SYSTEM}
\label{metrices}
\begin{tabular}{|cccc|}
\hline
\rowcolor[HTML]{C0C0C0} 
Classification &
  Critical Metrics &
  Literature &
  Description \\ \hline
\multicolumn{1}{|c|}{} &
  \multicolumn{1}{c|}{Time to Collision (TTC)} &
  \multicolumn{1}{c|}{ \cite{weng2020model}} &
  \begin{tabular}[c]{@{}c@{}}It defines the minimum time interval that the two agents \\ will collide\end{tabular} \\ \cline{2-4} 
\multicolumn{1}{|c|}{} &
  \multicolumn{1}{c|}{Worst Time to Collision (WTTC)} &
  \multicolumn{1}{c|}{\cite{wachenfeld2016worst}} &
  \begin{tabular}[c]{@{}c@{}}The WTTC metric is an extension of the traditional TTC that \\ takes numerous traces of actors into consideration\end{tabular} \\ \cline{2-4} 
\multicolumn{1}{|c|}{} &
  \multicolumn{1}{c|}{Time to Maneuver (TTM)} &
  \multicolumn{1}{c|}{ \cite{weng2020model}} &
  \begin{tabular}[c]{@{}c@{}}The TTM yields the latest time in the range [0, TTC] at which an \\ expert actor may conduct a movement that avoids a collision\end{tabular} \\ \cline{2-4} 
\multicolumn{1}{|c|}{} &
  \multicolumn{1}{c|}{Time to React (TTR)} &
  \multicolumn{1}{c|}{\cite{tian2018deeptest}} &
  \begin{tabular}[c]{@{}c@{}}TTR metric provides an approximation of the latest time before a \\ reaction is necessary\end{tabular} \\ \cline{2-4} 
\multicolumn{1}{|c|}{\multirow{-5}{*}{\begin{tabular}[c]{@{}c@{}}\\ Temporal \\ metrics\\ \\ \\ \\ \\ \end{tabular}}} &
  \multicolumn{1}{c|}{Time Headway (THW)} &
  \multicolumn{1}{c|}{\cite{weng2020model}} &
  \begin{tabular}[c]{@{}c@{}}The THW measure determines the amount of time it will take an \\ actor to get to the location of other vehicle\end{tabular} \\ \hline
\multicolumn{1}{|c|}{} &
  \multicolumn{1}{c|}{Deceleration Safety Time (DST)} &
  \multicolumn{1}{c|}{\cite{li2020make}} &
  \begin{tabular}[c]{@{}c@{}}It calculates the deceleration required to maintain the \\ safe distance\end{tabular} \\ \cline{2-4} 
\multicolumn{1}{|c|}{} &
  \multicolumn{1}{c|}{Stopping Distance (SD)} &
  \multicolumn{1}{c|}{\cite{li2020av}} &
  \begin{tabular}[c]{@{}c@{}}Minimum stopping distance at the time of \\ deceleration\end{tabular} \\ \cline{2-4} 
\multicolumn{1}{|c|}{} &
  \multicolumn{1}{c|}{Crash Potential Index (CPI)} &
  \multicolumn{1}{c|}{\cite{alhajyaseen2015integration}} &
  \begin{tabular}[c]{@{}c@{}}It measures the probability that the vehicle cannot avoid the \\ collision by deceleration\end{tabular} \\ \cline{2-4} 
\multicolumn{1}{|c|}{\multirow{-4}{*}{\begin{tabular}[c]{@{}c@{}} Non-Temporal\\ metrics\\ \\ \\ \\ \\ \end{tabular}}} &
  \multicolumn{1}{c|}{Conflict Index (CI)} &
  \multicolumn{1}{c|}{\cite{alhajyaseen2015integration}} &
  \begin{tabular}[c]{@{}c@{}}It estimates the collision probability and the severity \\ factors \end{tabular} \\ \hline
\end{tabular}
\end{table*}

Ensuring safety in End-to-End autonomous driving systems is a complex challenge. While these systems offer high-performance potential, several considerations and approaches are essential for maintaining safety throughout the pipeline. First, training the system with diverse and high-quality data that covers a wide range of scenarios, including rare and critical situations. Hanselmann et al. \cite{hanselmann2022king2orc99}, Chen et al. \cite{chen2022learning5orc98}, Chitta et al. \cite{chitta2022transfuserc116}, Xiao et al. \cite{prakash2021multic121}, and Ohn-Bar et al. \cite{ohn2020learningc62} demonstrate that training on critical scenarios helps the system learn robust and safe behaviors and prepares it for environmental conditions and potential hazards. These scenarios include unprotected turnings at intersections, pedestrians emerging from occluded regions, aggressive lane-changing, and other safety heuristics, as shown in Fig.~\ref{safe}(b) and Fig.~\ref{safe}(c). Hanselmann et al. \cite{hanselmann2022king2orc99} focus on improving robustness by inducing adversarial scenarios (collision scenarios) and collecting an observation-waypoint dataset using experts, which is then used to fine-tune the policy.

Integrating safety constraints and rules into the End-to-End system is another vital aspect. The system can prioritize safe behavior by incorporating safety considerations during learning or post-processing system outputs. Safety constraints include a safety cost function \cite{cui2021lookout,casas2021mp3,Zeng2020DSDNetDS,sadat2020perceive}, avoiding unsafe maneuvers \cite{zhang2022learning,jia2023think}, and collision avoidance strategies \cite{zhang2023coaching,wu2022trajectoryc120,zhao2021samc65}. Zeng et al. \cite{zeng2021endtoend} define the cost volume responsible for safe planning; Kendall et al. \cite{8793742} propose a practical safety reward function for safety-sensitive outputs. Li et al. \cite{li2022efficient4} and Hu et al. \cite{hu2022st} demonstrate safety by utilizing the Safety Cost function that penalizes jerk, significant acceleration, and safety violations. To avoid unsafe maneuvers, Zhang et al. \cite{zhang2021learning} eliminate unsafe waypoints, and Shao et al. \cite{shao2023safetyc117} introduce InterFuser (Fig. \ref{safe}(a)), which constrains only the actions within the safety set and steers only the safest action. The above constraints ensure that the system operates within predefined safety boundaries.

Implementing additional safety modules and testing mechanisms (Tables \ref{testing0}, \ref{testing1}) enhances the system's safety. Real-time monitoring of the system's behavior allows for detecting abnormalities or deviations from safe operation. Hu et al. \cite{hu2023planningoriented}, Renz et al. \cite{renz2022plantc121c}, Wu et al. \cite{wu2022trajectoryc120}, and Hawke et al. \cite{hawke2020urban} implement a planner that identifies collision-free routes, reduces possible infractions, and compensates for potential failures or inaccuracies. Renz et al. \cite{renz2022plantc121c} use a rule-based expert algorithm for their planner, while Wu et al. \cite{wu2022trajectoryc120} propose a trajectory + control model that predicts a safe trajectory over the long horizon. Hu et al. \cite{hu2023planningoriented} also employ a goal planner to ensure safety. Codevilla et al. \cite{codevilla2019exploringc61} demonstrate the system's ability to respond appropriately and return the vehicle to a safe state when encountering potential inconsistencies. Similarly, Zhao et al. \cite{zhao2021samc65} incorporate stop intentions to help avoid hazardous traffic situations and respond appropriately. These mechanisms ensure that the system can detect and respond to abnormal or unexpected situations, thereby reducing the risk of accidents or unsafe behavior.

Adversarial attack \cite{wu2023adversarial} methods, as shown in the Table \ref{testing0}, are utilized in driving testing to evaluate the correctness of the output control signal. These testing methodologies aim to identify vulnerabilities and assess the robustness against adversaries. The End-to-End testing oracle (Table~\ref{testing1}) determines the correct control decision within a given scenario. Metamorphic testing tackles the oracle problem by verifying the consistency of the steering angle \cite{hanselmann2022king2orc99} across various weather and lighting conditions. It provides a reliable way to ensure that the steering angle remains stable and unaffected by these factors. Differential testing \cite{zhao2021samc65} exposes inconsistencies among different DNN models by comparing their inference results for the same scenario. If the models produce different outcomes, it indicates unexpected behavior and potential issues in the system. The model-based oracle employs a trained probabilistic model to assess and predict potential risks \cite{codevilla2019exploringc61} in real scenarios. By monitoring the environment, it can identify situations that the system may not adequately handle.

Safety metrics provide quantitative measures to evaluate the performance of autonomous driving systems and assess how well the system functions in terms of safety. Time to Collision (TTC), Conflict Index (CI), Crash Potential Index (CPI), Time to React (TTR), and others are some of the metrics that can provide additional objective comparisons between the safety performance of various approaches and identify areas that require improvement. A description of these metrics is provided in Table \ref{metrices}.

\section{Explainability}\label{explainability}

Explainability \cite{rosenfeld2019explainability} refers to the ability to understand the logic of an agent and is focused on how a user interprets the relationships between the input and output of a model. It encompasses two main concepts: interpretability, which relates to the understandability of explanations, and completeness, which pertains to exhaustively defining the behavior of the model through explanations. Choi et al. \cite{choi2015investigating} distinguish three types of confidence in autonomous vehicles: transparency, which refers to the person's ability to foresee and comprehend vehicle operation; technical competence, which relates to understanding vehicle performance; and situation management, which involves the notion that the user can regain vehicle control at any time. According to Haspiel et al. \cite{haspiel2018explanations}, explanations play a crucial role when humans are involved, as the ability to explain an autonomous vehicle's actions significantly impacts consumer trust, which is essential for the widespread acceptance of this technology.

\begin{figure}[t]
\centerline{\includegraphics[scale=0.35]{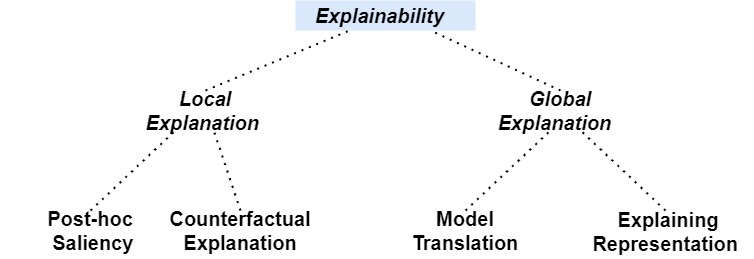}}
\caption{Categorization of Explainability Approaches.}
\label{fig2}
\end{figure}

In the context of explainability for end-to-end autonomous driving systems, we can categorize explanation approaches into two main types (Fig. \ref{fig2}): local explanations and global explanations. A local explanation aims to describe the rationale behind the predictions of the model. On the other hand, global explanations aim to comprehensively comprehend the model's behavior by describing the underlying knowledge. As of now, there is no available research on global explanations in the context of end-to-end autonomous driving \cite{tian2017deeptest}. Therefore, future research should focus on addressing this gap.

\begin{figure*}
  \centering

  \begin{tabular}{@{}c@{}}
    \includegraphics[width=.45\linewidth,height=120pt]{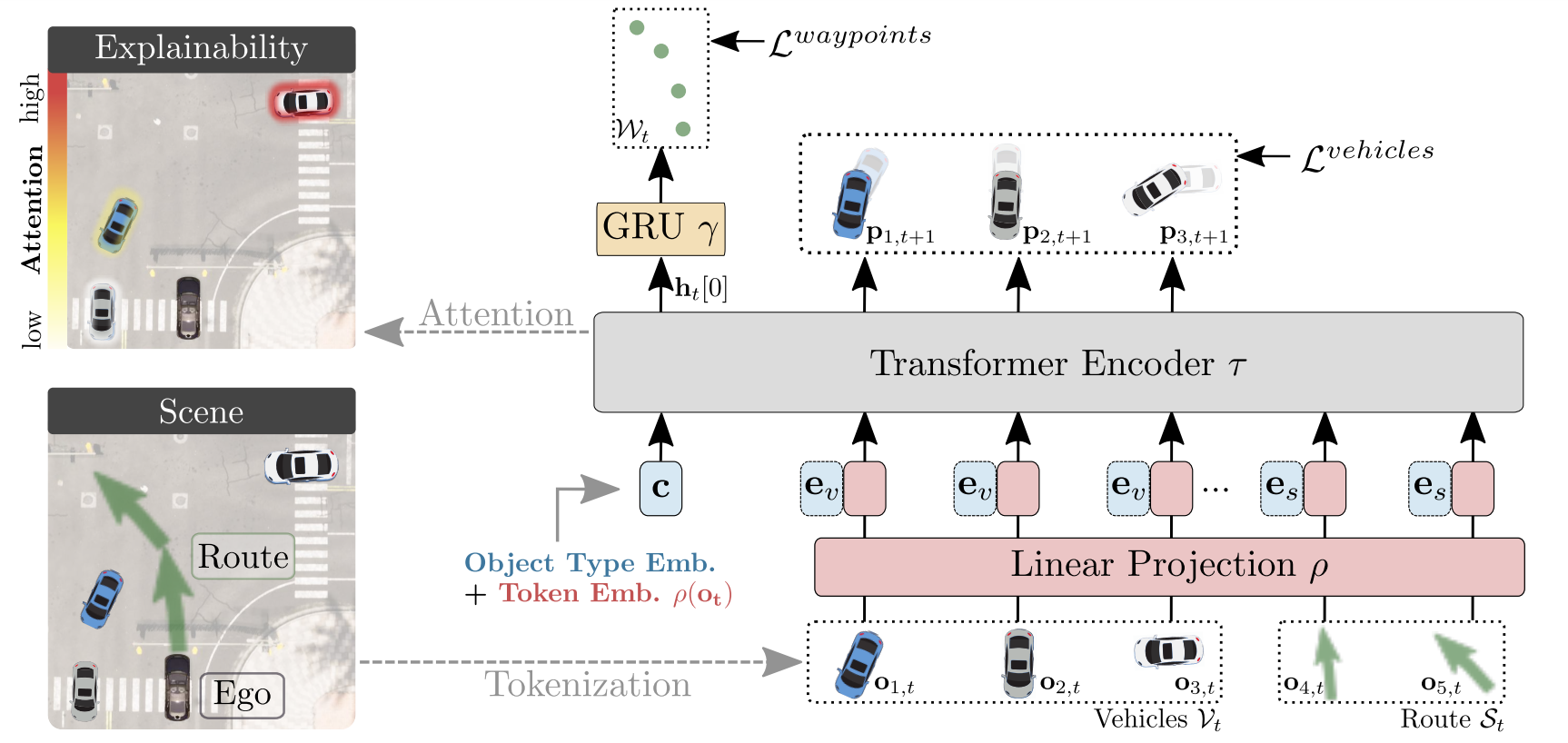} \\[\abovecaptionskip]
    \small (a) PlanT agent's attention 
  \end{tabular}
  \begin{tabular}{@{}c@{}}
    \includegraphics[width=.45\linewidth,height=120pt]{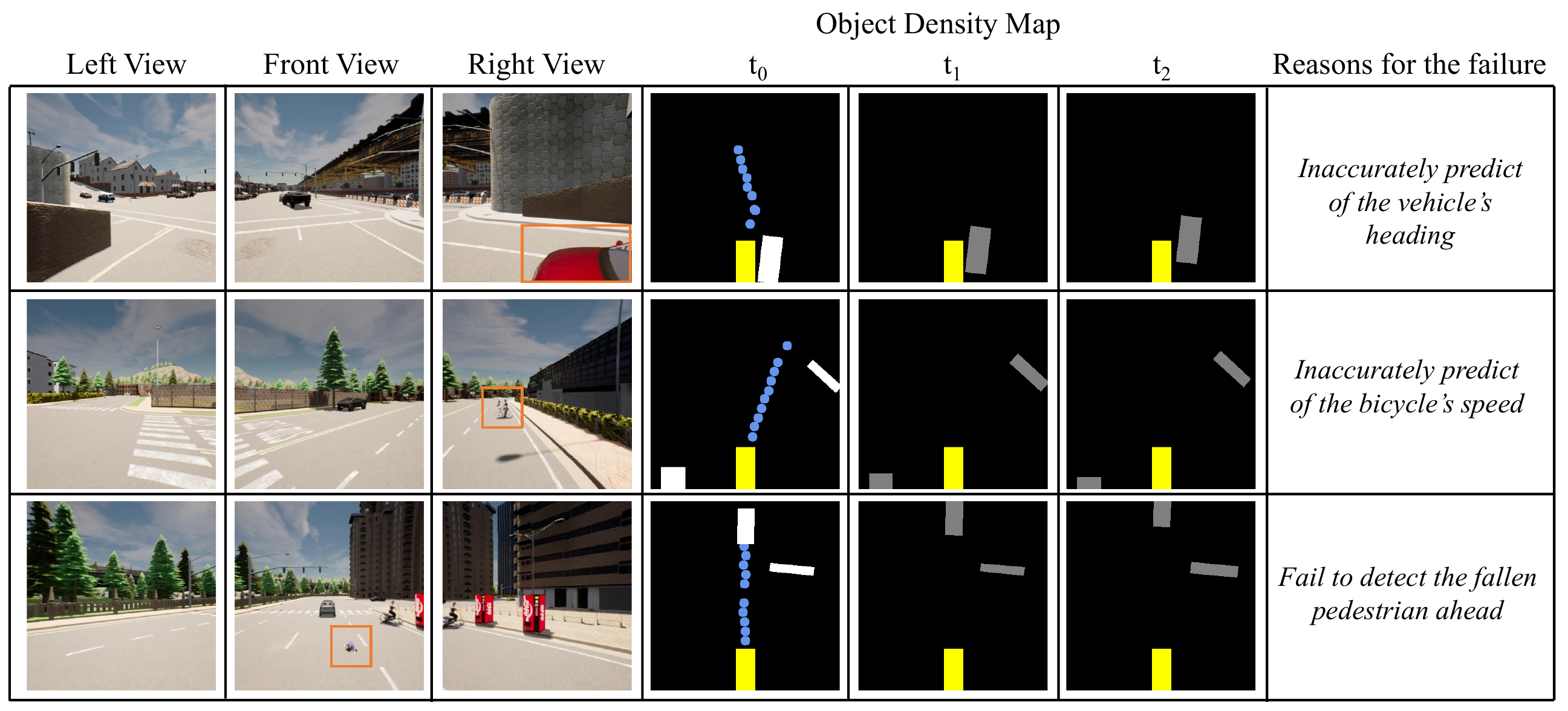} \\[\abovecaptionskip]
    \small (b) InterFuser failure explanation 
  \end{tabular}

  \caption{Explainability Methods: (a) PlanT \cite{renz2022plantc121c} visualization showing the attention given to the agent in various scenarios. (b) Using InterFuser \cite{shao2023safetyc117}, failure cases can be visualized by integrating three RGB views and a predicted object density map. The orange boxes indicate objects that pose a collision risk to the ego-vehicle. The object density map offers predictions for the current traffic scene ($t_{0}$) and future traffic scenes at 1-second ($t_{1}$) and 2-second ($t_{2}$) intervals.}
 
  \label{exp}
  
\end{figure*}

\subsection{Local explanations}\label{A}

A local explanation describes why the model $f$ produces its prediction $y = f(x)$ given an input $x$. There are two approaches: in \ref{post}, we determine which visual region has the most impact, and in \ref{counter}, we identify the factors that caused the model to predict $f(x)$.

\subsubsection{Post-hoc saliency methods}\label{post}

A post-hoc saliency technique attempts to explain which portions of the input space have the most effect on the model's output. These approaches provide a saliency map that illustrates the locations where the model made the most significant decisions.

Post-hoc saliency methods primarily focus on the perception component of the driving architecture. Bojarski et al. \cite{bojarski2018visualbackprop} introduced the first post-hoc saliency approach for visualizing the impact of inputs in autonomous driving. Renz et al. \cite{renz2022plantc121c} proposed the PlanT method (Fig. \ref{exp}(a)), which utilizes an attention mechanism for post-hoc saliency visualization to provide object-level representations using the attention weights of the transformer to identify the most relevant objects. 
Mori et al. \cite{mori2019visual} proposed an attention mechanism that utilizes the model's predictions of steering angle and throttle. These local predictions are employed as visual attention maps and combined with learned parameters using a linear combination to make the final decision. While attention-based methods are often believed to improve the transparency of neural networks, it should be noted that learned attention weights may exhibit weak correlations with several features. The attention weights can provide accurate predictions when measuring different input features during driving. Overall, evaluating the post-hoc effectiveness of attention mechanisms is challenging and often relies on subjective human evaluation.

\subsubsection{Counterfactual explanation}\label{counter}

Saliency approaches focus on answering the `where' question, identifying influential input locations for the model's decision. In contrast, counterfactual explanations address the `what' question by seeking small changes in the input that alter the model's prediction. Counterfactual analysis aims to identify features $X$ within the input $x$ that led to the outcome $y = f(x)$ by creating a new input instance $x'$ where $X$ is modified, resulting in a different outcome $y'$. The modified input instance $x'$ serves as the counterfactual example, and $y'$ represents the contrasting class, such as `What changes in the traffic scenario would cause the vehicle to stop moving?' It could be a red light.

Since the input space consists of semantic dimensions and is modifiable, assessing the causality of input components is straightforward. Li et al. \cite{li2020make} proposed a causal inference technique for identifying risky objects. Steex \cite{jacob2022steex} developed a counterfactual method that modifies the style of the region to explain the visual model. 
The semantic input provides a high-level object representation, making it more interpretable compared to pixel-level representations.

Bansal et al. \cite{bansal2018chauffeurnet} explore the underlying causes of particular outcomes by examining the ChauffeurNet model using manually crafted inputs that involve omitting particular objects.

In End-to-End driving, the steering, throttle, and brake driving outputs can be complemented with auxiliary outputs such as the occupancy and interpretable semantics to demonstrate a specific degree of counterfactual understandability. Chitta et al. \cite{chitta2022transfuserc116} introduce an auxiliary output (semantics map) that employs the A* planner to address a counterfactual inquiry of ``What is the possibility of a collision without braking".
Shao et al. \cite{shao2023safetyc117} designed a system, as shown in Fig. \ref{exp}(b), which infers counterfactual reasoning for the potential failures with the assistance of an intermediate object density map. Sadat et al. \cite{sadat2020perceive} generate a probabilistic semantic occupancy map over
space and time, capturing the positions of diverse road agents. Occupancy maps provide a counterfactual explanation as they act as an intermediary representation to the motion planning system, higher occupancy probabilities will discourages the maneuvers while lower occupancy will encourage them.

\subsection{Global explanations}\label{B}

Global explanations aim to provide an overall understanding of a model's behavior by describing the knowledge it possesses. They are classified into model translation (\ref{mt}) and representation explanation techniques (\ref{er}) for analyzing global explanations.

\subsubsection{Model translation}\label{mt}

The objective of model translation is to transfer the information from the original model to a different model that is inherently interpretable. This involves training an explainable model to mimic the input-output relationship. Recent studies have explored translating deep learning models into decision trees \cite{frosst2017distilling}, rule-based models \cite{zilke2016deepred}, or causal models \cite{harradon2018causal}. However, one limitation of this approach is the potential discrepancies between the interpretable translated model and the original self-driving model.

\subsubsection{Explaining representations}\label{er}

Explaining representations aims to explain the information captured by the model's structures at various scales. Zhang et al. \cite{zhang2018visual} and Bau et al. \cite{bau2017network} make efforts to gain insights into what the neurons capture. The activation of a neuron can be understood by examining input patterns that maximize its activity. For example, one can sample the input using gradient ascent \cite{simonyan2013deep} or generative networks \cite{nguyen2016synthesizing}. Tian et al. \cite{tian2018deeptest} employ the concept of neuron coverage to identify false actions that could potentially lead to fatalities. They partition the input space based on neuron coverage, assuming that inputs with the same neuron coverage will result in the same model decision. Their objective is to increase neuron coverage through transformations such as linear changes in image intensity and affine transformations like rotation and convolution.

\begin{table*}[]
\caption{CARLA AUTONOMOUS DRIVING LEADERBOARD 1.0 SUBMISSION UNTIL AUGUST 2023}
\label{carlaa}
\begin{tabular}{|c|c|c|c|c|cccccccc|c|}
\hline
\rowcolor[HTML]{C0C0C0} 
Rank &
  Submission &
  DS &
  RC &
  IP &
  \multicolumn{1}{c|}{\cellcolor[HTML]{C0C0C0}CP} &
  \multicolumn{1}{c|}{\cellcolor[HTML]{C0C0C0}CV} &
  \multicolumn{1}{c|}{\cellcolor[HTML]{C0C0C0}CL} &
  \multicolumn{1}{c|}{\cellcolor[HTML]{C0C0C0}RLI} &
  \multicolumn{1}{c|}{\cellcolor[HTML]{C0C0C0}SSI} &
  \multicolumn{1}{c|}{\cellcolor[HTML]{C0C0C0}OI} &
  \multicolumn{1}{c|}{\cellcolor[HTML]{C0C0C0}RD} &
  AB &
  Type \\ \hline
 &
   &
  \% &
  \% &
  [0,1] &
  \multicolumn{8}{c|}{infractions/km} &
  E/M \\ \hline
1 &
  ReasonNet \cite{shao2023reasonnet} &
  79.95 &
  89.89 &
  0.89 &
  \multicolumn{1}{c|}{0.02} &
  \multicolumn{1}{c|}{0.13} &
  \multicolumn{1}{c|}{0.01} &
  \multicolumn{1}{c|}{0.08} &
  \multicolumn{1}{c|}{0.00} &
  \multicolumn{1}{c|}{0.04} &
  \multicolumn{1}{c|}{0.00} &
  0.33 &
  E \\ \hline
1 &
  InterFuser \cite{shao2023safetyc117} &
  76.18 &
  88.23 &
  0.84 &
  \multicolumn{1}{c|}{0.04} &
  \multicolumn{1}{c|}{0.37} &
  \multicolumn{1}{c|}{0.14} &
  \multicolumn{1}{c|}{0.22} &
  \multicolumn{1}{c|}{0.00} &
  \multicolumn{1}{c|}{0.13} &
  \multicolumn{1}{c|}{0.00} &
  0.43 &
  E \\ \hline
2 &
  TCP \cite{wu2022trajectoryc120} &
  75.14 &
  85.63 &
  0.87 &
  \multicolumn{1}{c|}{0.00} &
  \multicolumn{1}{c|}{0.32} &
  \multicolumn{1}{c|}{0.00} &
  \multicolumn{1}{c|}{0.09} &
  \multicolumn{1}{c|}{0.00} &
  \multicolumn{1}{c|}{0.04} &
  \multicolumn{1}{c|}{0.00} &
  0.54 &
  E \\ \hline
3 &
  TF++ \cite{Jaeger2023ICCV} &
  66.32 &
  78.57 &
  0.84 &
  \multicolumn{1}{c|}{0.00} &
  \multicolumn{1}{c|}{0.50} &
  \multicolumn{1}{c|}{0.00} &
  \multicolumn{1}{c|}{0.01} &
  \multicolumn{1}{c|}{0.00} &
  \multicolumn{1}{c|}{0.12} &
  \multicolumn{1}{c|}{0.00} &
  0.71 &
  E \\ \hline
3 &
  LAV \cite{chen2022learning5orc98} &
  61.85 &
  94.46 &
  0.64 &
  \multicolumn{1}{c|}{0.04} &
  \multicolumn{1}{c|}{0.70} &
  \multicolumn{1}{c|}{0.02} &
  \multicolumn{1}{c|}{0.17} &
  \multicolumn{1}{c|}{0.00} &
  \multicolumn{1}{c|}{0.25} &
  \multicolumn{1}{c|}{0.09} &
  0.10 &
  E \\ \hline
4 &
  TransFuser \cite{chitta2022transfuserc116} &
  61.18 &
  86.69 &
  0.71 &
  \multicolumn{1}{c|}{0.04} &
  \multicolumn{1}{c|}{0.81} &
  \multicolumn{1}{c|}{0.01} &
  \multicolumn{1}{c|}{0.05} &
  \multicolumn{1}{c|}{0.00} &
  \multicolumn{1}{c|}{0.23} &
  \multicolumn{1}{c|}{0.00} &
  0.43 &
  E \\ \hline
5 &
  Latent TransFuser \cite{chitta2022transfuserc116} &
  45.20 &
  66.31 &
  0.72 &
  \multicolumn{1}{c|}{0.02} &
  \multicolumn{1}{c|}{1.11} &
  \multicolumn{1}{c|}{0.02} &
  \multicolumn{1}{c|}{0.05} &
  \multicolumn{1}{c|}{0.00} &
  \multicolumn{1}{c|}{0.16} &
  \multicolumn{1}{c|}{0.00} &
  1.82 &
  E \\ \hline
6 &
  GRIAD \cite{chekroun2021gric20} &
  36.79 &
  61.85 &
  0.60 &
  \multicolumn{1}{c|}{0.00} &
  \multicolumn{1}{c|}{2.77} &
  \multicolumn{1}{c|}{0.41} &
  \multicolumn{1}{c|}{0.48} &
  \multicolumn{1}{c|}{0.00} &
  \multicolumn{1}{c|}{1.39} &
  \multicolumn{1}{c|}{1.11} &
  0.84 &
  E \\ \hline
7 &
  TransFuser+ \cite{chitta2022transfuserc116} &
  34.58 &
  69.84 &
  0.56 &
  \multicolumn{1}{c|}{0.04} &
  \multicolumn{1}{c|}{0.70} &
  \multicolumn{1}{c|}{0.03} &
  \multicolumn{1}{c|}{0.75} &
  \multicolumn{1}{c|}{0.00} &
  \multicolumn{1}{c|}{0.18} &
  \multicolumn{1}{c|}{0.00} &
  2.41 &
  E \\ \hline
8 &
  World on Rails \cite{chen2021learning11orc136} &
  31.37 &
  57.65 &
  0.56 &
  \multicolumn{1}{c|}{0.61} &
  \multicolumn{1}{c|}{1.35} &
  \multicolumn{1}{c|}{1.02} &
  \multicolumn{1}{c|}{0.79} &
  \multicolumn{1}{c|}{0.00} &
  \multicolumn{1}{c|}{0.96} &
  \multicolumn{1}{c|}{1.69} &
  0.47 &
  E \\ \hline
9 &
  MaRLn  \cite{toromanoff2020endc66} &
  24.98 &
  46.97 &
  0.52 &
  \multicolumn{1}{c|}{0.00} &
  \multicolumn{1}{c|}{2.33} &
  \multicolumn{1}{c|}{2.47} &
  \multicolumn{1}{c|}{0.55} &
  \multicolumn{1}{c|}{0.00} &
  \multicolumn{1}{c|}{1.82} &
  \multicolumn{1}{c|}{1.44} &
  0.94 &
  E \\ \hline
10 &
  NEAT \cite{chitta2021neatc100} &
  21.83 &
  41.71 &
  0.65 &
  \multicolumn{1}{c|}{0.04} &
  \multicolumn{1}{c|}{0.74} &
  \multicolumn{1}{c|}{0.62} &
  \multicolumn{1}{c|}{0.70} &
  \multicolumn{1}{c|}{0.00} &
  \multicolumn{1}{c|}{2.68} &
  \multicolumn{1}{c|}{0.00} &
  5.22 &
  E \\ \hline
11 &
  AIM-MT \cite{chitta2021neatc100} &
  19.38 &
  67.02 &
  0.39 &
  \multicolumn{1}{c|}{0.18} &
  \multicolumn{1}{c|}{1.53} &
  \multicolumn{1}{c|}{0.12} &
  \multicolumn{1}{c|}{1.55} &
  \multicolumn{1}{c|}{0.00} &
  \multicolumn{1}{c|}{0.35} &
  \multicolumn{1}{c|}{0.00} &
  2.11 &
  E \\ \hline
12 &
  TransFuser \cite{prakash2021multic121} &
  16.93 &
  51.82 &
  0.42 &
  \multicolumn{1}{c|}{0.91} &
  \multicolumn{1}{c|}{1.09} &
  \multicolumn{1}{c|}{0.19} &
  \multicolumn{1}{c|}{1.26} &
  \multicolumn{1}{c|}{0.00} &
  \multicolumn{1}{c|}{0.57} &
  \multicolumn{1}{c|}{0.00} &
  1.96 &
  E \\ \hline
13 &
  CNN-Planner \cite{9995888} &
  15.40 &
  50.05 &
  0.41 &
  \multicolumn{1}{c|}{0.08} &
  \multicolumn{1}{c|}{4.67} &
  \multicolumn{1}{c|}{0.42} &
  \multicolumn{1}{c|}{0.35} &
  \multicolumn{1}{c|}{0.00} &
  \multicolumn{1}{c|}{2.78} &
  \multicolumn{1}{c|}{0.12} &
  4.63 &
  M \\ \hline
14 &
  Learning by \cite{chen2020learningc60} &
  8.94 &
  17.54 &
  0.73 &
  \multicolumn{1}{c|}{0.00} &
  \multicolumn{1}{c|}{0.40} &
  \multicolumn{1}{c|}{1.16} &
  \multicolumn{1}{c|}{0.71} &
  \multicolumn{1}{c|}{0.00} &
  \multicolumn{1}{c|}{1.52} &
  \multicolumn{1}{c|}{0.03} &
  4.69 &
  E \\ \hline
15 &
  MaRLn \cite{toromanoff2020endc66} &
  5.56 &
  24.72 &
  0.36 &
  \multicolumn{1}{c|}{0.77} &
  \multicolumn{1}{c|}{3.25} &
  \multicolumn{1}{c|}{13.23} &
  \multicolumn{1}{c|}{0.85} &
  \multicolumn{1}{c|}{0.00} &
  \multicolumn{1}{c|}{10.73} &
  \multicolumn{1}{c|}{2.97} &
  11.41 &
  E \\ \hline
16 &
  CILRS \cite{chitta2021neatc100} &
  5.37 &
  14.40 &
  0.55 &
  \multicolumn{1}{c|}{2.69} &
  \multicolumn{1}{c|}{1.48} &
  \multicolumn{1}{c|}{2.35} &
  \multicolumn{1}{c|}{1.62} &
  \multicolumn{1}{c|}{0.00} &
  \multicolumn{1}{c|}{4.55} &
  \multicolumn{1}{c|}{4.14} &
  4.28 &
  E \\ \hline
17 &
  CaRINA \cite{rosero2020software} &
  4.56 &
  23.80 &
  0.41 &
  \multicolumn{1}{c|}{0.01} &
  \multicolumn{1}{c|}{7.56} &
  \multicolumn{1}{c|}{51.52} &
  \multicolumn{1}{c|}{20.64} &
  \multicolumn{1}{c|}{0.00} &
  \multicolumn{1}{c|}{14.32} &
  \multicolumn{1}{c|}{0.00} &
  10055.99 &
  M \\ \hline
\end{tabular}
\begin{tablenotes}
    \item Route Completion (RC), Infraction Score/penalty (IS), Driving score (DS), Collisions pedestrians (CP)/(PC), Collisions vehicles (CV), Collisions layout (CL)/(LC), Red light infractions (RLI), Red light violation (RV), Stop sign infractions (SSI), Off-road infractions (OI), Route deviations (RD), Agent blocked (AB), End-to-End Architecture (E), Modular Architecture (M).
\end{tablenotes}
\label{carlaa}
\end{table*}

\section{Evaluation}\label{evaluation}

The evaluation of the End-to-End system consists of open-loop evaluation and closed-loop evaluation. The open loop is assessed using real-world benchmark datasets such as KITTI \cite{geiger2013vision} and nuScenes \cite{caesar2020nuscenes}. It compares the system's driving behavior with expert actions and measures the deviation. Measures such as MinADE, MinFDE \cite{hu2023planningoriented}, L2 error \cite{hu2022st}, and collision rate \cite{zeng2021endtoend} are some of the evaluation metrics presented in Table \ref{literature}. In contrast the closed-loop evaluation directly assesses the system in controlled real-world or simulated settings by allowing it to drive independently and learn safe driving maneuvers. 

In the open-loop evaluation of End-to-End driving systems, the system's inputs, such as camera images or LiDAR data, are provided to the system. The resulting outputs, such as steering commands and vehicle speed, are evaluated against predefined driving behaviors. The evaluation metrics commonly used in the open-loop evaluation include measures of the system's ability to follow the desired trajectory or driving behaviors, such as the mean squared error \cite{zhao2021samc65}, L2 \cite{hu2023planningoriented,casas2021mp3} between the predicted and actual trajectories or the percentage of time the system remains within a certain distance of the desired trajectory \cite{wu2022trajectoryc120}. Other evaluation metrics may also be employed to assess the system's performance in specific driving scenarios \cite{prakash2021multic121, hanselmann2022king2orc99}, such as the system's capability to navigate intersections, handle obstacles, or perform lane changes. The open loop provides faster initial assessment based on functionalities and is also helpful for testing specific components or behaviors in isolation. However, they inherit drawbacks from the benchmark datasets as they cannot generalize to wider geographical distribution.

Most of the recent End-to-End systems are evaluated in closed-loop settings such as LEADERBOARD and NOCRASH \cite{dosovitskiy2017carla}. Table \ref{carlaa} compares all the state-of-the-art methods on the CARLA public leaderboard. The CARLA leaderboard analyzes autonomous driving systems in unanticipated environments. Vehicles are tasked with completing a set of specified routes, incorporating risky scenarios such as unexpectedly crossing pedestrians or sudden lane changes. The leaderboard measures how far the vehicle has successfully traveled on the given Town route within a time constraint and how many times it has incurred infractions. Several metrics provide a comprehensive understanding of the driving system, which are mentioned below:

\begin{itemize}
    \item Route Completion (RC): \cite{chitta2022transfuserc116,chen2022learning5orc98,renz2022plantc121c,wu2022trajectoryc120,shao2023safetyc117}  measures the percentage of the distance that an agent can complete.
   
    \item Infraction Score/penalty (IS): \cite{prakash2021multic121,zhang2021learning,chitta2021neatc100} is a geometric series that tracks infractions and aggregates the infraction penalties. It measures how often an agent drives without causing infractions.
    \item Driving score (DS): \cite{zhang2021endc94,chen2021learningc136,ohn2020learningc62} is a primary metric calculated as the multiplication of the route completion and the infraction penalty. It measures the route completion rate weighted by infractions per route.
\end{itemize}

There are specific metrics that evaluate infractions; each metric has penalty coefficients applied every time an infraction takes place. Collisions with pedestrians, collisions with other vehicles, collisions with static elements, collisions layout, red light infractions, stop sign infractions, and off-road infractions are some of the metrics used \cite{9995888}. Closed-loop evaluation provides dynamic testing adaptability where one can provide customized configuration and sensor settings. The feedback loop in it allows for iterative refinement, enabling the system to learn and improve from mistakes and experiences. However, several challenges are associated with closed-loop. These include the complexity of the initial setup and the domain gap, which might require additional fine-tuning.

\section{Datasets and simulator} 
\label{s_dataset}

\subsection{Datasets}

In End-to-End models, the quality and richness of data are critical aspects of model training. Instead of using different hyperparameters, the training data is the most crucial factor influencing the model's performance. The amount of information fed into the model determines the kind of outcomes it produces. We summarized the self-driving dataset based on their sensor modalities, including camera, LiDAR, GNSS, and dynamics. The content of the datasets includes urban driving, traffic, and different road conditions. Weather conditions also influence the model's performance. Some datasets, such as ApolloScape \cite{huang2019apolloscape}, capture all weather conditions from sunny to snowy. The details are provided in Table \ref{dataset1}.

\subsection{Simulators and toolsets}

Standard testing of End-to-End driving and learning pipelines requires advanced software simulators to process information and make conclusions for their various functionalities. Experimenting with such driving systems is expensive, and conducting tests on public roads is heavily restricted. Simulation environments assist in training specific algorithms/modules before road testing. Simulators like Carla \cite{dosovitskiy2017carla} offer flexibility to simulate the environment based on experimental requirements, including weather conditions, traffic flow, road agents, etc. Simulators play a crucial role in generating safety-critical scenarios and contribute to model generalization for detecting and preventing such scenarios.

Widely used platforms for training End-to-End driving pipelines are compared in Table \ref{sim}. MATLAB/Simulink \cite{noauthor_automated_nodate} is used for various settings; it contains efficient plot functions and has the ability to co-simulate with other software, such as CarSim \cite{noauthor_mechanical_nodate}, which simplifies the creation of different settings. PreScan \cite{noauthor_prescan_nodate} can mimic real-world environments, including weather conditions, which MATLAB and CarSim lack. It also supports the MATLAB Simulink interface, making modeling more effective. Gazebo \cite{noauthor_gazebo_nodate} is well-known for its high versatility and easy connection with ROS. In contrast to the CARLA and LGSVL \cite{noauthor_svl_nodate} simulators, creating a simulated environment with Gazebo requires mechanical effort. CARLA and LGSVL offer high-quality simulation frameworks that require a GPU processing unit to operate at a decent speed and frame rate. CARLA is built on the Unreal Engine, while LGSVL is based on the Unity game engine. The API allows users to access various capabilities in CARLA and LGSVL, from developing customizable sensors to map generation. LGSVL generally links to the driving stack through various bridges, and CARLA allows built-in bridge connections via ROS and Autoware.

\section{Future research directions}
\label{s_future}

This section will highlight the possible research directions that can drive future advancements in the domain from the perspective of learning principles, safety, explainability, and others.

\subsection{Learning robustness}

Current research in End-to-End autonomous driving mainly focuses on reinforcement learning (Section \ref{AA}) and imitation learning (Section \ref{AAI}) methods. RL trains agents by interacting with simulated environments, while IL learns from expert agents without extensive environmental interaction. However, challenges like distribution shift in IL and computational instability in RL highlight the need for further improvements.

\subsection{Enhanced safety}
Ensuring the behavioral safety of vehicles and accurately predicting uncertain behaviors are key aspects in safety research as discussed in Section \ref{safety}. An effective system should be capable of handling various driving situations, contributing to comfortable and reliable transportation. To facilitate the widespread adoption of End-to-End approaches, it is essential to refine safety constraints and enhance their effectiveness. 

\subsection{Advancing model explainability}
The lack of interpretability poses a new challenge for the advancement of End-to-End driving. However, ongoing efforts (Section \ref{explainability}) are being made to address this issue by designing and generating interpretable features. These efforts have shown promising improvements in both performance and explainability. However, further exploration is required in global explanation strategies, including designing novel approaches to explain model actions leading to failures and suggesting potential solutions. Future research can also explore ways to improve feedback mechanisms, allowing users to understand the decision-making process and infuse confidence in the reliability of End-to-End driving systems.

\subsection{Collaboration perception systems}
Vehicles can communicate directly utilizing collaborative perception to observe surroundings beyond their line of sight and field of view. This approach addresses issues related to occlusion and limited receptive fields. Cooperative or collaborative perception enables vehicles in the same area to communicate and jointly assess the scene.

\begin{landscape}
\begin{table}[]

\caption{CUMULATIVE LIST OF DATASETS WITH THEIR DYNAMICS FOR END-TO-END TRAINING}
\label{dataset1}
\begin{tabular}{|c|c|ccccccc|ccc|ccc|c|c|c|}
\hline
\rowcolor[HTML]{C0C0C0} 
Datasets &
  Year &
  \multicolumn{7}{c|}{\cellcolor[HTML]{C0C0C0}Sensors Modalities} &
  \multicolumn{3}{c|}{\cellcolor[HTML]{C0C0C0}Content} &
  \multicolumn{3}{c|}{\cellcolor[HTML]{C0C0C0}Weather} &
  Size &
  Location &
  License \\ \cline{3-15}
\rowcolor[HTML]{FFFFFF} 
 &
   &
  \multicolumn{1}{c|}{\cellcolor[HTML]{FFFFFF}\rotatebox{90}{\parbox{2cm}{Cameras}}} &
  \multicolumn{1}{c|}{\cellcolor[HTML]{FFFFFF}\rotatebox{90}{\parbox{2cm}{LiDAR}}} &
  \multicolumn{1}{c|}{\cellcolor[HTML]{FFFFFF}\rotatebox{90}{\parbox{2cm}{GNSS}}} &
  \multicolumn{1}{c|}{\cellcolor[HTML]{FFFFFF}\rotatebox{90}{\parbox{2cm}{Steering}}} &
  \multicolumn{1}{c|}{\cellcolor[HTML]{FFFFFF}\rotatebox{90}{\parbox{2cm}{Speed, \\ Acceleration}}} &
  \multicolumn{1}{c|}{\cellcolor[HTML]{FFFFFF}\rotatebox{90}{\parbox{2cm}{Navigational \\ command}}} &
  \rotatebox{90}{\parbox{2cm}{Route planner}} &
  \multicolumn{1}{c|}{\cellcolor[HTML]{FFFFFF}\rotatebox{90}{\parbox{2cm}{Obstacles }}} &
  \multicolumn{1}{c|}{\cellcolor[HTML]{FFFFFF}\rotatebox{90}{\parbox{2cm}{Traffic}}} &
  \rotatebox{90}{\parbox{2cm}{Raods}} &
  \multicolumn{1}{c|}{\cellcolor[HTML]{FFFFFF}\rotatebox{90}{\parbox{2cm}{Sunny}}} &
  \multicolumn{1}{c|}{\cellcolor[HTML]{FFFFFF}\rotatebox{90}{\parbox{2cm}{Rain}}} &
  \rotatebox{90}{\parbox{2cm}{Snow or Fog}} &
   &
   &
   \\ \hline
Udacity \cite{udacity_dataset:2017} &
  2016 &
  \multicolumn{1}{c|}{\checkmark} &
  \multicolumn{1}{c|}{\checkmark} &
  \multicolumn{1}{c|}{\checkmark} &
  \multicolumn{1}{c|}{\checkmark} &
  \multicolumn{1}{c|}{\checkmark} &
  \multicolumn{1}{c|}{} &
   &
  \multicolumn{1}{c|}{\checkmark} &
  \multicolumn{1}{c|}{\checkmark} &
   &
  \multicolumn{1}{c|}{\checkmark} &
  \multicolumn{1}{c|}{} &
   &
  5h &
  Mountain View &
  MIT \\ \hline
Drive360 \cite{hecker2018end} &
  2019 &
  \multicolumn{1}{c|}{\checkmark} &
  \multicolumn{1}{c|}{} &
  \multicolumn{1}{c|}{\checkmark} &
  \multicolumn{1}{c|}{\checkmark} &
  \multicolumn{1}{c|}{\checkmark} &
  \multicolumn{1}{c|}{} &
  \checkmark &
  \multicolumn{1}{c|}{\checkmark} &
  \multicolumn{1}{c|}{} &
   &
  \multicolumn{1}{c|}{\checkmark} &
  \multicolumn{1}{c|}{\checkmark} &
   &
  55h &
  Switzerland &
  Academic \\ \hline
Comma.ai 2016 \cite{santana2016learning} &
  2016 &
  \multicolumn{1}{c|}{\checkmark} &
  \multicolumn{1}{c|}{} &
  \multicolumn{1}{c|}{\checkmark} &
  \multicolumn{1}{c|}{\checkmark} &
  \multicolumn{1}{c|}{\checkmark} &
  \multicolumn{1}{c|}{} &
   &
  \multicolumn{1}{c|}{} &
  \multicolumn{1}{c|}{} &
   &
  \multicolumn{1}{c|}{\checkmark} &
  \multicolumn{1}{c|}{} &
   &
  7h 15min &
  San Francisco &
  CC BY-NC-SA 3.0 \\ \hline
Comma.ai 2019 \cite{schafer2018commute} &
  2019 &
  \multicolumn{1}{c|}{\checkmark} &
  \multicolumn{1}{c|}{} &
  \multicolumn{1}{c|}{\checkmark} &
  \multicolumn{1}{c|}{\checkmark} &
  \multicolumn{1}{c|}{\checkmark} &
  \multicolumn{1}{c|}{} &
   &
  \multicolumn{1}{c|}{} &
  \multicolumn{1}{c|}{} &
   &
  \multicolumn{1}{c|}{\checkmark} &
  \multicolumn{1}{c|}{} &
   &
  30h &
  San Jose California  &
  MIT \\ \hline
BDD 100 \cite{yu2018bdd100k} &
  2018 &
  \multicolumn{1}{c|}{\checkmark} &
  \multicolumn{1}{c|}{} &
  \multicolumn{1}{c|}{\checkmark} &
  \multicolumn{1}{c|}{} &
  \multicolumn{1}{c|}{} &
  \multicolumn{1}{c|}{} &
   &
  \multicolumn{1}{c|}{\checkmark} &
  \multicolumn{1}{c|}{\checkmark} &
   &
  \multicolumn{1}{c|}{\checkmark} &
  \multicolumn{1}{c|}{\checkmark} &
   &
  1100h &
  US &
  Berkley \\ \hline
Oxford RobotCar \cite{maddern20171c26} &
  2019 &
  \multicolumn{1}{c|}{\checkmark} &
  \multicolumn{1}{c|}{\checkmark} &
  \multicolumn{1}{c|}{\checkmark} &
  \multicolumn{1}{c|}{} &
  \multicolumn{1}{c|}{} &
  \multicolumn{1}{c|}{} &
   &
  \multicolumn{1}{c|}{\checkmark} &
  \multicolumn{1}{c|}{\checkmark} &
   &
  \multicolumn{1}{c|}{\checkmark} &
  \multicolumn{1}{c|}{\checkmark} &
  \checkmark &
  214h &
  Oxford &
  CC BY-NC-SA 4.0 \\ \hline
HDD \cite{ramanishka2018toward} &
  2018 &
  \multicolumn{1}{c|}{\checkmark} &
  \multicolumn{1}{c|}{\checkmark} &
  \multicolumn{1}{c|}{\checkmark} &
  \multicolumn{1}{c|}{\checkmark} &
  \multicolumn{1}{c|}{\checkmark} &
  \multicolumn{1}{c|}{\checkmark} &
   &
  \multicolumn{1}{c|}{\checkmark} &
  \multicolumn{1}{c|}{\checkmark} &
  \checkmark &
  \multicolumn{1}{c|}{} &
  \multicolumn{1}{c|}{} &
   &
  104h &
  San Francisco &
  Academic \\ \hline
Brain4Cars \cite{jain2015car} &
  2016 &
  \multicolumn{1}{c|}{\checkmark} &
  \multicolumn{1}{c|}{} &
  \multicolumn{1}{c|}{\checkmark} &
  \multicolumn{1}{c|}{} &
  \multicolumn{1}{c|}{\checkmark} &
  \multicolumn{1}{c|}{} &
   &
  \multicolumn{1}{c|}{\checkmark} &
  \multicolumn{1}{c|}{\checkmark} &
   &
  \multicolumn{1}{c|}{} &
  \multicolumn{1}{c|}{} &
   &
  1180 miles &
  US &
  Academic \\ \hline
Li-Vi \cite{chen2018lidar, dbnet:2018} &
  2018 &
  \multicolumn{1}{c|}{\checkmark} &
  \multicolumn{1}{c|}{\checkmark} &
  \multicolumn{1}{c|}{\checkmark} &
  \multicolumn{1}{c|}{\checkmark} &
  \multicolumn{1}{c|}{\checkmark} &
  \multicolumn{1}{c|}{} &
   &
  \multicolumn{1}{c|}{} &
  \multicolumn{1}{c|}{} &
   &
  \multicolumn{1}{c|}{} &
  \multicolumn{1}{c|}{} &
   &
  10h &
  China &
  Academic \\ \hline
DDD17 \cite{binas2017ddd17} &
  2017 &
  \multicolumn{1}{c|}{\checkmark} &
  \multicolumn{1}{c|}{} &
  \multicolumn{1}{c|}{\checkmark} &
  \multicolumn{1}{c|}{\checkmark} &
  \multicolumn{1}{c|}{\checkmark} &
  \multicolumn{1}{c|}{} &
   &
  \multicolumn{1}{c|}{\checkmark} &
  \multicolumn{1}{c|}{\checkmark} &
   &
  \multicolumn{1}{c|}{\checkmark} &
  \multicolumn{1}{c|}{\checkmark} &
   &
  12h &
  Switzerland, Germany &
  CC-BY-NC-SA-4.0 \\ \hline
A2D2 \cite{geyer2019a2d2} &
  2020 &
  \multicolumn{1}{c|}{\checkmark} &
  \multicolumn{1}{c|}{\checkmark} &
  \multicolumn{1}{c|}{\checkmark} &
  \multicolumn{1}{c|}{\checkmark} &
  \multicolumn{1}{c|}{\checkmark} &
  \multicolumn{1}{c|}{} &
   &
  \multicolumn{1}{c|}{\checkmark} &
  \multicolumn{1}{c|}{\checkmark} &
   &
  \multicolumn{1}{c|}{\checkmark} &
  \multicolumn{1}{c|}{} &
   &
  390k frames &
  South of Germany &
  CC BY-ND 4.0 \\ \hline
nuScenes \cite{caesar2020nuscenes} &
  2019 &
  \multicolumn{1}{c|}{\checkmark} &
  \multicolumn{1}{c|}{\checkmark} &
  \multicolumn{1}{c|}{\checkmark} &
  \multicolumn{1}{c|}{} &
  \multicolumn{1}{c|}{} &
  \multicolumn{1}{c|}{} &
   &
  \multicolumn{1}{c|}{\checkmark} &
  \multicolumn{1}{c|}{} &
   &
  \multicolumn{1}{c|}{\checkmark} &
  \multicolumn{1}{c|}{\checkmark} &
   &
  5.5h &
  Boston, Singapore &
  Non-commercial \\ \hline
Waymo \cite{sun2019scalability} &
  2019 &
  \multicolumn{1}{c|}{\checkmark} &
  \multicolumn{1}{c|}{\checkmark} &
  \multicolumn{1}{c|}{\checkmark} &
  \multicolumn{1}{c|}{\checkmark} &
  \multicolumn{1}{c|}{\checkmark} &
  \multicolumn{1}{c|}{} &
   &
  \multicolumn{1}{c|}{\checkmark} &
  \multicolumn{1}{c|}{} &
   &
  \multicolumn{1}{c|}{\checkmark} &
  \multicolumn{1}{c|}{} &
   &
  5.5h &
  California&
  Non-commercial \\ \hline
H3D \cite{patil2019h3d} &
  2019 &
  \multicolumn{1}{c|}{\checkmark} &
  \multicolumn{1}{c|}{\checkmark} &
  \multicolumn{1}{c|}{\checkmark} &
  \multicolumn{1}{c|}{\checkmark} &
  \multicolumn{1}{c|}{\checkmark} &
  \multicolumn{1}{c|}{} &
   &
  \multicolumn{1}{c|}{} &
  \multicolumn{1}{c|}{} &
   &
  \multicolumn{1}{c|}{\checkmark} &
  \multicolumn{1}{c|}{} &
   &
  N/A &
  Japan &
  Academic \\ \hline
HAD \cite{kim2019grounding} &
  2019 &
  \multicolumn{1}{c|}{\checkmark} &
  \multicolumn{1}{c|}{} &
  \multicolumn{1}{c|}{\checkmark} &
  \multicolumn{1}{c|}{\checkmark} &
  \multicolumn{1}{c|}{\checkmark} &
  \multicolumn{1}{c|}{\checkmark} &
   &
  \multicolumn{1}{c|}{} &
  \multicolumn{1}{c|}{\checkmark} &
   &
  \multicolumn{1}{c|}{} &
  \multicolumn{1}{c|}{} &
   &
  30h &
  San Francisco &
  Academic \\ \hline
BIT \cite{articlebit} &
  2015 &
  \multicolumn{1}{c|}{\checkmark} &
  \multicolumn{1}{c|}{} &
  \multicolumn{1}{c|}{} &
  \multicolumn{1}{c|}{} &
  \multicolumn{1}{c|}{} &
  \multicolumn{1}{c|}{} &
   &
  \multicolumn{1}{c|}{\checkmark} &
  \multicolumn{1}{c|}{} &
   &
  \multicolumn{1}{c|}{\checkmark} &
  \multicolumn{1}{c|}{} &
   &
  9850 frames &
  Beijing &
  Academics \\ \hline
UA-DETRAC \cite{wen2020ua} &
  2015 &
  \multicolumn{1}{c|}{\checkmark} &
  \multicolumn{1}{c|}{} &
  \multicolumn{1}{c|}{} &
  \multicolumn{1}{c|}{} &
  \multicolumn{1}{c|}{} &
  \multicolumn{1}{c|}{} &
   &
  \multicolumn{1}{c|}{\checkmark} &
  \multicolumn{1}{c|}{} &
   &
  \multicolumn{1}{c|}{\checkmark} &
  \multicolumn{1}{c|}{} &
   &
  140k frames &
  Beijing, Tianjing &
  CC BY-NC-SA 3.0 \\ \hline
DFG \cite{Tabernik2019ITS} &
  2019 &
  \multicolumn{1}{c|}{\checkmark} &
  \multicolumn{1}{c|}{} &
  \multicolumn{1}{c|}{} &
  \multicolumn{1}{c|}{} &
  \multicolumn{1}{c|}{} &
  \multicolumn{1}{c|}{} &
   &
  \multicolumn{1}{c|}{} &
  \multicolumn{1}{c|}{\checkmark} &
   &
  \multicolumn{1}{c|}{\checkmark} &
  \multicolumn{1}{c|}{\checkmark} &
   &
  7k+8k &
  Slovenia &
  CC BY-NC-SA 4.0 \\ \hline
Bosch \cite{BehrendtNovak2017ICRA} &
  2017 &
  \multicolumn{1}{c|}{\checkmark} &
  \multicolumn{1}{c|}{} &
  \multicolumn{1}{c|}{} &
  \multicolumn{1}{c|}{} &
  \multicolumn{1}{c|}{} &
  \multicolumn{1}{c|}{} &
   &
  \multicolumn{1}{c|}{} &
  \multicolumn{1}{c|}{\checkmark} &
   &
  \multicolumn{1}{c|}{\checkmark} &
  \multicolumn{1}{c|}{} &
   &
  8334 frames &
  Germany &
  Research Only \\ \hline
Tencent 100k \cite{Zhe_2016_CVPR} &
  2016 &
  \multicolumn{1}{c|}{\checkmark} &
  \multicolumn{1}{c|}{} &
  \multicolumn{1}{c|}{} &
  \multicolumn{1}{c|}{} &
  \multicolumn{1}{c|}{} &
  \multicolumn{1}{c|}{} &
   &
  \multicolumn{1}{c|}{} &
  \multicolumn{1}{c|}{\checkmark} &
   &
  \multicolumn{1}{c|}{\checkmark} &
  \multicolumn{1}{c|}{} &
   &
  30k &
  China &
  CC-BY-NC \\ \hline
LISA \cite{jensen2016vision}&
  2012 &
  \multicolumn{1}{c|}{\checkmark} &
  \multicolumn{1}{c|}{} &
  \multicolumn{1}{c|}{} &
  \multicolumn{1}{c|}{} &
  \multicolumn{1}{c|}{} &
  \multicolumn{1}{c|}{} &
   &
  \multicolumn{1}{c|}{} &
  \multicolumn{1}{c|}{\checkmark} &
   &
  \multicolumn{1}{c|}{\checkmark} &
  \multicolumn{1}{c|}{} &
   &
  20k &
  California &
  Research Only \\ \hline
STSD \cite{inproceedingsstrd}&
  2011 &
  \multicolumn{1}{c|}{\checkmark} &
  \multicolumn{1}{c|}{} &
  \multicolumn{1}{c|}{} &
  \multicolumn{1}{c|}{} &
  \multicolumn{1}{c|}{} &
  \multicolumn{1}{c|}{} &
   &
  \multicolumn{1}{c|}{} &
  \multicolumn{1}{c|}{\checkmark} &
   &
  \multicolumn{1}{c|}{\checkmark} &
  \multicolumn{1}{c|}{} &
   &
  2503 frames &
  Sweden &
  CC BY-SA 4.0 \\ \hline
GTSRB \cite{stallkamp2011german} &
  2013 &
  \multicolumn{1}{c|}{\checkmark} &
  \multicolumn{1}{c|}{} &
  \multicolumn{1}{c|}{} &
  \multicolumn{1}{c|}{} &
  \multicolumn{1}{c|}{} &
  \multicolumn{1}{c|}{} &
   &
  \multicolumn{1}{c|}{} &
  \multicolumn{1}{c|}{\checkmark} &
   &
  \multicolumn{1}{c|}{\checkmark} &
  \multicolumn{1}{c|}{\checkmark} &
   &
  50k &
  Germany &
  CC0 1.0 \\ \hline
KUL \cite{mathias2013traffic} &
  2013 &
  \multicolumn{1}{c|}{\checkmark} &
  \multicolumn{1}{c|}{} &
  \multicolumn{1}{c|}{} &
  \multicolumn{1}{c|}{} &
  \multicolumn{1}{c|}{} &
  \multicolumn{1}{c|}{} &
   &
  \multicolumn{1}{c|}{} &
  \multicolumn{1}{c|}{\checkmark} &
   &
  \multicolumn{1}{c|}{\checkmark} &
  \multicolumn{1}{c|}{} &
   &
  16k &
  Flanders &
  CC0 1.0 \\ \hline
Caltech \cite{dollar2009pedestrian}&
  2009 &
  \multicolumn{1}{c|}{\checkmark} &
  \multicolumn{1}{c|}{} &
  \multicolumn{1}{c|}{} &
  \multicolumn{1}{c|}{} &
  \multicolumn{1}{c|}{} &
  \multicolumn{1}{c|}{} &
   &
  \multicolumn{1}{c|}{\checkmark} &
  \multicolumn{1}{c|}{} &
   &
  \multicolumn{1}{c|}{\checkmark} &
  \multicolumn{1}{c|}{} &
   &
  10 hours &
  California &
  CC4.0 \\ \hline
CamVid \cite{brostow2009semantic}&
  2009 &
  \multicolumn{1}{c|}{\checkmark} &
  \multicolumn{1}{c|}{} &
  \multicolumn{1}{c|}{} &
  \multicolumn{1}{c|}{} &
  \multicolumn{1}{c|}{} &
  \multicolumn{1}{c|}{} &
   &
  \multicolumn{1}{c|}{\checkmark} &
  \multicolumn{1}{c|}{\checkmark} &
  \checkmark &
  \multicolumn{1}{c|}{\checkmark} &
  \multicolumn{1}{c|}{} &
   &
  22 min, 14 s &
  Cambridge &
  Academic \\ \hline
Ford\cite{Agarwal_2020}&
  2018 &
  \multicolumn{1}{c|}{\checkmark} &
  \multicolumn{1}{c|}{\checkmark} &
  \multicolumn{1}{c|}{} &
  \multicolumn{1}{c|}{} &
  \multicolumn{1}{c|}{} &
  \multicolumn{1}{c|}{} &
   &
  \multicolumn{1}{c|}{\checkmark} &
  \multicolumn{1}{c|}{} &
   &
  \multicolumn{1}{c|}{\checkmark} &
  \multicolumn{1}{c|}{} &
   &
  66 km &
  Michigan &
  CC-BY-NC-SA 4.0 \\ \hline
KITTI \cite{geiger2013vision}&
  2013 &
  \multicolumn{1}{c|}{\checkmark} &
  \multicolumn{1}{c|}{\checkmark} &
  \multicolumn{1}{c|}{\checkmark} &
  \multicolumn{1}{c|}{} &
  \multicolumn{1}{c|}{} &
  \multicolumn{1}{c|}{} &
   &
  \multicolumn{1}{c|}{\checkmark} &
  \multicolumn{1}{c|}{\checkmark} &
  \checkmark &
  \multicolumn{1}{c|}{\checkmark} &
  \multicolumn{1}{c|}{} &
   &
  43 k &
  Karlsruhe &
  Apache License 2.0 \\ \hline
CityScapes \cite{cordts2016cityscapes} &
  2016 &
  \multicolumn{1}{c|}{\checkmark} &
  \multicolumn{1}{c|}{} &
  \multicolumn{1}{c|}{\checkmark} &
  \multicolumn{1}{c|}{} &
  \multicolumn{1}{c|}{} &
  \multicolumn{1}{c|}{} &
   &
  \multicolumn{1}{c|}{\checkmark} &
  \multicolumn{1}{c|}{\checkmark} &
   &
  \multicolumn{1}{c|}{\checkmark} &
  \multicolumn{1}{c|}{} &
   &
  20+5 K frams &
  Germany, France, Scotland &
  Apache License 2.0 \\ \hline
Mapillary \cite{neuhold2017mapillary}&
  2017 &
  \multicolumn{1}{c|}{\checkmark} &
  \multicolumn{1}{c|}{} &
  \multicolumn{1}{c|}{} &
  \multicolumn{1}{c|}{} &
  \multicolumn{1}{c|}{} &
  \multicolumn{1}{c|}{} &
   &
  \multicolumn{1}{c|}{\checkmark} &
  \multicolumn{1}{c|}{} &
   &
  \multicolumn{1}{c|}{\checkmark} &
  \multicolumn{1}{c|}{\checkmark} &
  \checkmark &
  25000 frames &
  Germany &
  Research Only \\ \hline
ApolloScape \cite{huang2019apolloscape}&
  2018 &
  \multicolumn{1}{c|}{\checkmark} &
  \multicolumn{1}{c|}{\checkmark} &
  \multicolumn{1}{c|}{} &
  \multicolumn{1}{c|}{} &
  \multicolumn{1}{c|}{} &
  \multicolumn{1}{c|}{} &
   &
  \multicolumn{1}{c|}{\checkmark} &
  \multicolumn{1}{c|}{\checkmark} &
   &
  \multicolumn{1}{c|}{\checkmark} &
  \multicolumn{1}{c|}{\checkmark} &
  \checkmark &
  147 k frames &
  China &
  Non-commercial \\ \hline
VERI-Wild \cite{lou2019veri}&
  2019 &
  \multicolumn{1}{c|}{\checkmark} &
  \multicolumn{1}{c|}{} &
  \multicolumn{1}{c|}{} &
  \multicolumn{1}{c|}{} &
  \multicolumn{1}{c|}{} &
  \multicolumn{1}{c|}{} &
   &
  \multicolumn{1}{c|}{\checkmark} &
  \multicolumn{1}{c|}{} &
   &
  \multicolumn{1}{c|}{\checkmark} &
  \multicolumn{1}{c|}{\checkmark} &
   &
  125, 280 hours &
  China &
  Research Only \\ \hline
D2 -City \cite{che2019d2city}&
  2019 &
  \multicolumn{1}{c|}{\checkmark} &
  \multicolumn{1}{c|}{} &
  \multicolumn{1}{c|}{} &
  \multicolumn{1}{c|}{} &
  \multicolumn{1}{c|}{} &
  \multicolumn{1}{c|}{} &
   &
  \multicolumn{1}{c|}{\checkmark} &
  \multicolumn{1}{c|}{} &
  \checkmark &
  \multicolumn{1}{c|}{\checkmark} &
  \multicolumn{1}{c|}{\checkmark} &
   &
  10000 video &
  China &
  Research Only \\ \hline
DriveSeg \cite{mmkedv0320}&
  2020 &
  \multicolumn{1}{c|}{\checkmark} &
  \multicolumn{1}{c|}{} &
  \multicolumn{1}{c|}{} &
  \multicolumn{1}{c|}{} &
  \multicolumn{1}{c|}{} &
  \multicolumn{1}{c|}{} &
   &
  \multicolumn{1}{c|}{\checkmark} &
  \multicolumn{1}{c|}{\checkmark} &
  \checkmark &
  \multicolumn{1}{c|}{\checkmark} &
  \multicolumn{1}{c|}{} &
   &
  500 minutes &
  Massachusetts &
  CC BY-NC 4.0 \\ \hline
\end{tabular}
\end{table}
\end{landscape}

\begin{table*}[] 
\caption{PROMINENT SIMULATORS USED FOR CREATING VIRTUAL ENVIRONMENTS FOR END-TO-END SYSTEMS}
\label{tab:my-table}
\begin{tabular}{|cccccc|}
\hline
\rowcolor[HTML]{C0C0C0} 
\begin{tabular}[c]{@{}c@{}}Simulators\\ \end{tabular} &
  \begin{tabular}[c]{@{}c@{}}MATLAB\\ \cite{noauthor_automated_nodate}\end{tabular} &
  \begin{tabular}[c]{@{}c@{}}CarSim\\ \cite{noauthor_mechanical_nodate}\end{tabular} &
  \begin{tabular}[c]{@{}c@{}}PreScan\\\cite{noauthor_prescan_nodate}\end{tabular} &
  \begin{tabular}[c]{@{}c@{}}CARLA\\ \cite{dosovitskiy2017carla}\end{tabular} &
  \begin{tabular}[c]{@{}c@{}}LGSVL\\\cite{noauthor_svl_nodate} \end{tabular} \\ \hline
\multicolumn{1}{|c|}{\begin{tabular}[c]{@{}c@{}}Sensors \\ support\end{tabular}} &
  \multicolumn{1}{c|}{\checkmark} &
  \multicolumn{1}{c|}{\checkmark} &
  \multicolumn{1}{c|}{\checkmark} &
  \multicolumn{1}{c|}{\checkmark} &
  \checkmark \\ \hline
\multicolumn{1}{|c|}{\begin{tabular}[c]{@{}c@{}}Weather \\ condition\end{tabular}} &
  \multicolumn{1}{c|}{} &
  \multicolumn{1}{c|}{} &
  \multicolumn{1}{c|}{\checkmark} &
  \multicolumn{1}{c|}{\checkmark} &
  \checkmark \\ \hline
\multicolumn{1}{|c|}{\begin{tabular}[c]{@{}c@{}}Camera \\ calibration\end{tabular}} &
  \multicolumn{1}{c|}{\checkmark} &
  \multicolumn{1}{c|}{} &
  \multicolumn{1}{c|}{\checkmark} &
  \multicolumn{1}{c|}{\checkmark} &
   \\ \hline
\multicolumn{1}{|c|}{\begin{tabular}[c]{@{}c@{}}Path \\ planning\end{tabular}} &
  \multicolumn{1}{c|}{\checkmark} &
  \multicolumn{1}{c|}{\checkmark} &
  \multicolumn{1}{c|}{\checkmark} &
  \multicolumn{1}{c|}{\checkmark} &
  \checkmark \\ \hline
\multicolumn{1}{|c|}{\begin{tabular}[c]{@{}c@{}}Vehicle \\ dynamics\end{tabular}} &
  \multicolumn{1}{c|}{\checkmark} &
  \multicolumn{1}{c|}{\checkmark} &
  \multicolumn{1}{c|}{\checkmark} &
  \multicolumn{1}{c|}{\checkmark} &
  \checkmark \\ \hline
\multicolumn{1}{|c|}{\begin{tabular}[c]{@{}c@{}}Virtual \\ environment\end{tabular}} &
  \multicolumn{1}{c|}{} &
  \multicolumn{1}{c|}{\checkmark} &
  \multicolumn{1}{c|}{\checkmark} &
  \multicolumn{1}{c|}{\checkmark} &
  \checkmark \\ \hline
\multicolumn{1}{|c|}{\begin{tabular}[c]{@{}c@{}}Infrastructure\\ fabrication\end{tabular}} &
  \multicolumn{1}{c|}{\checkmark} &
  \multicolumn{1}{c|}{\checkmark} &
  \multicolumn{1}{c|}{\checkmark} &
  \multicolumn{1}{c|}{\checkmark} &
  \checkmark \\ \hline
\multicolumn{1}{|c|}{\begin{tabular}[c]{@{}c@{}}Scenarios \\ simulator\end{tabular}} &
  \multicolumn{1}{c|}{\checkmark} &
  \multicolumn{1}{c|}{\checkmark} &
  \multicolumn{1}{c|}{\checkmark} &
  \multicolumn{1}{c|}{\checkmark} &
  \checkmark \\ \hline
\multicolumn{1}{|c|}{\begin{tabular}[c]{@{}c@{}}Ground \\ truth\end{tabular}} &
  \multicolumn{1}{c|}{\checkmark} &
  \multicolumn{1}{c|}{} &
  \multicolumn{1}{c|}{} &
  \multicolumn{1}{c|}{\checkmark} &
  \checkmark \\ \hline
\multicolumn{1}{|c|}{\begin{tabular}[c]{@{}c@{}}Simulator \\ connectivity\end{tabular}} &
  \multicolumn{1}{c|}{\checkmark} &
  \multicolumn{1}{c|}{\checkmark} &
  \multicolumn{1}{c|}{\checkmark} &
  \multicolumn{1}{c|}{\checkmark} &
  \checkmark \\ \hline
\multicolumn{1}{|c|}{\begin{tabular}[c]{@{}c@{}}System\\ scalability\end{tabular}} &
  \multicolumn{1}{c|}{} &
  \multicolumn{1}{c|}{} &
  \multicolumn{1}{c|}{} &
  \multicolumn{1}{c|}{\checkmark} &
  \checkmark \\ \hline
\multicolumn{1}{|c|}{\begin{tabular}[c]{@{}c@{}}Open \\ source\end{tabular}} &
  \multicolumn{1}{c|}{} &
  \multicolumn{1}{c|}{} &
  \multicolumn{1}{c|}{} &
  \multicolumn{1}{c|}{\checkmark} &
  \checkmark \\ \hline
\multicolumn{1}{|c|}{\begin{tabular}[c]{@{}c@{}}System\\ stable\end{tabular}} &
  \multicolumn{1}{c|}{\checkmark} &
  \multicolumn{1}{c|}{\checkmark} &
  \multicolumn{1}{c|}{\checkmark} &
  \multicolumn{1}{c|}{\checkmark} &
  \checkmark \\ \hline
\multicolumn{1}{|c|}{\begin{tabular}[c]{@{}c@{}}System\\ portable\end{tabular}} &
  \multicolumn{1}{c|}{\checkmark} &
  \multicolumn{1}{c|}{\checkmark} &
  \multicolumn{1}{c|}{\checkmark} &
  \multicolumn{1}{c|}{\checkmark} &
  \checkmark \\ \hline
\multicolumn{1}{|c|}{\begin{tabular}[c]{@{}c@{}}API \\ flexibility\end{tabular}} &
  \multicolumn{1}{c|}{\checkmark} &
  \multicolumn{1}{c|}{\checkmark} &
  \multicolumn{1}{c|}{} &
  \multicolumn{1}{c|}{\checkmark} &
  \checkmark \\ \hline
\end{tabular}
\label{sim}
\end{table*}

 Cooperative perception methods \cite{CoCa3D:23,xiang2023hm,wang2023core} include V2V (vehicle-to-vehicle), V2I (vehicle-to-infrastructure), and V2X (vehicle-to-everything) modes. Future works should focus on enhancing the transmission efficiency within collaboration systems while safeguarding data privacy.

\subsection{Large language and vision models}
Large vision models have emerged as a prominent trend in AI. By harnessing the advancements in these models, various domains can benefit from their integration. Visual prompts have become essential aids for understanding visuals across diverse domains and enhancing model capacity for interpreting visual data. Presently, SAM-Track \cite{cheng2023segment} for object tracking and VIMA \cite{jiang2022vima} for robot action manipulation showcase potential, implying that these large models can optimize visual recognition systems. Moreover, we can effectively utilize large language and vision models through transfer learning, domain adaptation, and fine-tuning. We can transfer insights from a larger model to a smaller one, emphasizing the importance of compact transfer of knowledge and applying it to novel tasks while upholding performance and adaptability, especially in contexts like autonomous driving. Future efforts must focus on designing large vision models tailored explicitly to autonomous driving and prompt fine-tuning to guide tasks related to perception and control.

\section{Conclusion}\label{conclusion}

Over the past few years, there has been significant interest in End-to-End autonomous driving due to the simplicity of its design compared to conventional modular autonomous driving. We develop a taxonomy based on modalities, learning, and training methodology and investigate the potential of leveraging domain adaptation approaches to optimize the training process. Furthermore, the paper explores evaluation framework that encompasses both open and closed-loop assessments, enabling a comprehensive analysis of system performance. To facilitate further research and development in the domain, we compile a summarized list of advancements, publicly available datasets and simulators. The paper also explores potential solutions proposed by different articles regarding safety and explainability. Despite the impressive performance of End-to-End approaches, there is a need for continued exploration and improvement in safety and interpretability to achieve broader technology acceptance.

\bibliographystyle{elsarticle-num}

\bibliography{cas-refs}

\newpage

\bio{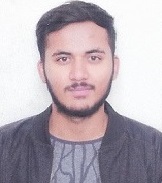}
Pranav Singh Chib is a Ph.D. candidate in the Computer Science and Engineering department at the Indian Institute of Technology, Roorkee. He holds a Post-Graduate Computer Science and Technology Specialization from Jawaharlal Nehru University, New Delhi. Pranav's research interests lie in machine learning, computer vision, and autonomous driving. His ongoing doctoral studies focus on contributing to advancements in autonomous driving and deep learning. 
\endbio

\bio{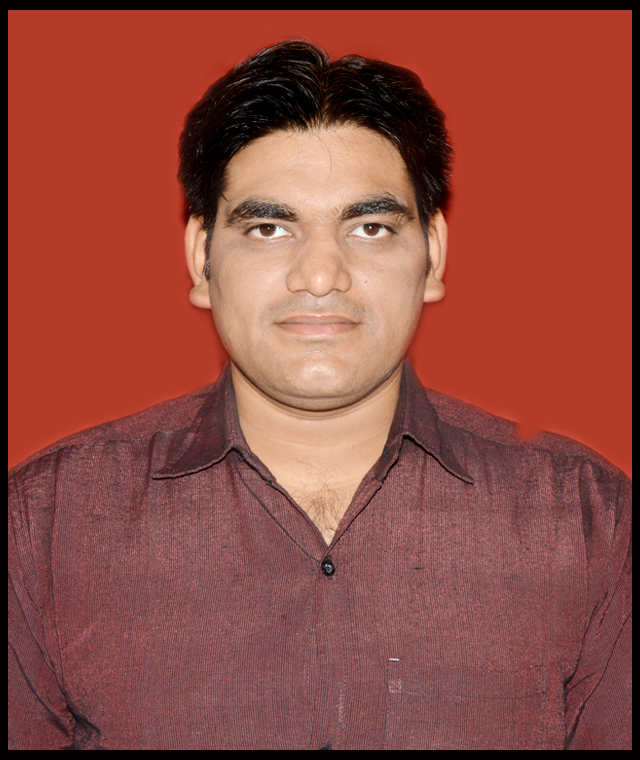}
Pravendra Singh received his Ph.D. degree from IIT Kanpur. He is currently an Assistant Professor in the CSE department at IIT Roorkee, India. His research interests include deep learning, machine learning, computer vision, and artificial intelligence. He has published papers at internationally reputable conferences and journals, including IEEE TPAMI, IJCV, CVPR, ECCV, NeurIPS, AAAI, IJCAI, IJCV, Pattern Recognition, Neural Networks, Knowledge-Based Systems, Neurocomputing, and others.
\endbio

\end{document}